\documentclass{article}

\usepackage[preprint]{neurips_2020}

\usepackage[utf8]{inputenc} 
\usepackage[T1]{fontenc}    
\usepackage[hidelinks]{hyperref}       
\usepackage{url}            
\usepackage{booktabs}       
\usepackage{amsfonts}       
\usepackage{nicefrac}       
\usepackage{microtype}      

\usepackage{algorithm}
\usepackage[noend]{algpseudocode}
\usepackage{graphicx}
\usepackage{amsmath}
\usepackage{amssymb}
\usepackage{amsthm}
\usepackage{multirow}
\usepackage{multicol}
\usepackage{enumitem}
\usepackage{caption}
\usepackage{subcaption}

\newcommand{\ddx}[2]{\frac{\partial #1}{\partial #2}}
\newcommand{\vtheta}{\boldsymbol{\theta}}

\title{Revisiting Loss Modelling for Unstructured Pruning}

\author{%
  C\'esar Laurent$^{1}$, Camille Ballas$^{2}$, Thomas George$^{1}$, Nicolas Ballas$^{3}$, Pascal Vincent$^{1,3,4}$ \\
  $^{1}$Mila, Universit\'e de Montr\'eal; \\
  $^{2}$Insight Centre for Data Analytics, Dublin City University; \\
  $^{3}$Facebook AI Research; \\
  $^{4}$Canadian Institute for Advanced Research (CIFAR)
}

\begin{document}

\maketitle

\begin{abstract}

By removing parameters from deep neural networks, unstructured pruning methods aim at cutting down memory footprint and computational cost, while maintaining prediction accuracy. In order to tackle this otherwise intractable problem, many of these methods model the loss landscape using first or second order Taylor expansions to identify which parameters can be discarded. We revisit loss modelling for unstructured pruning: we show the importance of ensuring locality of the pruning steps. We systematically compare first and second order Taylor expansions and empirically show that both can reach similar levels of performance. Finally, we show that better preserving the original network function does not necessarily transfer to better performing networks after fine-tuning, suggesting that only considering the impact of pruning on the loss might not be a sufficient objective to design good pruning criteria.

\end{abstract}

\section{Introduction}

Neural networks are getting bigger and bigger, requiring more and more computational resources not only for training, but also when used for inference.
However, resources are sometimes limited, especially on mobile devices and low-power chips.
In unstructured pruning, the goal is to remove some parameters (i.e. setting them to zeros), while still maintaining good prediction performances. This is fundamentally a combinatorial optimization problem which is intractable even for small scale neural networks, and thus various heuristics have been developed to prune the model either before training \citep{lee2018snip, Wang2020Picking}, during training \citep{louizos2017learning, molchanov2017variational, ding2019global}, or in an iterative training/fine-tuning fashion \citep{lecun1990optimal, hassibi1993second, song2015learning, Frankle2018TheLT, Renda2020Comparing}.

Early pruning work Optimal Brain Damage (OBD) \citep{lecun1990optimal}, and later Optimal Brain Surgeon (OBS) \citep{hassibi1993second}, proposed to estimate the importance of each parameter by approximating the effect of removing it, using the second order term of a Taylor expansion of the loss function around converged parameters. This type of approach involves computing the Hessian, which is challenging to compute since it scales quadratically with the number of parameters in the network. Several approximations have thus been explored in the literature \citep{lecun1990optimal, hassibi1993second, heskes2000natural, zeng2019mlprune, wang2019eigendamage}. However, state-of-the-art unstructured pruning methods typically rely on Magnitude Pruning (MP) \citep{song2015learning}, a simple and computationally cheap criterion based on weight magnitude, that works extremely well in practice \citep{Renda2020Comparing}.

This paper revisits linear and quadratic models of the local loss landscape for unstructured pruning. In particular, since these models are \emph{local} approximations and thus assume that pruning steps correspond to small vectors in parameter space, we propose to investigate how this locality assumption affects their performance. Moreover, we show that the \emph{convergence} assumption behind OBD and OBS, which is overlooked and violated in current methods, can be relaxed by maintaining the gradient term in the quadratic model, removing the need of fine-tuning phases after each pruning iteration. Finally, to prevent having to compute second order information, we propose to compare quadratic models to simpler linear models.

While our empirical study demonstrates that pruning criteria based on linear and quadratic loss models are good at preserving the training loss, it also shows that this benefit does not necessarily transfer to better networks after fine-tuning, suggesting that preserving the loss might not be the best objective to optimize for.

Our contributions can be summarized as follows:
\setlist{nolistsep}
\begin{enumerate}[noitemsep]
    \item We present pruning criteria based on both linear and quadratic models of the loss, and show how they compare at preserving training loss compared to OBD and MP.
    \item We study two strategies to better enforce locality in the pruning steps, iterative pruning and step size regularisation, and show how they improve the quality of the criteria.
    \item We show that using pruning criteria that are better at preserving the loss does not necessarily transfer to better fine-tuned networks, raising questions about the adequacy of such criteria.
\end{enumerate}

\section{Background: Unstructured Pruning}

\subsection{Unstructured Pruning Problem Formulation}

For a given architecture, neural networks are a family of functions $f_{\vtheta}: \mathcal X\to \mathcal Y$ from an input space $\mathcal X$ to an output space $\mathcal Y$, where $\vtheta \in \mathbb{R}^D$ is the vector that contains all the parameters of the network. Neural networks are usually trained by seeking parameters $\vtheta$ that minimize the empirical risk $\mathcal L(\vtheta) = \frac{1}{N} \sum_{i} \ell \left( f_{\vtheta} \left( x_i \right), t_i \right)$ of a loss function $\ell$ on a training dataset $\mathcal{D}=\left\{ \left(x_{i},t_{i}\right)\right\} _{1\leq i\leq N}$, composed of $N$ (example, target) pairs.

The goal of unstructured pruning is to find a step $\Delta\vtheta$ to add to the current parameters $\vtheta$ such that $\Vert\vtheta + \Delta \vtheta\Vert_0 = (1 - \kappa)D$, i.e. the parameter vector after pruning is of desired sparsity $\kappa \in [0, 1]$. While doing so, the performance of the pruned network should be maintained, so $\mathcal L(\vtheta + \Delta\vtheta)$ should not differ much from $\mathcal L(\vtheta)$. Unstructured pruning thus amounts to the following minimization problem:
\begin{equation}
    \underset{\Delta\boldsymbol{\theta}}{\text{minimize}} \quad  \Delta \mathcal L (\vtheta, \Delta\vtheta) \stackrel{\text{def}}{=} \left\vert \mathcal{L}(\vtheta + \Delta\vtheta) - \mathcal L(\vtheta) \right\vert \qquad \text{s.t.} \quad \left\Vert \vtheta + \Delta\vtheta \right\Vert_{0} = (1 - \kappa)D
    \label{eq:pruning_delta}
\end{equation}
Directly solving this problem would require evaluating $\mathcal{L}(\vtheta + \Delta\vtheta)$ for all possible values of $\Delta\vtheta$, which is prohibitively expensive, so one needs to rely on heuristics to find good solutions.

\subsection{Optimal Brain Damage Criterion}

Optimal Brain Damage (OBD) \citep{lecun1990optimal} proposes to use a quadratic modelling of $\mathcal{L}(\vtheta + \Delta\vtheta)$, leading to the following approximation of $\Delta \mathcal L (\vtheta, \Delta\vtheta)$:
\begin{align}
    \Delta \mathcal L^{QM} (\vtheta, \Delta\vtheta) &= \left\vert \ddx{\mathcal L(\vtheta)}{\vtheta}^\top \Delta\vtheta + \frac{1}{2} \Delta\vtheta ^\top \mathbf H (\vtheta) \Delta\vtheta \right\vert
    \label{eq:qm}
\end{align}
where $\mathbf H(\vtheta)$ is the Hessian of $\mathcal{L}(\vtheta)$. $\mathbf H(\vtheta)$ being intractable, even for small-scale networks, its Generalized Gauss-Newton approximation $\mathbf G(\vtheta)$ \citep{schraudolph2002fast} is used in practice, as detailed in Appendix~\ref{sec:GGN}.\footnote{Although \citet{lecun1990optimal} uses $\mathbf H(\vtheta)$ in the equations of OBD, it is actually $\mathbf G(\vtheta)$ which is used in practice \citep{YannNIPS}.}
Then, two more approximations are made: first, it assumes the training of the network has converged, thus the gradient of the loss wrt $\vtheta$ is $0$, which makes the linear term vanish. Then, it neglects the interactions between parameters, which corresponds to a diagonal approximation of $\mathbf G(\vtheta)$, leading to the following model:
\begin{align}
    \Delta \mathcal L^{OBD}(\vtheta, \Delta\vtheta_k) \approx \frac{1}{2} \mathbf G_{kk}(\vtheta) \Delta\vtheta_k^2 
        \qquad \Rightarrow \qquad s_k^{\text{OBD}} = \frac{1}{2} \mathbf G_{kk}(\vtheta) \vtheta_k^2
    \label{eq:sens_OBD}
\end{align}
$s_k^{\text{OBD}}$ is the \textit{saliency} of each parameter, estimating how much the loss will change if that parameter is pruned, i.e. if $\Delta\vtheta_k = -\vtheta_k$. Parameters can thus be ranked by order of importance, and the ones with the smallest saliencies (i.e. the least influence on the loss) are pruned, while the ones with the biggest saliencies are kept unchanged. This can be interpreted as finding and applying a binary mask $\mathbf m \in \{0, 1\}^D$ to the parameters such that $\vtheta + \Delta\vtheta = \vtheta \odot \mathbf m$, where $\odot$ is the element-wise product.

\subsection{Magnitude Pruning Criterion}
Magnitude Pruning (MP) \citep{song2015learning}, is a popular pruning criterion in which the saliency is simply based on the norm of the parameter:
\begin{align}
    s_k^{\text{MP}} = \vtheta_k^2
    \label{eq:sens_mp}
\end{align}
Despite its simplicity, MP works extremely well in practice \citep{gale2019state}, and is used in current state-of-the-art methods \citep{Renda2020Comparing}. It will serve as baseline in all our experiments.

\subsection{Optimal Brain Surgeon}

Optimal Brain Surgeon (OBS) \citep{hassibi1993second}, relies on the same quadratic model as OBD to solve the minimization problem given in Equation~\ref{eq:pruning_delta}, but uses the Lagrangian formulation to include the constraint to the solution of the minimization problem. Since OBS requires to compute the inverse of $\mathbf H(\vtheta)$, several approximations have been explored in the literature, including diagonal, as in the original OBS, Kronecker-factored \citep{martens2015optimizing} as in ML-Prune \citep{zeng2019mlprune}, or diagonal, but in an Kronecker-factored Eigenbasis \citep{george2018fast}, as in EigenDamage \citep{wang2019eigendamage}. While we use OBD in our demonstrations and experimental setup, everything presented in this paper can also be used in OBS-based methods. We leave that for future work.

\section{Revisiting Loss Modelling for Unstructured Pruning}
\label{sec:revisiting}

In this work, we investigate linear and quadratic models of the loss function and their performance at pruning neural networks. In our empirical study, we aim at answering the following questions:
\setlist{nolistsep}
\begin{enumerate}[noitemsep]
    \item How do criteria based on weight magnitude, linear and quadratic models compare at preserving training loss (i.e. at solving the minimization problem in Equation~\ref{eq:pruning_delta})?
    \item How does the locality assumption behind criteria based on linear and quadratic models affect their performances?
    \item Do pruning criteria better at preserving the loss lead to better fine-tuned networks?
\end{enumerate}
We now describe the linear and quadratic models we use, as well as the strategies to enforce locality of the pruning steps.

\subsection{Linear and Quadratic Models}
\label{sec:LMQM}

In current training strategies, regularization techniques such as early stopping or dropout \citep{srivastava2014dropout} are often used to counteract overfitting. In these setups, there is no reason to assume that the training has converged, implying that the linear term in the Taylor expansion should not be neglected. Thus, one can build a pruning criterion similar to OBD that includes the gradient term in the quadratic model from Equation~\ref{eq:qm}, leading to the following saliencies:\footnote{Concurrent work explores similar idea for Optimal Brain Surgeon \citep{singh2020woodfisher}.}
\begin{equation}
    \Delta \mathcal L^{QM}(\vtheta, \Delta\vtheta_k) \approx \left\vert \ddx{\mathcal L(\vtheta)}{\vtheta_k}^\top \Delta\vtheta_k + \frac{1}{2} \mathbf G_{kk}(\vtheta) \Delta\vtheta_k^2  \right\vert \Rightarrow s_k^{\text{QM}} = \left\vert - \ddx{\mathcal L(\vtheta)}{\vtheta_k} \vtheta_k + \frac{1}{2} \mathbf G_{kk}(\vtheta) \vtheta_k^2 \right\vert \label{eq:sens_qm}
\end{equation}
Recall the constraint $\Delta\vtheta_k \in \{-\vtheta_k, 0\}$, hence the saliencies. This criterion generalizes OBD for networks that are not at convergence, and provide similar saliencies for networks that have converged.

To suppress the computational cost associated with second order information, which is prohibitive for large scale neural networks, one can use a linear model (LM) instead of a quadratic one to approximate $\Delta \mathcal L (\vtheta, \Delta\vtheta)$, leading to the following approximation and saliencies:\footnote{The saliencies of the linear model are very related to the criterion used in Single-shot Network Pruning \citep{lee2018snip}, as demonstrated by \citet{Wang2020Picking}.}
\begin{align}
    \Delta \mathcal L^{LM} (\vtheta, \Delta\vtheta) &= \left\vert \ddx{\mathcal L(\vtheta)}{\vtheta}^\top \Delta\vtheta \right\vert \quad \Rightarrow \quad s_k^{\text{LM}} = \left\vert \ddx{\mathcal L(\vtheta)}{\vtheta_k} \vtheta_k \right\vert
    \label{eq:sens_lm}
\end{align}

\subsection{Enforcing Locality}
\label{sec:loc}

One important point to keep in mind is that linear and quadratic models are \emph{local} approximations, and are only faithful in a small neighbourhood of the current parameters. Explicitly showing the terms that are neglected, we have:
\begin{align}
    \Delta \mathcal L(\vtheta, \Delta\vtheta) &= \Delta \mathcal L^{LM} (\vtheta, \Delta\vtheta) + \mathcal O(\Vert \Delta \vtheta \Vert^2_2) = \Delta \mathcal L^{QM} (\vtheta, \Delta\vtheta) + \mathcal O(\Vert \Delta \vtheta \Vert^3_2)
\end{align}
So when approximating $\Delta \mathcal L$ with $\Delta \mathcal L^{LM}$ we neglect the terms in $\mathcal O(\Vert \Delta \vtheta \Vert^2_2)$, and when approximating $\Delta \mathcal L$ with $\Delta \mathcal L^{QM}$ we neglect the terms in $\mathcal O(\Vert \Delta \vtheta \Vert^3_2)$. Both approximations are thus only valid in a small neighbourhood of $\vtheta$, and are extremely likely to be wrong when $\Vert \Delta \vtheta \Vert_2$ is large. We list here different tricks to prevent this from happening.

\paragraph{Pruning Iteratively} $\Vert \Delta \vtheta \Vert_2$ can be large when a large portion of the parameters is pruned at once. An easy fix typically used to mitigate this issue is to perform the pruning in several iterations, re-estimating the model at each iteration. The number of iterations, which we denote by $\pi$, is typically overlooked (e.g. both \citet{zeng2019mlprune} and \citet{wang2019eigendamage} use only 6 pruning iterations). Our experiments show that it has a drastic impact on the performances. Note that, without fine-tuning phases between the different pruning iterations, this strategy violates the \emph{convergence} assumption behind OBD and OBS, since after the first iteration of pruning the network is no more at convergence.

The sparsity at each iteration can be increased either linearly, where each step prunes the same number of parameter, or exponentially, where the number of parameters pruned at each iteration gets smaller and smaller. The later is typically used in the literature \citep{zeng2019mlprune,wang2019eigendamage,Frankle2018TheLT,Renda2020Comparing}. We compare them in Section~\ref{sec:before_ft}.

\paragraph{Constraining the Step Size}
As it is often done when using quadratic models (e.g. \citet{nocedal2006numerical}), one can penalize the model when it decides to take steps that are too large, in order to stay in a region where we can trust the model. This can be done by simply adding the norm penalty $\frac{\lambda}{2} \left\Vert \vtheta_k \right\Vert_2^2$ to the saliencies computed by any criterion (Equations~\ref{eq:sens_OBD}, \ref{eq:sens_qm} or \ref{eq:sens_lm}), where $\lambda$ is an hyper-parameter that controls the strength of the constraint: a small value of $\lambda$ leaves the saliencies unchanged, and a large value of $\lambda$ transforms the pruning criterion into MP (Equation~\ref{eq:sens_mp}).

\paragraph{Other Considerations}
$\Vert \Delta \vtheta \Vert_2$ can be large if $\vtheta$ is large itself. This is dependent on the training procedure of the network, but can be easily mitigated by constraining the norm of the weights, which can be done using $L_2$ regularisation or weight decay. Since nowadays weight decay is almost systematically used by default when training networks (e.g. \citet{he2016identity, xie2017aggregated, devlin2018bert}), we do not investigate this further.

\section{Methodology}

We follow most of the recommendations from \citet{blalock2020state}. For fair comparison between criteria, all experiments are from our own PyTorch \citep{paszke2017automatic} re-implementation, and ran on V100 GPUs.\footnote{Our code available at: \url{https://github.com/Thrandis/loss-models-pruning}} All experiments are run using 5 different random seeds, and both mean and standard deviations are reported. We experiment on a MLP on MNIST, and both VGG11~\citep{simonyan2014very} and a pre-activation residual network 18 \citep{he2016identity}, on CIFAR10 \citep{krizhevsky2009learning}, to have variability in architectures, while using networks with good performance to number of parameters ratio. Although MNIST is not considered a good benchmark for pruning \citep{blalock2020state}, it can still be used to compare  the ability of different criteria to solve the minimization problem in Equation~\ref{eq:pruning_delta}. See Appendix~\ref{sec:setups} for details about splits, data augmentation strategies, architectures, initialisation and hyper-parameters.

\paragraph{Pruning Framework}

Algorithm~\ref{algo:framework} presents the pruning framework used in this work: the network is first trained, then pruned, then fine-tuned once, using the same hyper-parameters as for the original training. Note that because of their convergence assumption, OBD and OBS advocate for fine-tuning after each iteration of pruning. Since LM and QM are not based on this assumption, the proposed framework works with these criteria. While the fine tuning-phase would require hyper-parameters optimisation, \citet{Renda2020Comparing} showed that using the same ones as for the original training usually leads to good results. The hyper-parameters used in our experiments are provided in Appendix~\ref{sec:setups}.

\begin{algorithm}[htb]
    \caption{Pruning Framework}\label{algo:framework}
    \begin{algorithmic}[1]
        \Require Network $f_{\vtheta}$ with $\vtheta \in \mathbb R^D$, dataset $\mathcal D$, number of pruning iterations $\pi$, and sparsitiy $\kappa$.
        \State $f_{\vtheta} \gets$ Training($f_{\vtheta}$, $\mathcal D$)
        \State $\kappa_0 \gets 0$
        \State $\mathbf m \gets \mathbf 1^{D}$
        \For{$i=1$ to $\pi$}
            \State $\kappa_i \gets \kappa_{i-1} + \frac{(\kappa - \kappa_0)}{\pi}\;\:  \textbf{or} \;\: \kappa_i \gets \kappa_{i-1} + (\kappa - \kappa_0)^{i / \pi}$\label{algo:iterative_pruning:pr}
            \algorithmiccomment{Compute sparsity for iteration $i$}
            \State $\boldsymbol{s} \gets \text{Saliencies}(f_{\vtheta \odot \mathbf m}, \mathcal D)$ \phantom{} \algorithmiccomment{Compute saliencies (Equation \ref{eq:sens_OBD}, \ref{eq:sens_mp}, \ref{eq:sens_qm} or \ref{eq:sens_lm}).}
            \State $\boldsymbol{i} \gets \text{argsort}(\boldsymbol s)[:\kappa_i D]$ \algorithmiccomment{Get indexes of smallest saliencies.}
            \State $\mathbf m[\boldsymbol{i}] \gets 0$ \algorithmiccomment{Update the mask.}
        \EndFor
        \State $f_{\vtheta \odot \mathbf m} \gets$ Training($f_{\vtheta \odot \mathbf m}$, $\mathcal D$)
        \algorithmiccomment{Optional fine-tuning}
        \State \textbf{return }$f_{\vtheta \odot \mathbf m}, \mathbf m$
     \end{algorithmic}
\end{algorithm}

\paragraph{Performance Metrics} The performances of the pruning criteria are measured using two metrics: First, we use $\Delta \mathcal L (\vtheta, \Delta\vtheta) = \left\vert \mathcal{L}(\vtheta + \Delta\vtheta) - \mathcal L(\vtheta) \right\vert$, which is the quantity that the pruning criteria are designed to minimize (recall Equation~\ref{eq:pruning_delta}). 
We want to point out that this metric is never reported in practice (except in \citet{wang2019eigendamage}), and thus there is no way of knowing if better loss models indeed result in better loss-preserving criteria. 
Second, we use the validation error gap before/after fine-tuning, which is the metric we ultimately care about when designing pruning methods.

\section{Performances before Fine-tuning}
\label{sec:before_ft}

\begin{figure}[htb!]
    \centering
    \includegraphics[width=0.32\textwidth]{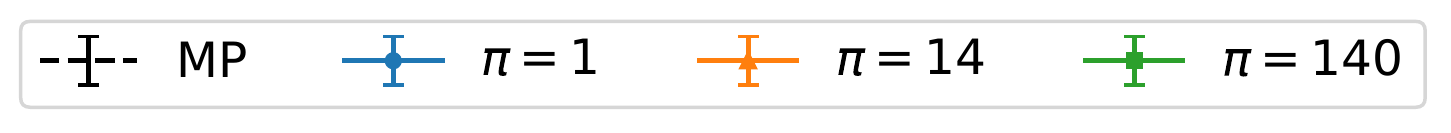}
    \begin{subfigure}[b]{\textwidth}
        \centering
        \includegraphics[width=0.32\textwidth]{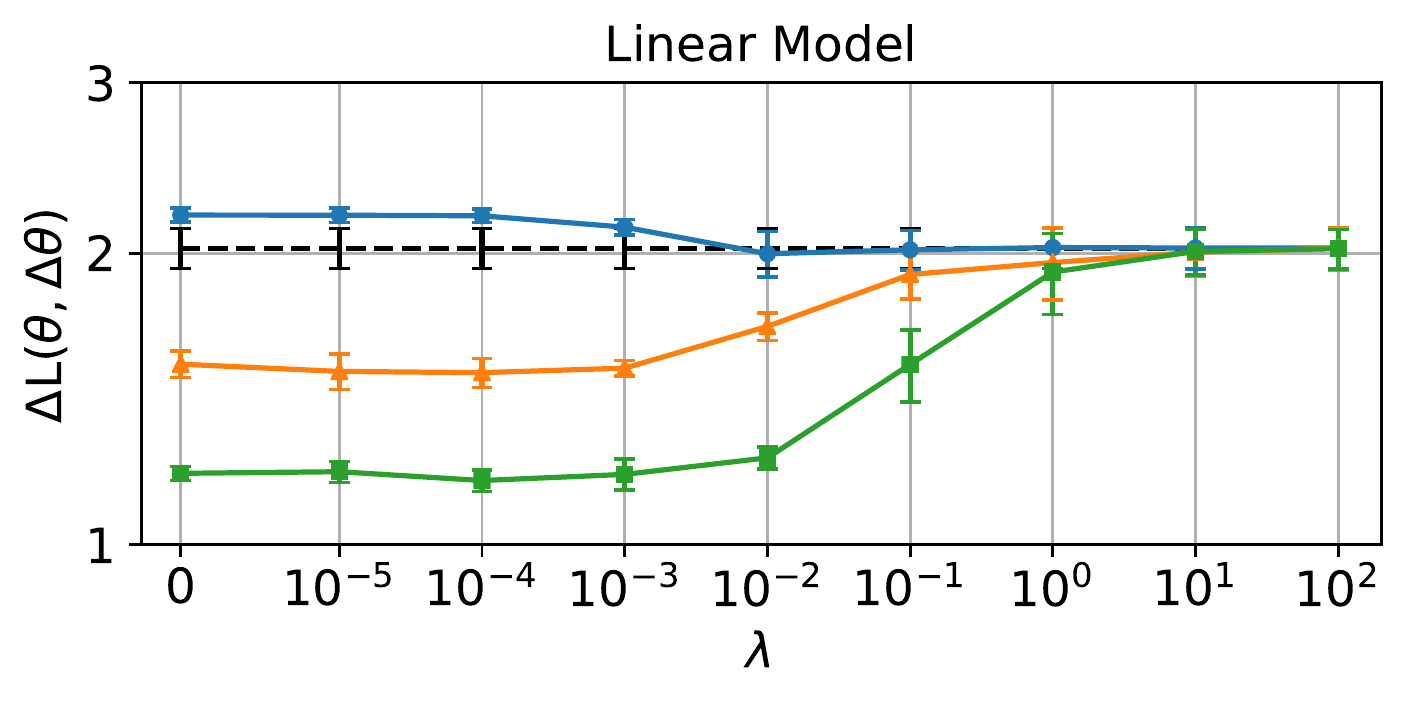}
        \includegraphics[width=0.32\textwidth]{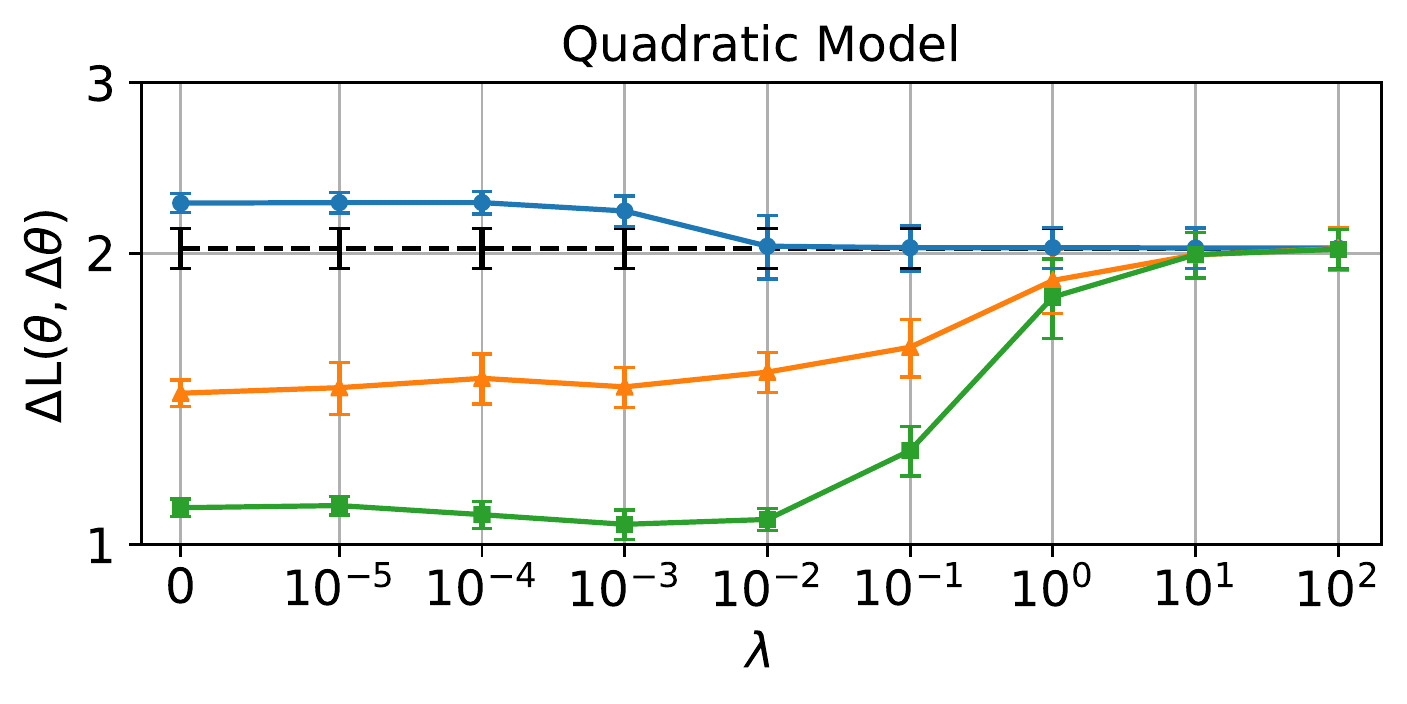}
        \includegraphics[width=0.32\textwidth]{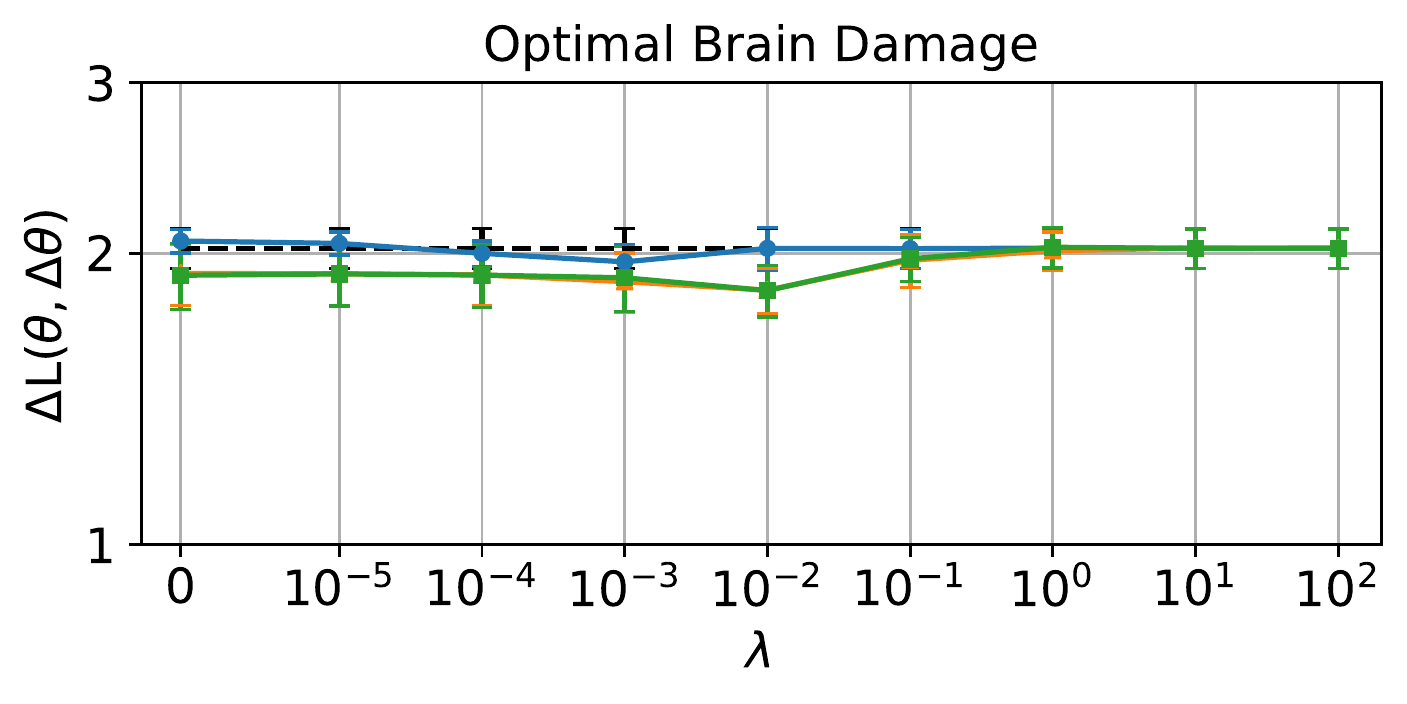}
        \label{fig:loss_lt_mlp}
        \caption{MLP on MNIST with 98.8\% sparsity.}
    \end{subfigure}\\
    \begin{subfigure}[b]{\textwidth}
        \centering
        \includegraphics[width=0.32\textwidth]{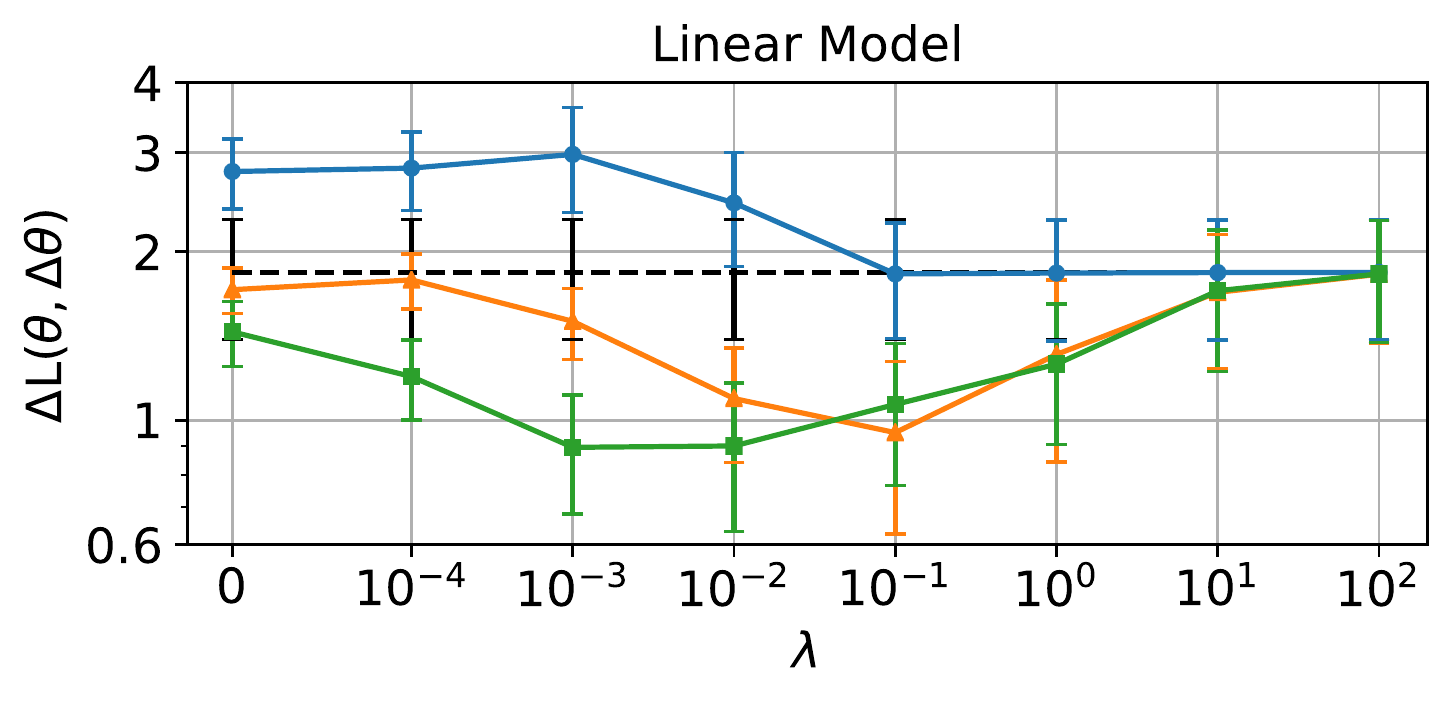}
        \includegraphics[width=0.32\textwidth]{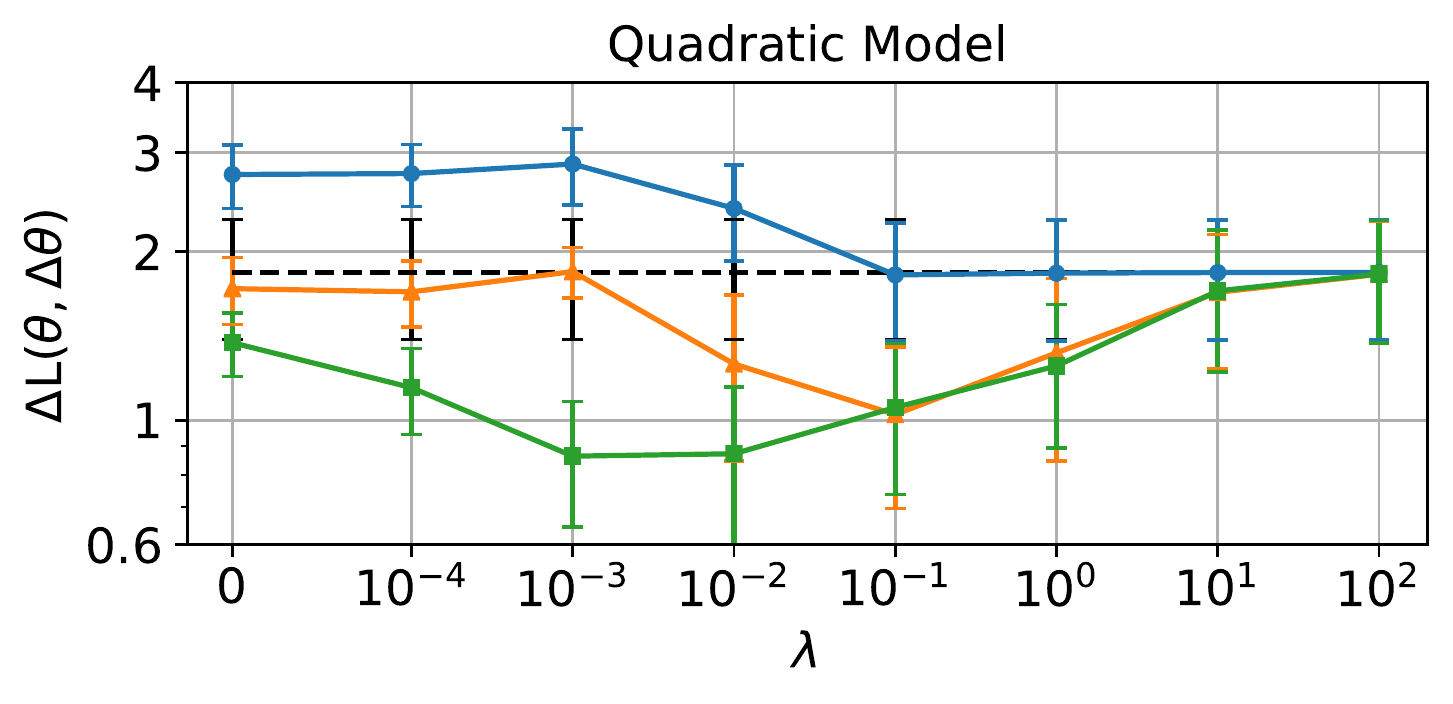}
        \includegraphics[width=0.32\textwidth]{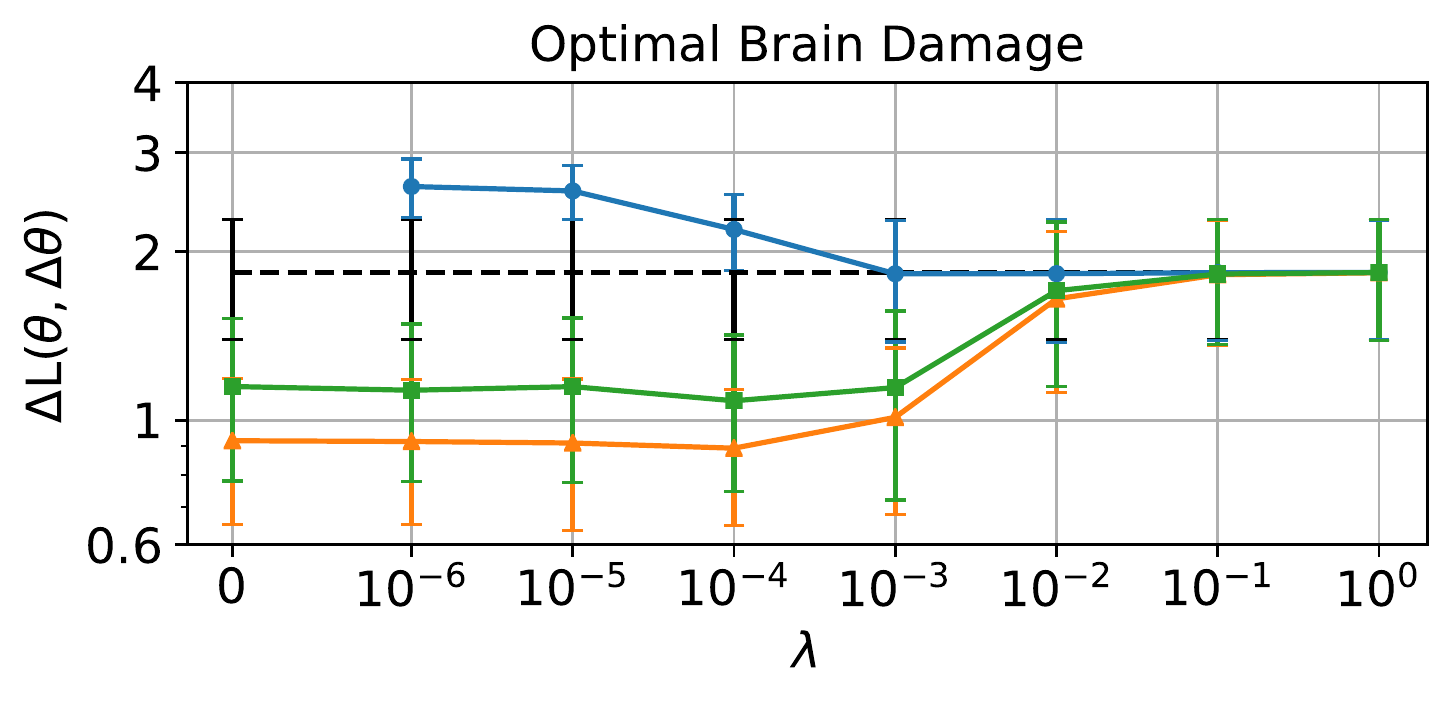}
        \label{fig:loss_lt_vgg}
        \caption{VGG11 on CIFAR10 with 95.6\% sparsity.}
    \end{subfigure}\\
    \begin{subfigure}[b]{\textwidth}
        \centering
        \includegraphics[width=0.32\textwidth]{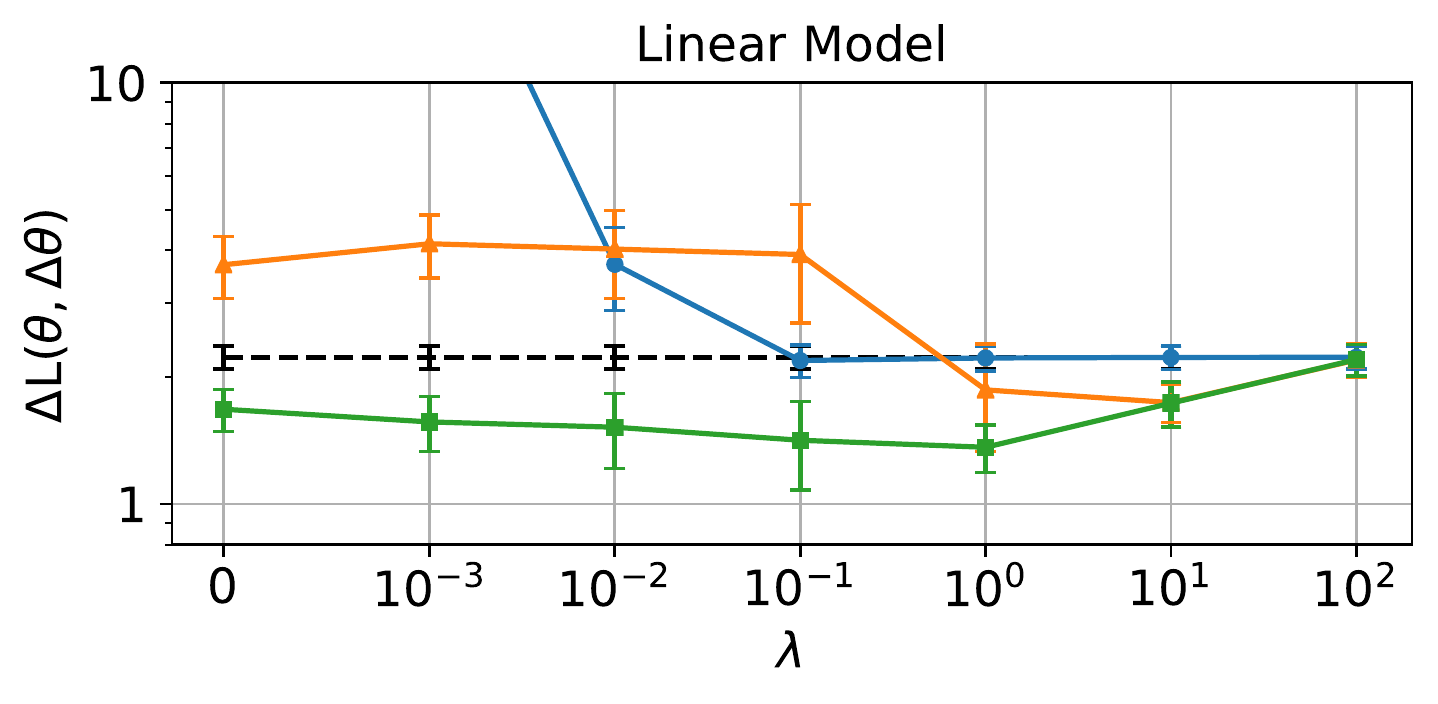}
        \includegraphics[width=0.32\textwidth]{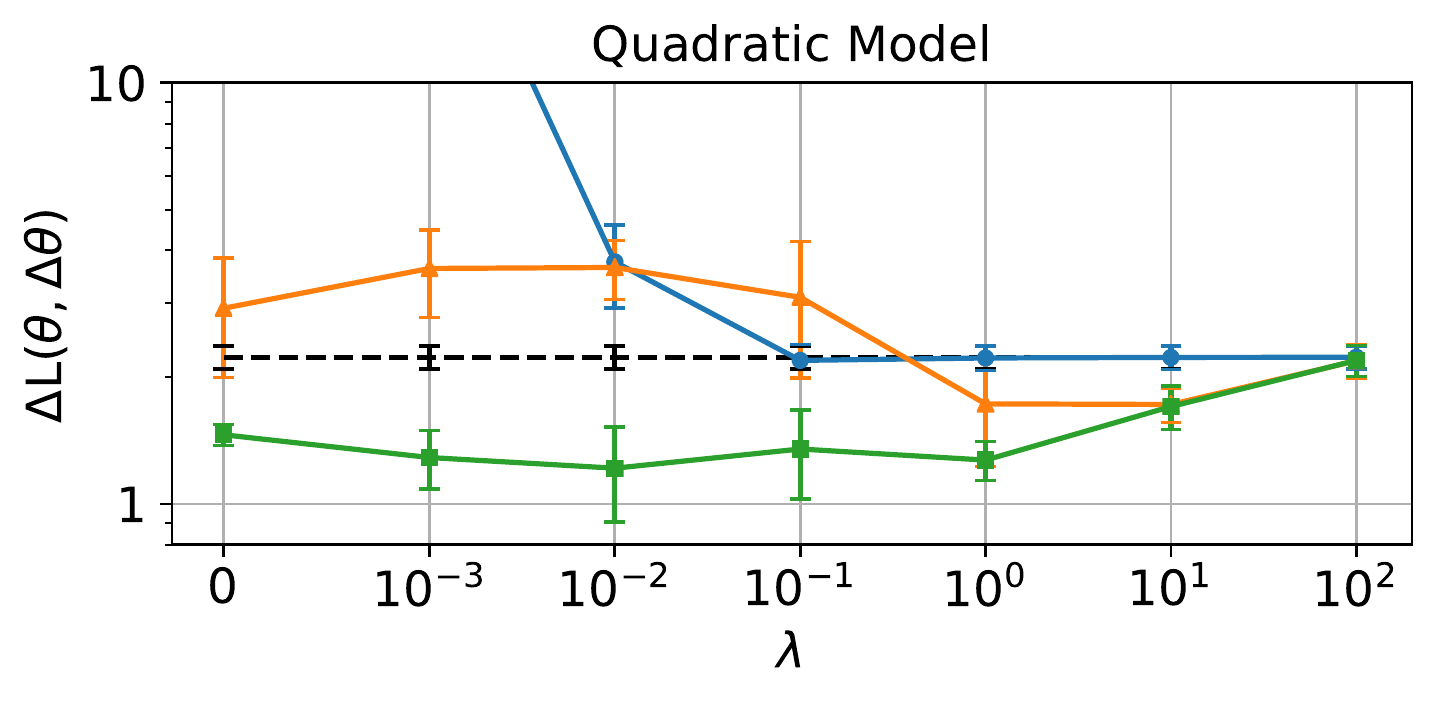}
        \includegraphics[width=0.32\textwidth]{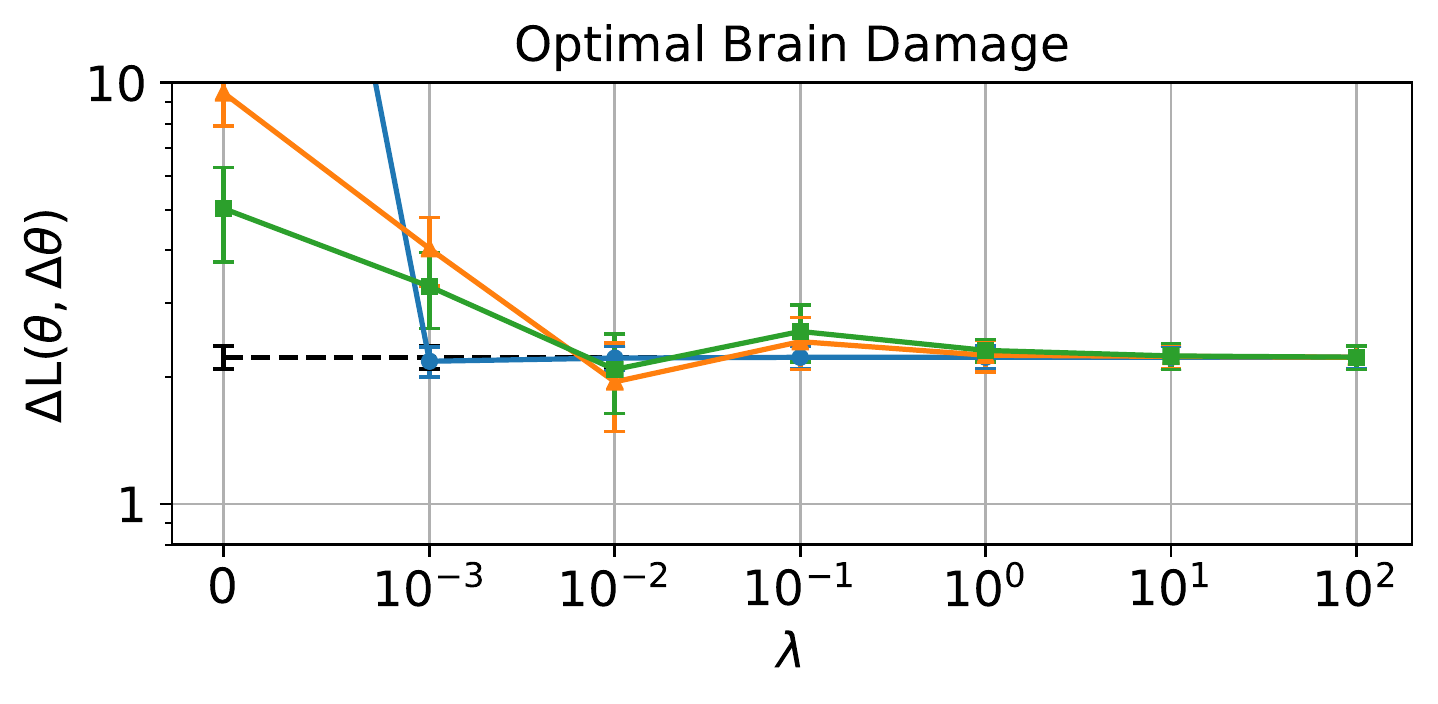}
        \label{fig:loss_lt_resnet}
        \caption{PreActResNet18 on CIFAR10 with 95.6\% sparsity.}
    \end{subfigure}
    \caption{$\Delta \mathcal L(\vtheta, \Delta\vtheta)$ for different number of pruning iteration $\pi$, as a function of $\lambda$, the step size constraint strength, using either (left) LM, (middle) QM or (right) OBD criteria. MP, which is invariant to $\lambda$ and to the number of pruning iterations, is displayed in dashed black. The curves are the mean and the error bars the standard deviation over 5 random seeds. OBD with $\pi=1$ and $\lambda=0$ diverged for all of the 5 seeds. Increasing the number of pruning iterations drastically reduces $\Delta \mathcal L(\vtheta, \Delta\vtheta)$. A $\lambda > 0$ can also help improving performances.}
    \label{fig:loss_lt}
\end{figure}

We evaluate the impact of enforcing locality in the LM, QM and OBS criteria. For each criterion, Figure~\ref{fig:loss_lt} reports $\Delta \mathcal L (\vtheta, \Delta\vtheta)$ as a function of $\lambda$, for different number of pruning iterations $\pi$, using exponential pruning steps, and Figure~\ref{fig:loss_eq} in Appendix show the same results for equally spaced steps. A typical usage of these criteria would be with a regularisation strength $\lambda = 0$ and a number of pruning iterations $\pi \approx 1$. MP, the baseline, which is invariant to both $\lambda$ and $\pi$, is also reported in dashed black. For reference, the networks reached a validation error rate before pruning of $1.47 \pm 0.04$~\% for the MLP, $10.16 \pm 0.29$~\% for VGG11 and $4.87 \pm 0.04$~\% for the PreActResNet18.

\subsection{Impact of the Assumptions behind the Different Criteria}

\paragraph{Locality Assumption} Figure~\ref{fig:loss_lt} shows that increasing the number of pruning iterations can drastically reduce $\Delta \mathcal L(\vtheta, \Delta\vtheta)$ when using LM, QM  and OBS criteria. It demonstrates the importance of applying local steps when pruning. Constraining the steps size through $\frac{\lambda}{2} \left\Vert \vtheta_k \right\Vert_2^2$ can also reduce $\Delta \mathcal L(\vtheta, \Delta\vtheta)$, on CIFAR10 in particular. The trend, however, is less pronounced on MNIST. We hypothesize that this behavior can be explained by the pruning step size. On MNIST, the MLP contains 260k parameters, while VGG11 has 9.7M parameters, so even if we perform 140 iterations of pruning in VGG11, the number of parameters pruned at each iteration is still rather large, which translates to a bigger $\Vert \Delta \vtheta \Vert_2$, which need to be controlled by the regularisation constraint.

\paragraph{Convergence Assumption} When pruning iteratively, we also observe that LM and QM can reach better performances than OBD. Without retraining phases between pruning iterations, we  violate the convergence assumption of OBD. This is however not the case for LM and QM, since they are not based on this assumption. Note that OBD still works reasonably well on VGG11. This could be be related to the depth of VGG11: VGG11 is deeper than the MLP, but not equipped with residual connections like the PreActResNet18. There is also links between OBD and the KL divergence (see Appendix~\ref{sec:GGN}), that could explain the performance of OBD on VGG11.

\subsection{Loss-preserving Capabilities of the Different Criteria}

Table~\ref{tab:loss} contains the best $\Delta \mathcal L(\vtheta, \Delta\vtheta)$ for each of the networks and pruning criteria. Our main observation is the criteria that model the loss (LM and QM in particular) are better at loss-preserving than MP. Furthermore, LM performs similarly as QM, while being less expensive computationally. This suggests that one could use the simpler LM instead of QM for pruning in OBD-like criteria. Similarly to Table~\ref{tab:loss}, Table~\ref{tab:error} in Appendix contains the best validation error gap before/after pruning, where we can observe similar tendencies.

\begin{table}[h!!]
	\centering
	\caption{Summary of the best $\Delta \mathcal L(\vtheta, \Delta\vtheta)$ across values of $\lambda$ for different networks and pruning criteria, with $\pi=140$. QM is better at loss-preserving than other criteria. OBD performs worse than QM, since we violate its convergence assumption when pruning iteratively.}
    \begin{tabular}{@{}lcccc@{}}
        \toprule
        \multirow{2}{*}{\bf Network} & \multicolumn{4}{c}{ \bf $\Delta \mathcal L(\vtheta, \Delta\vtheta)$} \\
        \cmidrule{2-5}
        & \bf MP & \bf OBD & \bf LM & \bf QM \\
        \midrule
        MLP on MNIST & 2.02 $\pm$ 0.10 & 1.83 $\pm$ 0.11 & 1.17 $\pm$ 0.03 & \bf 1.05 $\pm$ 0.04 \\
        VGG11 on CIFAR10 & 1.84 $\pm$ 0.44 & \bf 0.89 $\pm$ 0.24 & \bf 0.90 $\pm$ 0.21 & \bf 0.86 $\pm$ 0.22 \\
        PreActResNet18 on CIFAR10 & 2.23 $\pm$ 0.14 & 1.95 $\pm$ 0.46 & 1.36 $\pm$ 0.18 & \bf 1.22 $\pm$ 0.31 \\
        \bottomrule
    \end{tabular}
    \label{tab:loss}
\end{table}

\subsection{Linear vs Exponential Pruning Steps}

\begin{figure}[htb]
    \centering
    \includegraphics[width=0.33\textwidth]{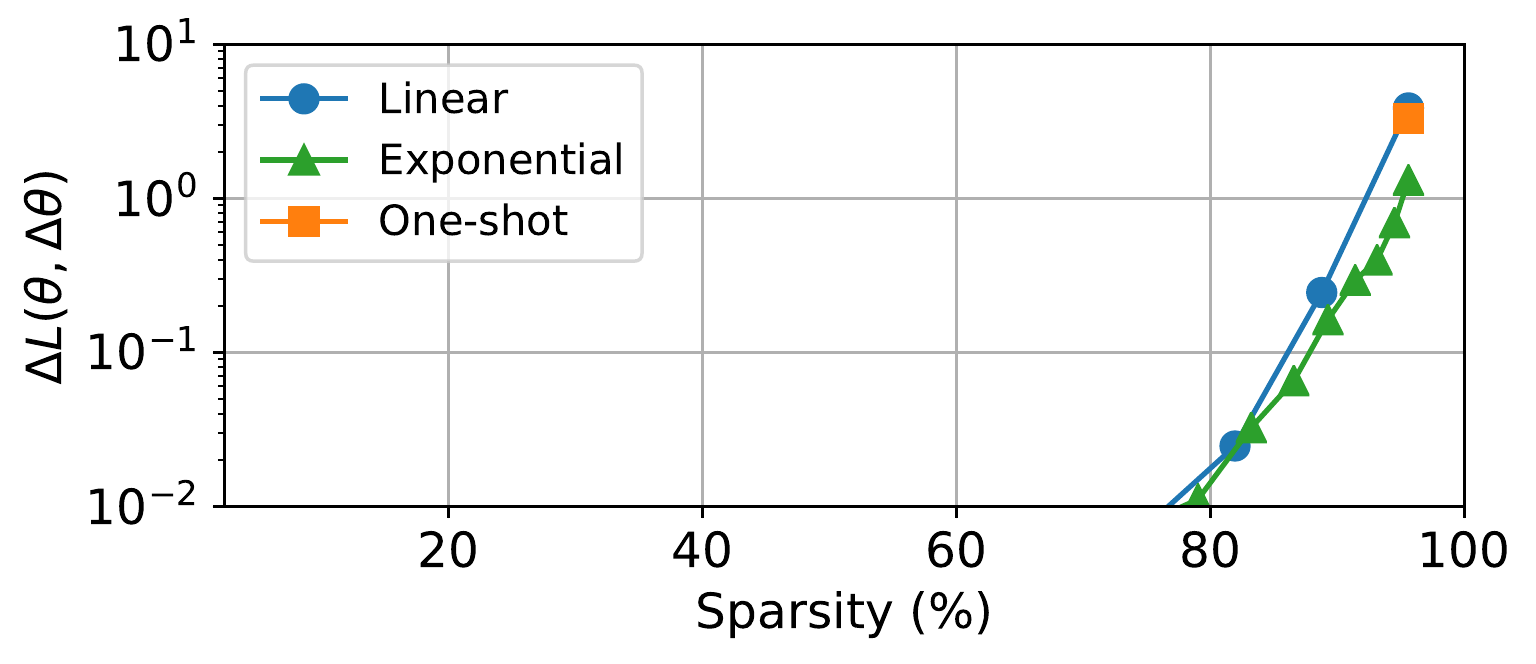}
    \includegraphics[width=0.33\textwidth]{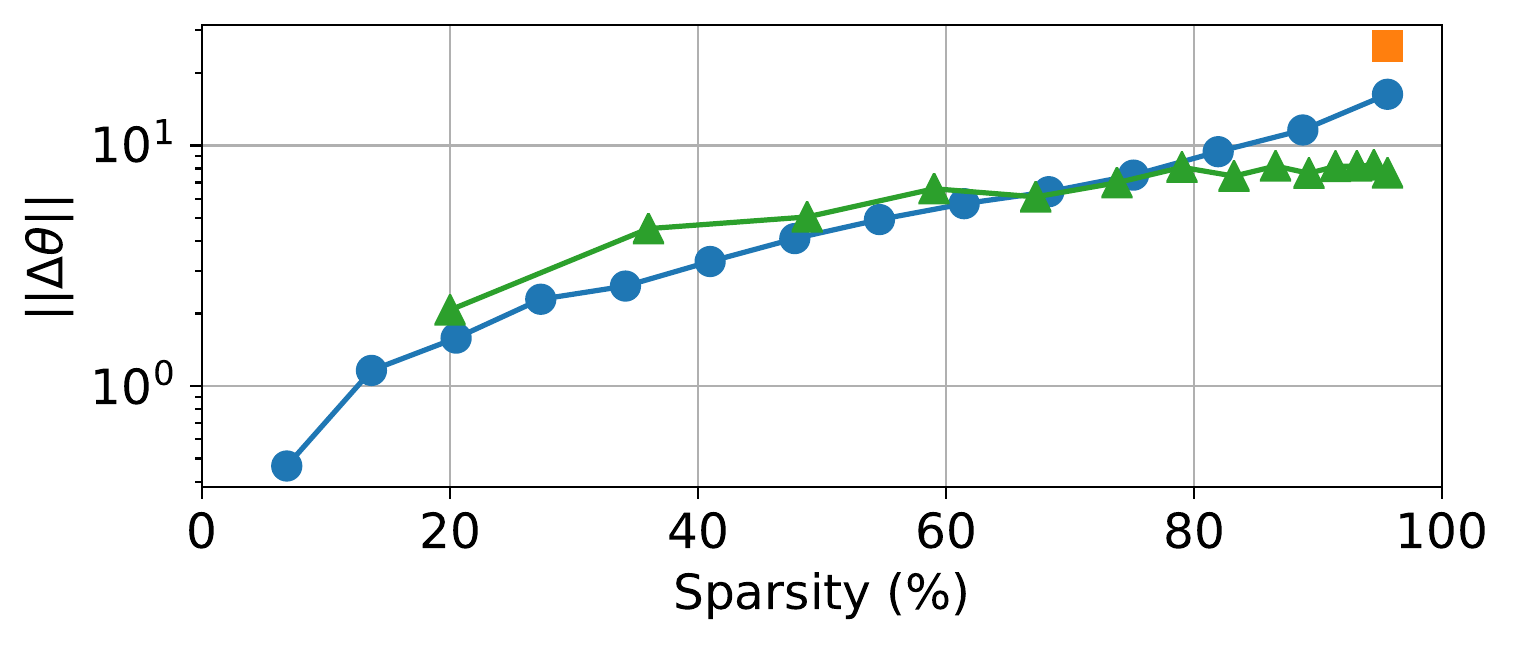}
    \includegraphics[width=0.32\textwidth]{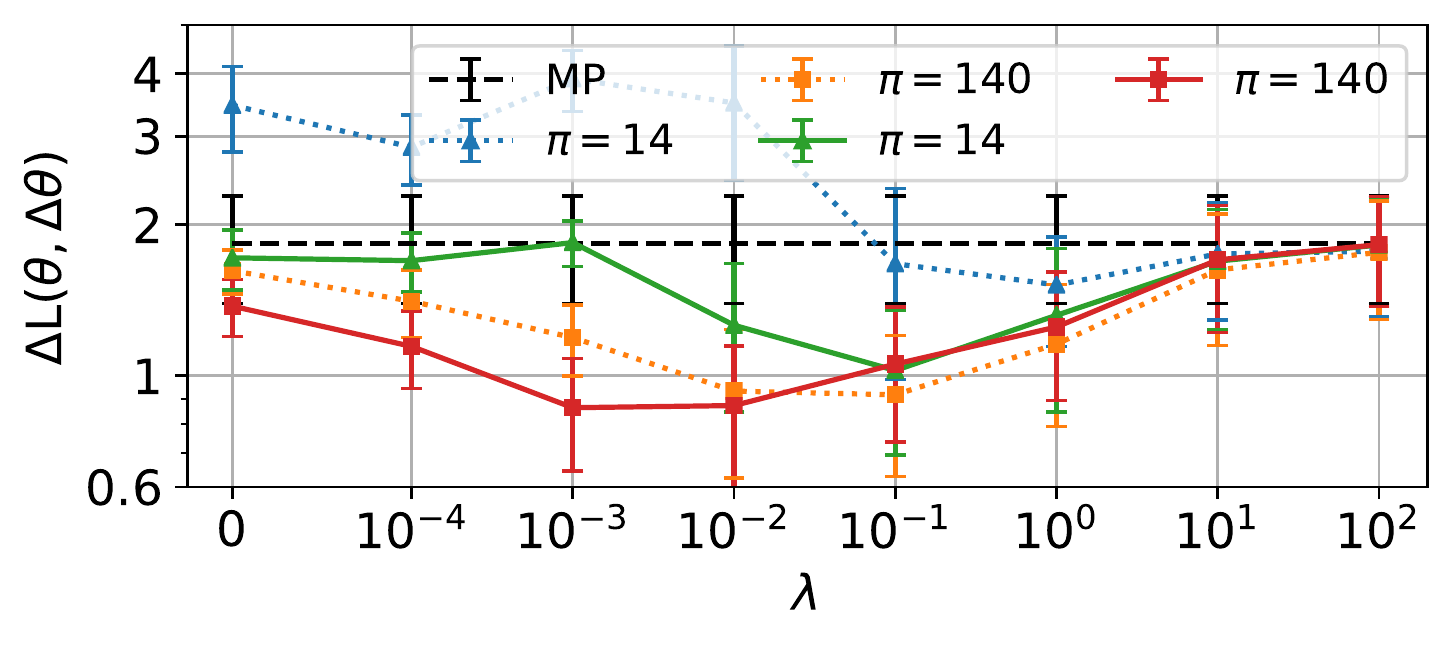}
    \caption{Linear vs exponential pruning steps using QM on VGG11. Left: $\Delta \mathcal L(\vtheta, \Delta\vtheta)$ as a function of the sparsity. We zoom on the end to better highlight the differences. The 14 pruning iterations are denoted by markers. Middle: $\Vert \Delta \vtheta \Vert_2$ at each iteration. Right: Same as Figure~\ref{fig:loss_lt}, but comparing exponential (solid) and linear (dotted) steps using $\pi \in \{14, 140 \}$, and a sparsity of 95.6\%. For a fixed $\pi$, we get smaller $\Vert \Delta \vtheta \Vert_2$ per pruning iteration when using exponential rather than linear steps, which results in a smaller $\Delta \mathcal L(\vtheta, \Delta\vtheta)$. There is a clear advantage for using exponential steps when the pruning budged is limited, i.e. when $\pi$ is small. This advantage disappears for larger values of $\pi$.}
    \label{fig:norms}
\end{figure}

Figure~\ref{fig:norms} compares the impact of $\Vert \Delta \vtheta \Vert_2$ and reports the training error gap when pruning iteratively VGG11 on CIFAR10, either linearly or exponentially. We also compare against one-shot pruning, as reference. Pruning with exponential steps maintains a more constant $\Vert \Delta \vtheta \Vert_2$ throughout the pruning procedure, which limits the maximum size of $\Vert \Delta \vtheta \Vert_2$, and thus enforces better the locality assumption. We hypothesize that this could be one of the reasons behind the success of the Lottery Ticket and Rewinding experiments \citep{Frankle2018TheLT, frankle2019lottery, Renda2020Comparing}.

\section{Performances after Fine-tuning}
\label{sec:after_ft}

We now fine-tune the pruned networks using the same hyper-parameters and number of epochs than for the original training. Table~\ref{tab:after_ft} contains the validation gap between the non-pruned networks and the pruned networks after fine-tuning for all the criteria. LM performs better than MP on both the MLP and VGG11 (0.5\% difference), but all criteria perform similarly on the PreActResNet18. These results are consistent with the observations of \citet{blalock2020state}. As reference, global random pruning resulted in validation error rate of $47.18 \pm 6.8$~\% for the MLP, and resulted in non-retrainable networks on CIFAR10 (with $90$~\% error rate).

\begin{table}[h!!]
	\centering
	\caption{Best validation error gap of the fine-tuned networks (lower is better), for different pruning criteria, across values of $\lambda$ and $\pi$. LM is better than MP on the MLP and VGG11. QM performs similarly or slightly worse than QM. All the methods have similar performance on the PreActResNet18.}
    \begin{tabular}{@{}lcccc@{}}
        \toprule
        \multirow{2}{*}{\bf Network} & \multicolumn{4}{c}{ \bf Gap of Validation Error (\%)} \\
        \cmidrule{2-5}
        & \bf MP & \bf OBD & \bf LM & \bf QM \\
        \midrule
        MLP on MNIST & 2.4 $\pm$ 0.3 & \phantom{-}2.0 $\pm$ 0.1 & \bf \phantom{-}1.9 $\pm$ 0.3 & \bf \phantom{-}1.9 $\pm$ 0.2 \\
        VGG11 on CIFAR10 & 0.2 $\pm$ 0.2 & -0.1 $\pm$ 0.2 & \bf -0.3 $\pm$ 0.1 & -0.1 $\pm$ 0.1 \\
        PreActResNet18 on CIFAR10 & 0.2 $\pm$ 0.2 & \phantom{-}0.2 $\pm$ 0.2 & \bf \phantom{-}0.1 $\pm$ 0.1 & \phantom{-}0.2 $\pm$ 0.2 \\
        \bottomrule
    \end{tabular}
    \label{tab:after_ft}
\end{table}

\subsection{Correlation between loss-preserving and performances after fine-tuning}

\begin{figure}[htb!]
    \centering
    \includegraphics[width=0.32\textwidth]{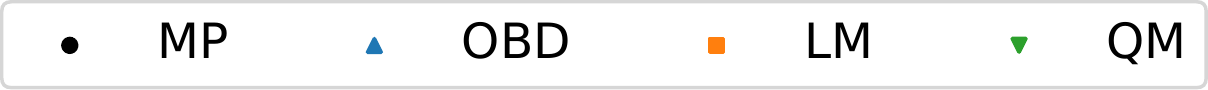}\\
    \includegraphics[height=0.32\textwidth]{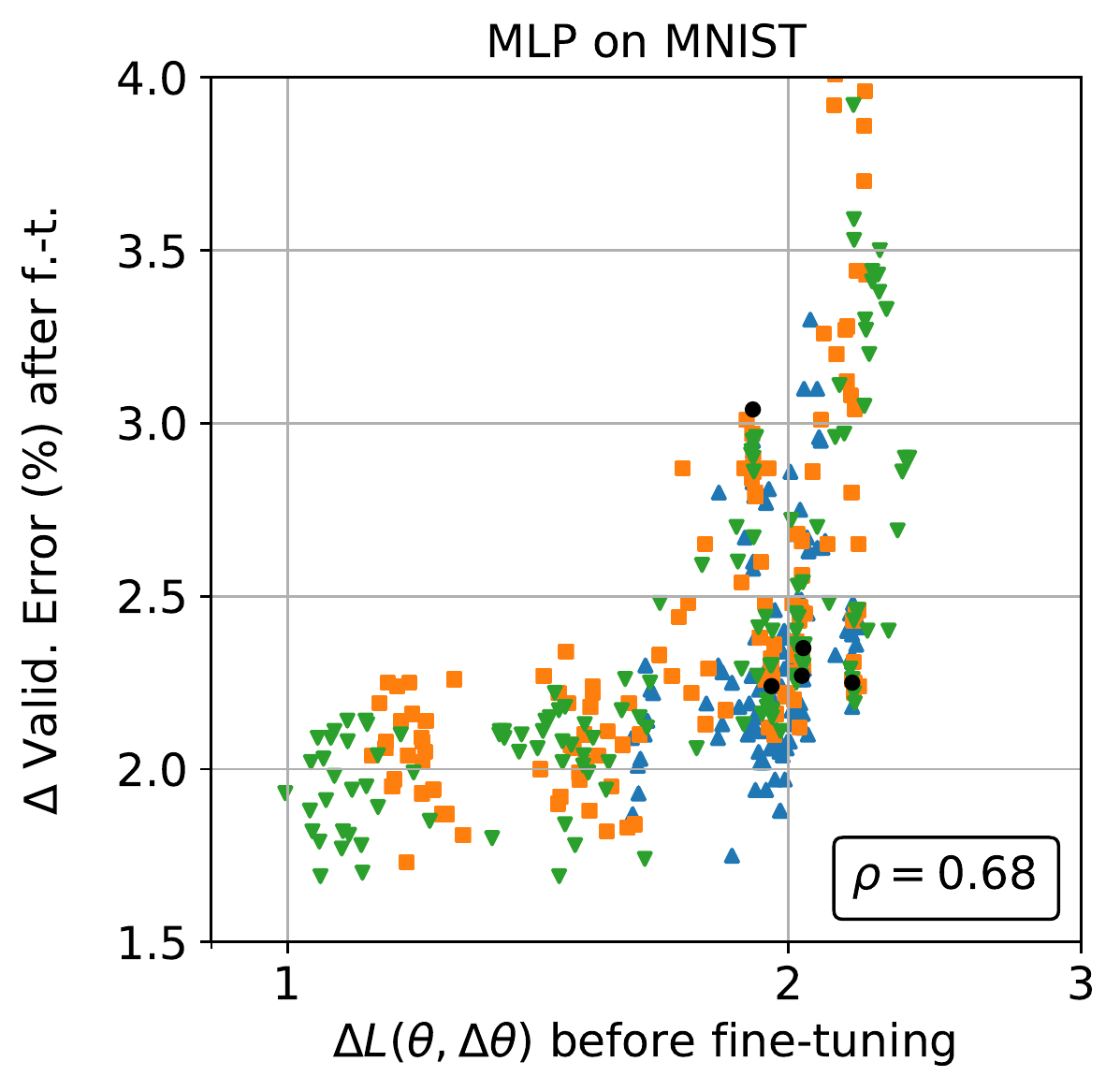}
    \includegraphics[height=0.32\textwidth]{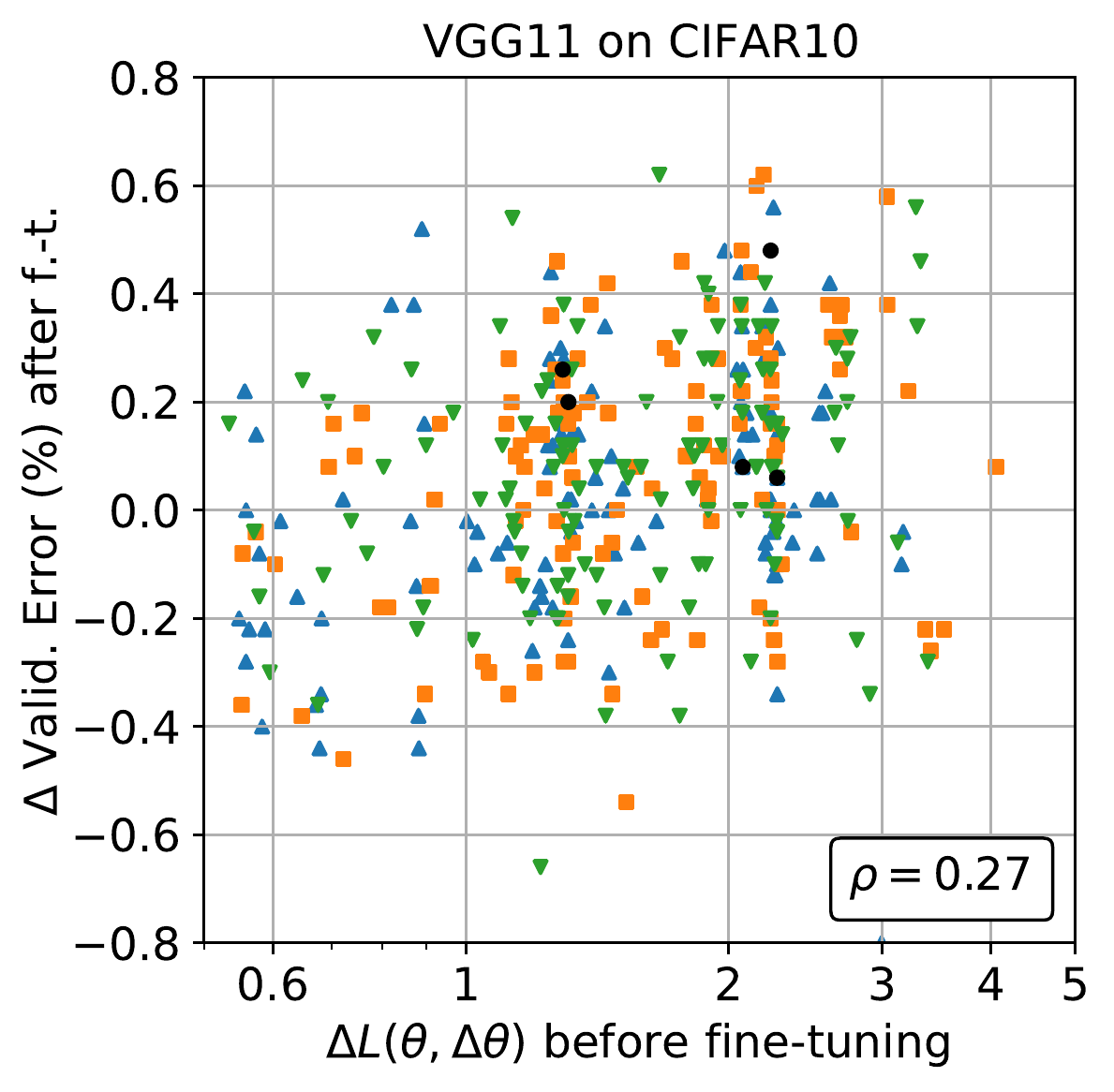}
    \includegraphics[height=0.32\textwidth]{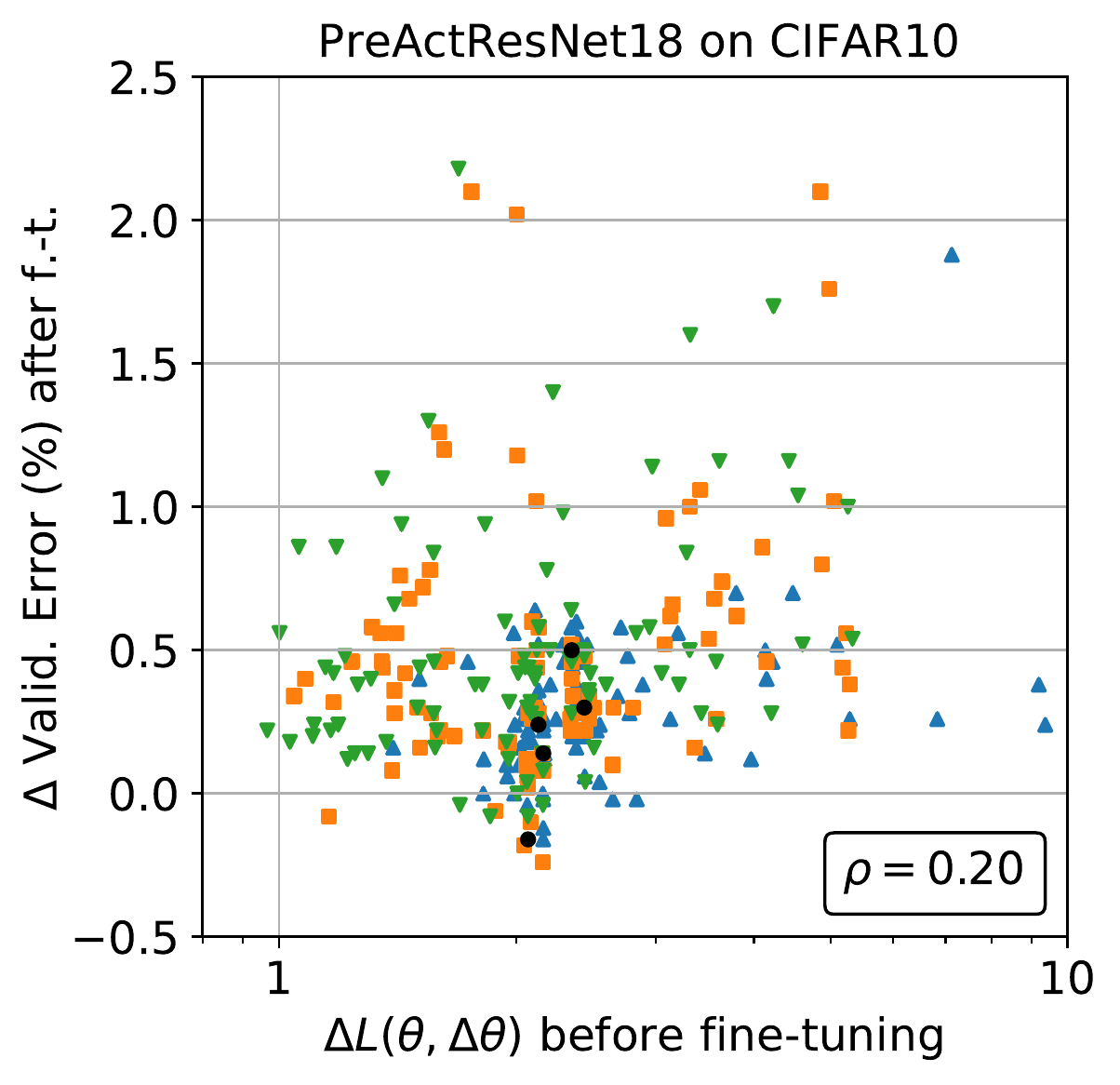}
    \caption{Scatter plot of the gap of validation error after fine-tuning as a function of $\Delta \mathcal L(\vtheta, \Delta\vtheta)$. Each point is one experiment (i.e. one one random seed, one $\pi$ and one $\lambda$). $\rho$ is the Spearman's rank correlation coefficient computed on all the data points. Except for the MLP on MNIST, there is only weak correlations between $\Delta \mathcal L(\vtheta, \Delta\vtheta)$ and the gap of validation after fine-tuning. Thus, the performance after pruning cannot be explained solely by the loss-preserving abilities of the pruning criteria, and other mechanisms might be at play.}
    \label{fig:scatter}
\end{figure}

An important observation is that the hyper-parameters $\lambda$ and $\pi$ that give the best performing criteria in terms of $\Delta\mathcal L(\vtheta, \Delta \vtheta)$ in Table~\ref{tab:loss} are not the same as the ones that give the best performing criteria after fine-tuning in Table~\ref{tab:after_ft}. We display in Figure~\ref{fig:scatter} scatter plots of all the experiments we ran, to show how well does loss-preserving correlate with performance after fine-tuning.

Quite surprisingly, although we are able to obtain networks with smaller $\Delta \mathcal L(\vtheta, \Delta\vtheta)$, and thus better performing networks right after pruning, the performances after fine-tuning do not correlate significantly with the gap. Except for the MLP on MNIST, whose Spearman's rank correlation coefficient is $\rho=0.67$, there is only weak correlations between $\Delta \mathcal L(\vtheta, \Delta\vtheta)$ and the validation error gap after fine-tuning ($\rho=0.27$ for VGG11 and $\rho=0.20$ for PreActResNet18). Figure~\ref{fig:scatter_loss} in Appendix contains the same scatter plots, but showing $\mathcal L(\vtheta \odot \mathbf m)$ after fine-tuning instead of the validation error gap, and similar trends can be observed. Figure~\ref{fig:scatter_pr}, also in Appendix, shows similar scatter plots, but for different sparsity levels on VGG11.

To verify that these observations are not due to a specific choice of fine-tuning hyper-parameters, we perform an hyper-parameter grid search and report similar results in Appendix~\ref{app:finetune}.
    
\subsection{Fine-tuning curves}

\begin{figure}[htb!]
    \centering
    \includegraphics[width=0.32\textwidth]{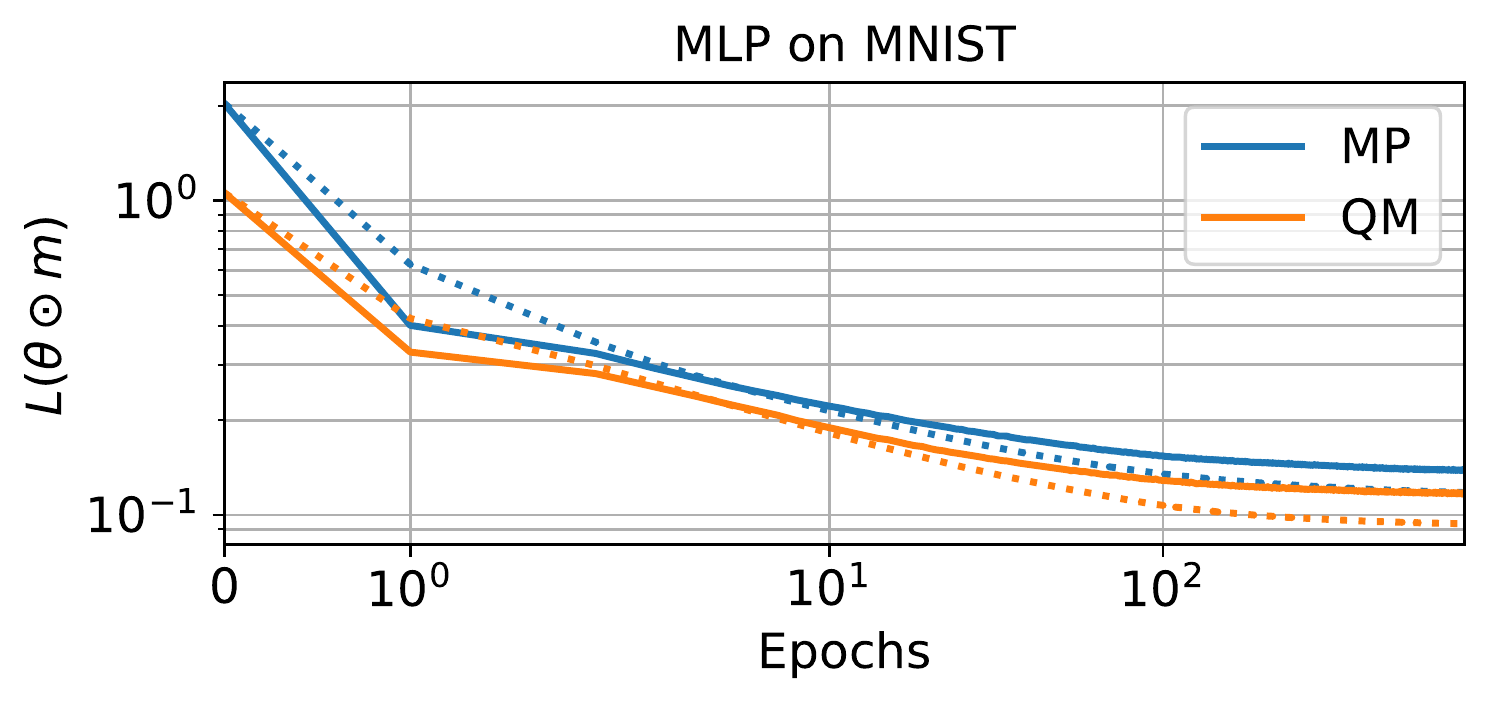}
    \includegraphics[width=0.32\textwidth]{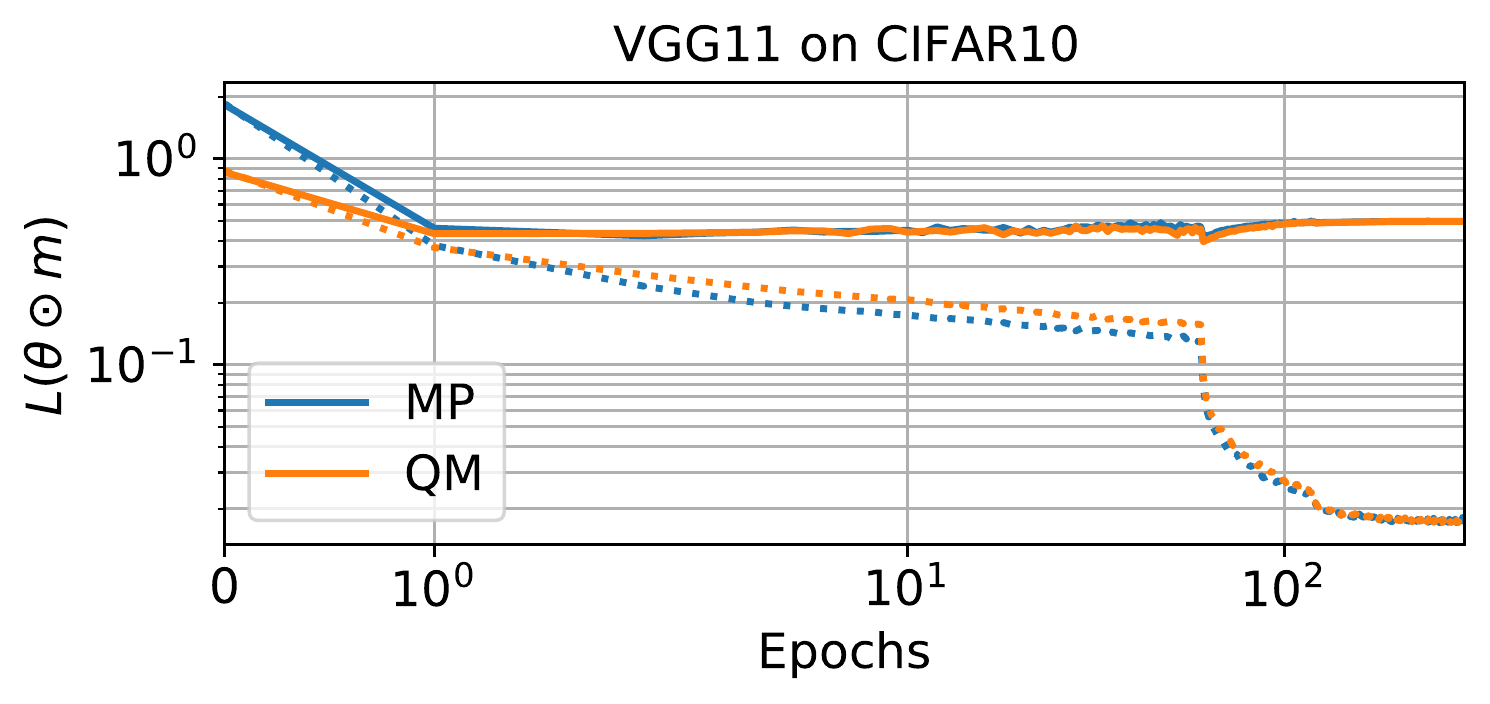}
    \includegraphics[width=0.32\textwidth]{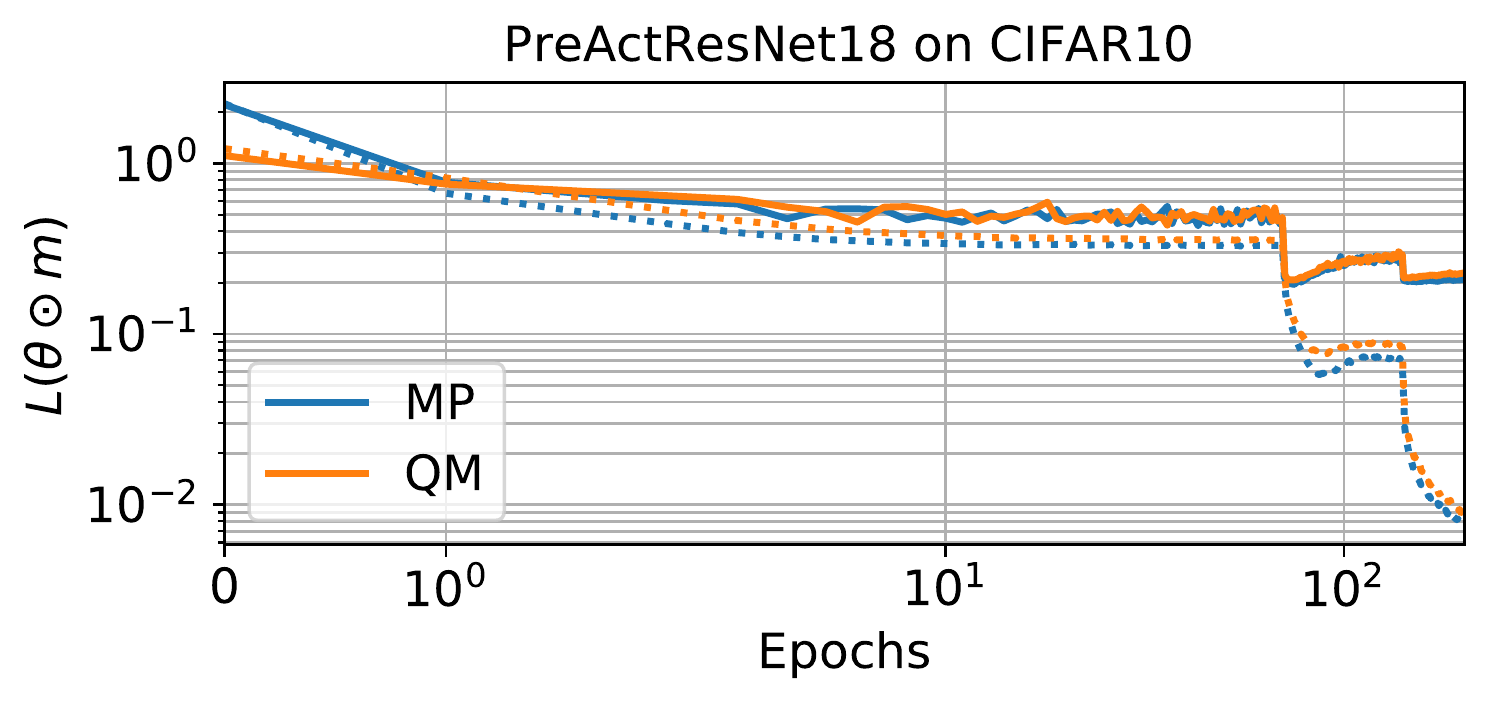}
    \caption{Fine-tuning losses (dotted is training, solid is validation) of networks pruned using MP and QM criteria. All the curves are the average over the 5 seeds. We do not show the standard deviation for clarity. Left: MLP, middle: VGG11 and right: PreActResNet18. Except for MNIST, the difference in loss right after pruning (i.e. at epoch 0) disappears after one epoch of fine-tuning.}
    \label{fig:retrain}
\end{figure}

To investigate whether one of the networks is suffering from optimization issues during fine-tuning, we show in Figure~\ref{fig:retrain} the fine-tuning curves of networks pruned using MP and our best QM criteria. We observe that, except for MNIST, the difference in training loss right after pruning disappears after only one epoch of fine-tuning, erasing the advantage of QM over MP.

\subsection{Discussion}

These results raise an important issue: minimizing $\Delta \mathcal L(\vtheta, \Delta\vtheta)$, no matter what model is used, might be used to design better pruning criteria, but it does not necessarily transfer to a better pruning method when fine-tuning is involved. The performance after pruning cannot be explained solely by the local loss-preserving abilities of the criteria, and other mechanisms might be at play. Thus, the fine-tuning should also be taken into account when designing pruning criteria.
For instance, \citet{lee2019signal} and \citet{Wang2020Picking} proposed different heuristics to take into account gradient propagation in the context of \emph{foresight} pruning, i.e. pruning untrained networks right after initialisation. \citet{Wang2020Picking} argues that minimizing $\Delta \mathcal L(\vtheta, \Delta\vtheta)$ in that context makes little sense, since the network is producing random predictions. Such methods should also be investigated in the context of fine-tuning.

\section{Conclusion}

In this paper, we revisited loss modelling for unstructured pruning. We showed that keeping the gradient term in the quadratic model allows to relax the convergence assumption behind OBS and OBD.
We also showed the importance of locality when using loss models for pruning: increasing the number of pruning iterations and constraining the step size are two improvements that produce better pruning criteria and that should be added to the recommendation list of \citet{blalock2020state}.
Moreover, we also showed that linear models perform similarly to quadratic diagonal models, at a lower computational cost.
Finally we observed that, even on our small-scale CIFAR10 experiments, the loss right after pruning does not correlate much with the performances after fine-tuning, suggesting that a better loss before fine-tuning is not solely responsible for the performances after fine-tuning. Thus, future research should focus on ways to model the actual effect of subsequent fine-tuning when designing pruning criteria.

\clearpage
\section*{Broader Impact}

Pruning methods allow to deploy neural networks that require less computational resources and storage. This may be beneficial in reducing the energy and environmental footprint required by a given system. They are also related to enabling the deployment of neural networks on embedded systems, that can be used for a multitude of applications, ranging from medical devices to weapon systems.

The consequences of failure of the system are the same as the original network. Pruning methods could leverage biases in the data, by maintaining good predictions only for the most represented classes. However, one can easily compare the performances between the original and the pruned model, to control whether such phenomenon is happening.

\begin{ack}
We thank Facebook for computational and financial resources. This research was also enabled in part by support provided by Calcul Qu\'ebec and Compute Canada, and Science Foundation Ireland (SFI) under Grant Number 12/RC/2289\_P2 and 16/SP/3804 (Insight Centre for Data Analytics). We also wish to thank Aristide Baratin for insightful discussions.
\end{ack}

\clearpage
\appendix
{\centering \Large\bf Appendix\par}

\section{Generalized Gauss-Newton}
\label{sec:GGN}

\subsection{Definition}

Having to compute $\mathbf H(\vtheta)$ is an obvious drawback of quadratic models, and thus a common first step is to approximate $\mathbf H(\vtheta)$ using the Generalized Gauss-Newton approximation \citep{schraudolph2002fast}:
\begin{align}
    \mathbf H(\vtheta) &= \underbrace{\frac{1}{N} \sum_{i=1}^{N} \frac{\partial f_{\boldsymbol{\theta}}\left(x_{i}\right)}{\partial\boldsymbol{\theta}}^{\top}\nabla_{u=f_{\boldsymbol{\theta}}\left(x_{i}\right)}^{2}\ell\left(u,t_{i}\right)\frac{\partial f_{\boldsymbol{\theta}}\left(x_{i}\right)}{\partial\boldsymbol{\theta}}}_{\mathbf G(\vtheta) \text{, the Generalized Gauss-Newton}} + \underbrace{\sum_{k}^K\frac{\partial\ell\left(u,t_{i}\right)}{\partial u_{k}}\Bigr|_{_{u=f_{\boldsymbol{\theta}}\left(x_{i}\right)}}\frac{\partial^{2}f_{\boldsymbol{\theta}}\left(x_{i}\right)_{k}}{\partial\boldsymbol{\theta}^{2}}}_{\approx 0}\\
    &\approx \mathbf G(\vtheta)
\end{align}
where K is the number of outputs of the network. $\mathbf G(\vtheta)$ has the advantage of being easier to compute and is also positive semi-definite by construction.

\subsection{Links with Kullback–Leibler Divergence}

\citet{pascanu2013revisiting} showed that, for networks $f_{\vtheta}$ that output probability distributions, $\mathbf G(\vtheta)$ is equal to the Fisher Information Matrix $\mathbf F(\vtheta)$, when the latter is approximated empirically. Moreover, $\mathbf F(\vtheta)$ is the Hessian of the Kullback-Leibler divergence $D_{KL}$:
\begin{align}
    D_{KL}(f_{\vtheta} \mid \mid f_{\vtheta + \Delta\vtheta}) = \frac{1}{2} \Delta\vtheta^T \mathbf F(\vtheta) \Delta\vtheta + \mathcal O(\Vert \Delta\vtheta \Vert^2_2)
\end{align}
Note that the first order term in the Taylor expansion of $D_{KL}$ is zero by construction. Interestingly, approximating $\mathbf H(\vtheta)$ with $\mathbf G(\vtheta)$ and neglecting the gradient term in OBD (Equation~\ref{eq:sens_OBD}) leads to minimizing $D_{KL}(f_{\vtheta} \mid \mid f_{\vtheta + \Delta\vtheta})$ rather than $\Delta\mathcal L(\vtheta, \Delta\vtheta)$ in Equation~\ref{eq:pruning_delta}, and we can highlight the difference between the two objectives:
\begin{align}
    \Delta\mathcal L(\vtheta, \Delta\vtheta) &= \left\vert \frac{1}{N}\sum_{i=1}^N \sum_{k=1}^K (t_i)_k \ln \left(\frac{f_{\vtheta}(x_i)_k}{f_{\vtheta + \Delta\vtheta}(x_i)_k}\right) \right\vert\\
    D_{KL}(f_{\vtheta} \mid \mid f_{\vtheta + \Delta\vtheta}) &= \frac{1}{N}\sum_{i=1}^N \sum_{k=1}^K f_{\vtheta}(x_i)_k \ln \left(\frac{f_{\vtheta}(x_i)_k}{f_{\vtheta + \Delta\vtheta}(x_i)_k}\right)
\end{align}
Since $D_{KL} \geq 0$, the only difference between the two objectives is whether they are evaluated using the distribution of the target $t$ or using the output distribution of the current network $f_{\vtheta}$.

\section{Details on the Experimental Setup}
\label{sec:setups}

\subsection{Setup}

\paragraph{Datasets} We use the MNIST dataset, and hold-out 10000 examples randomly sampled from the training set for validation. We also use CIFAR10 \citep{krizhevsky2009learning}, where the last 5000 examples of the training set are used for validation, and we apply standard data augmentation (random cropping and flipping, as in \citet{he2016identity}) during training phases.

\paragraph{Network Architectures}On MNIST, we use a MLP of dimensions 784-300-100-10, with Tanh activation functions. On CIFAR10, we use both: a VGG11 \citep{simonyan2014very}, equipped with ReLUs \citep{nair2010rectified}, but no Batch Normalisation~\citep{pmlr-v37-ioffe15}; and the PreActResNet18, which is the 18-layer pre-activation variant of residual networks \citep{he2016identity}. Except for the MLP, where \citet{glorot2010understanding} is used, the weights are initialized following \citet{he2015delving}, and the biases are initialized to 0.

\subsection{Experiments}

In all the experiments, the network is first trained for a fixed number of epochs, using early stopping on the validation set to select the best performing network. The hyper-parameters used for training are selected via grid search (before even considering pruning). Then we prune a large fraction of the parameters. For OBD, LM and QM, we randomly select, at each iteration of pruning, 1000 examples (10 mini-batches) from the training set to compute the gradients and second order terms of the models.\footnote{Using 1000 examples or the whole training set made no difference in our experiments. Using less examples started to degrade the performances, which concord with the observations of \citet{lee2018snip}}. Finally, we retrain the network using exactly the same hyper-parameters as for the initial training.

\paragraph{MLP on MNIST} We train the network for 400 epochs, using SGD with learning rate of 0.01, momentum factor of 0.9, l2 regularisation of 0.0005 and a mini-batch size of 100. We prune 98.85\% of the parameters.

\paragraph{VGG11 on CIFAR10} We train the network for 300 epoch, using SGD with a learning rate of 0.01, momentum factor of 0.9, a l2 regularisation of 0.0005 and a mini-batch size of 100. The learning rate is divided by 10 every 60 epochs. We prune 95.6\% of the parameters.

\paragraph{PreActResNet18 on CIFAR10}
We train the network for 200 epochs, using SGD with a learning rate of 0.1, momentum factor of 0.9, a l2 regularisation of 0.0005 and a mini-batch size of 100. The learning rate is divided by 10 every 70 epochs. We prune 95.6\% of the parameters.

\section{Supplementary Results}

\subsection{Performances before Fine-tuning}
\label{app:before}

\paragraph{Validation error Table} Table~\ref{tab:error} is the same as Table~\ref{tab:loss}, but containing the best validation error gap before/after pruning instead of $\Delta \mathcal L(\vtheta, \Delta\vtheta)$. We can observe a similar trend as in Table~\ref{tab:loss}: LM and QM give better performances than MP, and OBD performs poorly, since the convergence assumption is not respected.

\begin{table}[h!!]
	\centering
	\caption{Best validation error gap before/after pruning for different networks and pruning criteria.}
    \begin{tabular}{@{}lcccc@{}}
        \toprule
        \multirow{2}{*}{\bf Network} & \multicolumn{4}{c}{ \bf Gap of Validation Error (\%)} \\
        \cmidrule{2-5}
        & \bf MP & \bf OBD & \bf LM & \bf QM \\
        \midrule
        MLP on MNIST & 72.09 $\pm$ 3.72 & 64.89 $\pm$ 5.74 & \bf 16.35 $\pm$ 0.77 & \bf 15.22 $\pm$ 0.62 \\
        VGG11 on CIFAR10 & 56.19 $\pm$ 17.9 & 18.84 $\pm$ 5.54  & \bf \phantom{0}5.89 $\pm$ 1.52 & \bf \phantom{0}5.92 $\pm$ 2.14 \\
        PreActResNet18 on CIFAR10 & 74.13 $\pm$ 4.59 & 49.08 $\pm$ 8.18 & 26.79 $\pm$ 8.61 & \bf 21.48 $\pm$ 5.96 \\
        \bottomrule
    \end{tabular}
    \label{tab:error}
\end{table}

\begin{figure}[htb!]
    \centering
    \includegraphics[width=0.32\textwidth]{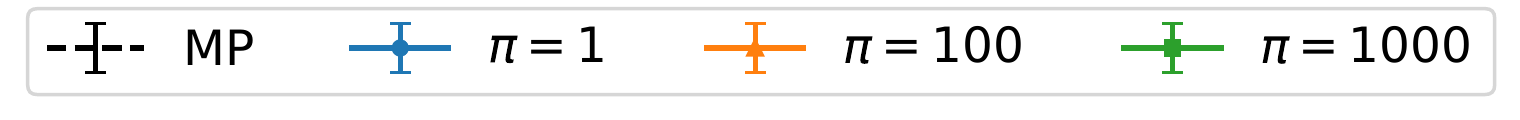}
    \begin{subfigure}[b]{\textwidth}
        \centering
        \includegraphics[width=0.32\textwidth]{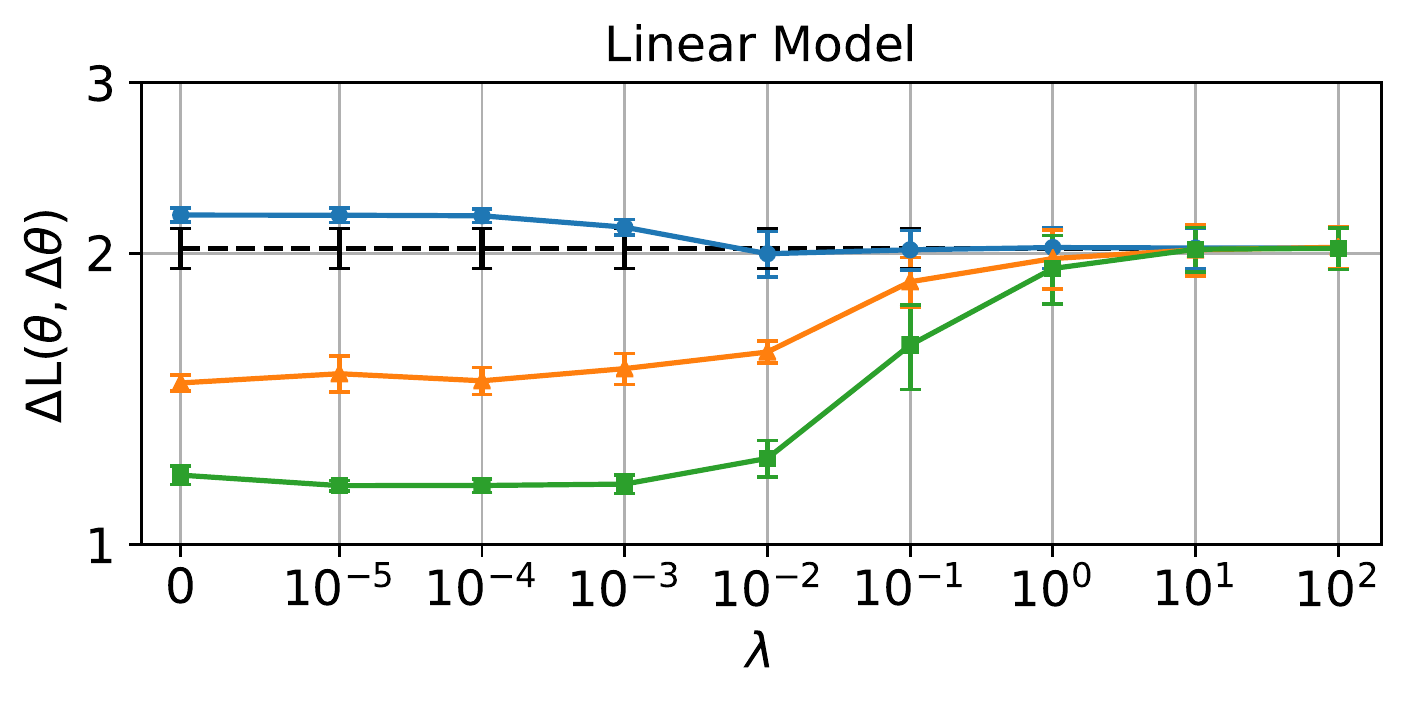}
        \includegraphics[width=0.32\textwidth]{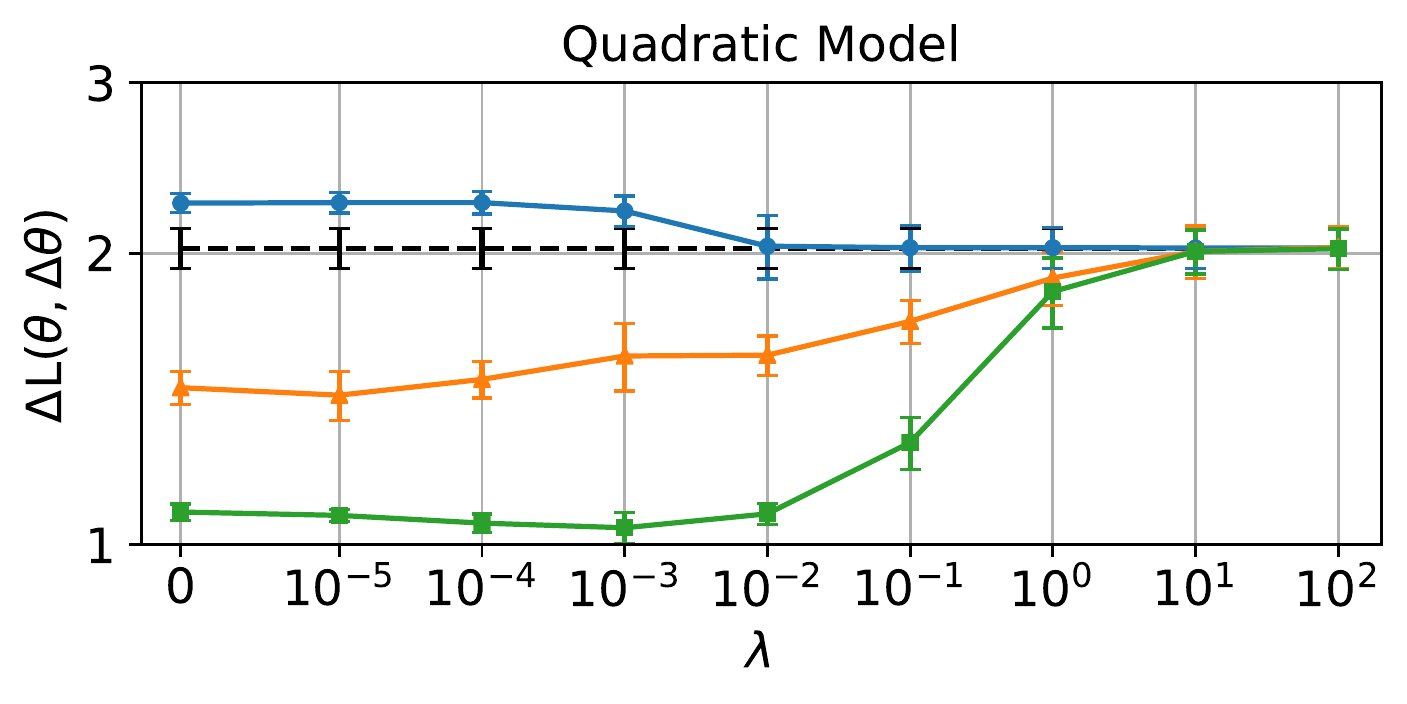}
        \includegraphics[width=0.32\textwidth]{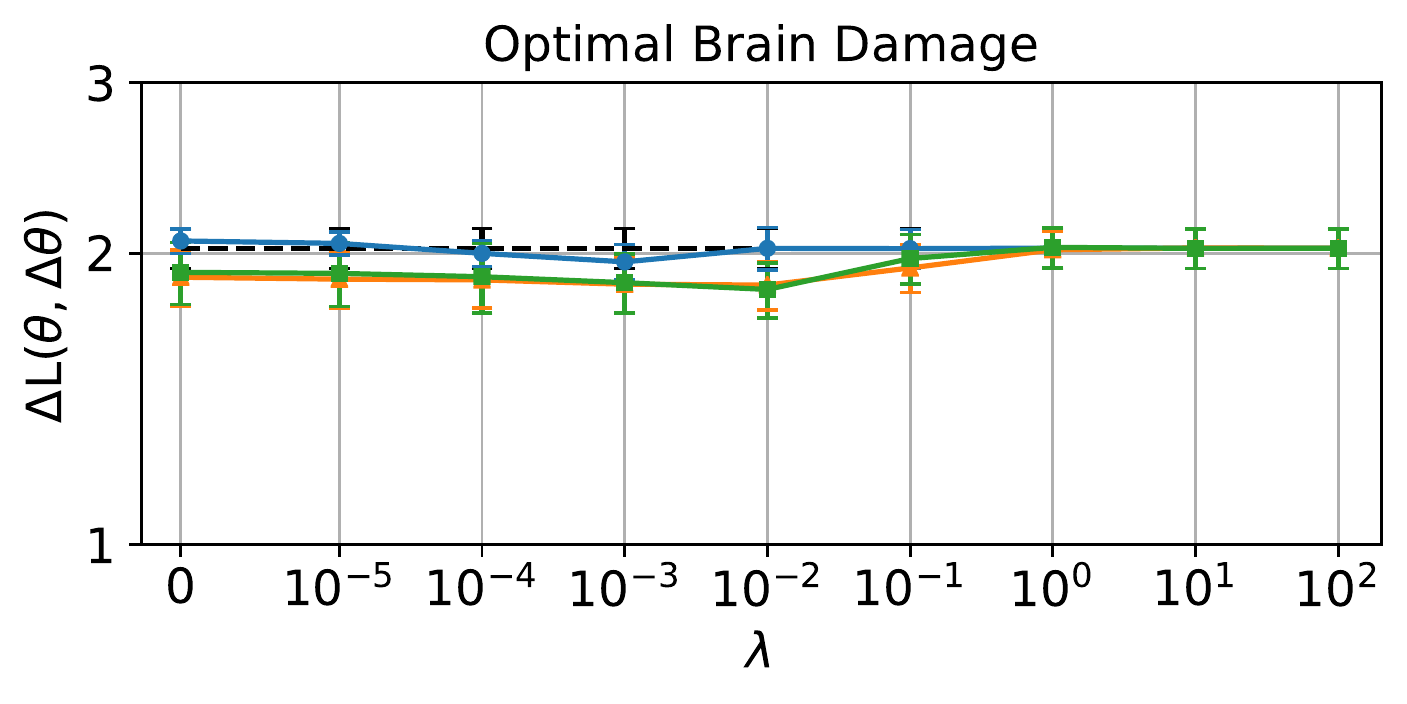}
        \label{fig:loss_eq_mlp}
        \caption{MLP on MNIST with 98.8\% sparsity.}
    \end{subfigure}\\
    \begin{subfigure}[b]{\textwidth}
        \centering
        \includegraphics[width=0.32\textwidth]{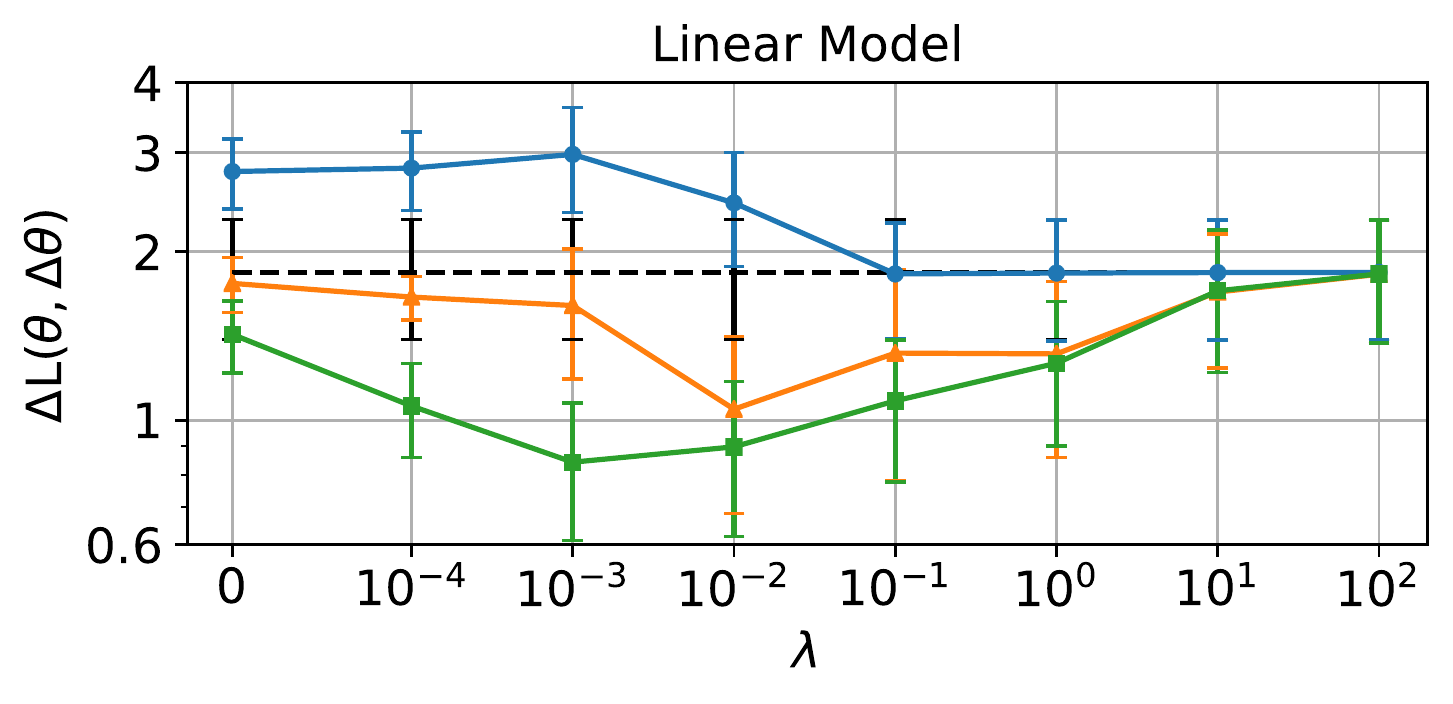}
        \includegraphics[width=0.32\textwidth]{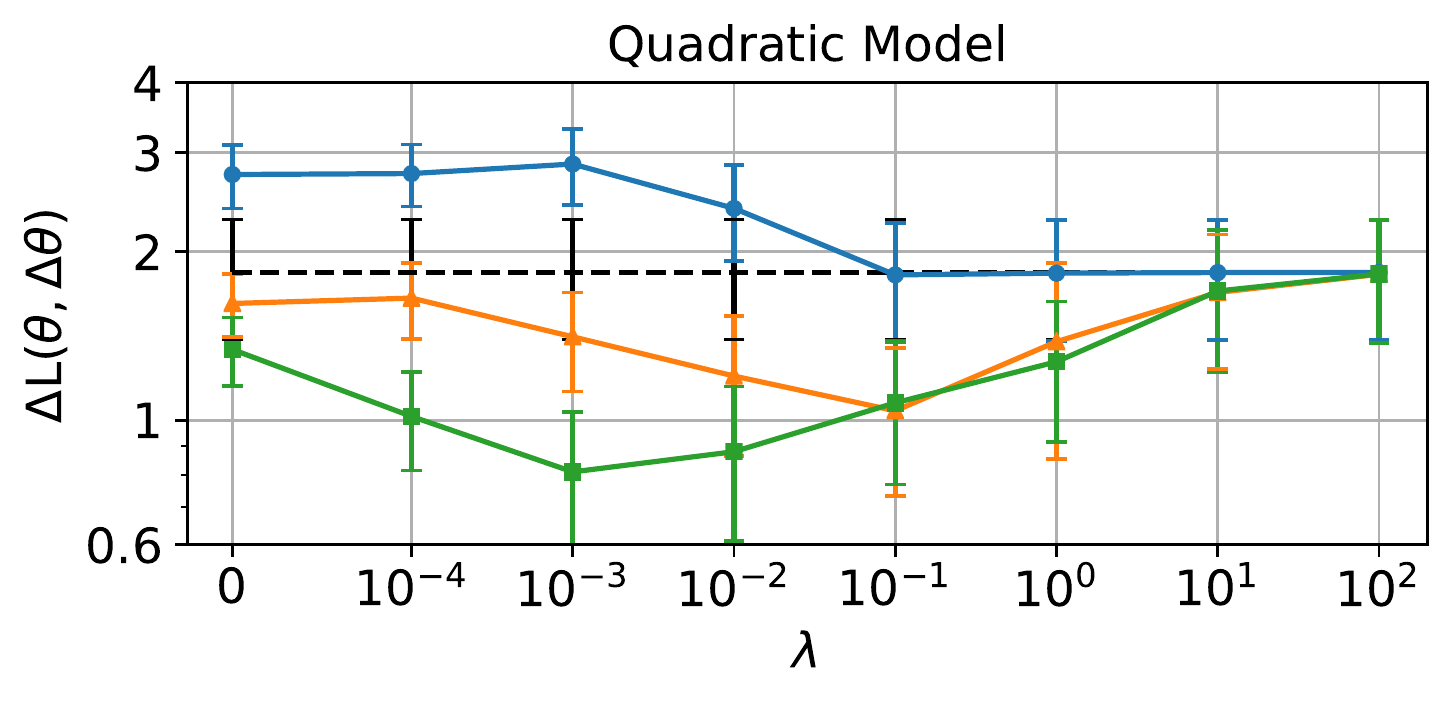}
        \includegraphics[width=0.32\textwidth]{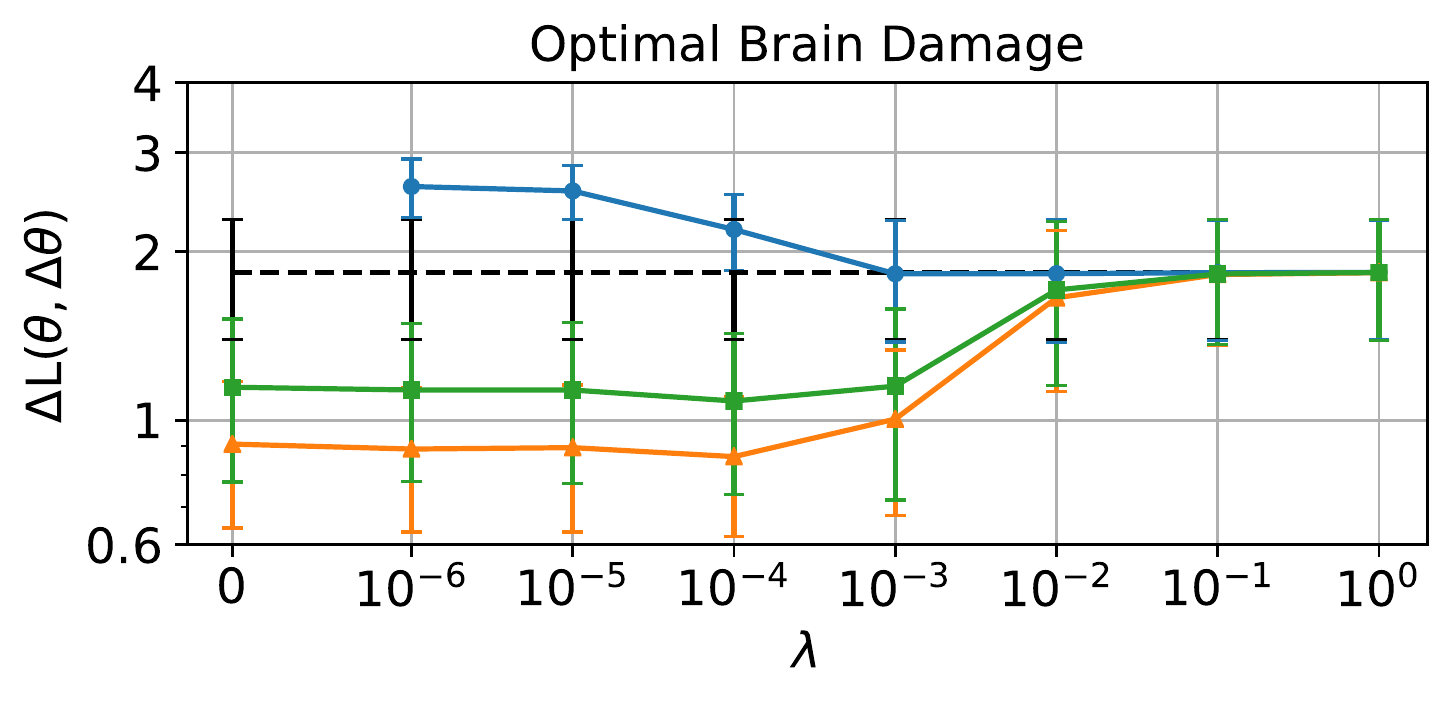}
        \label{fig:loss_eq_vgg}
        \caption{VGG11 on CIFAR10 with 95.6\% sparsity.}
    \end{subfigure}\\
    \begin{subfigure}[b]{\textwidth}
        \centering
        \includegraphics[width=0.32\textwidth]{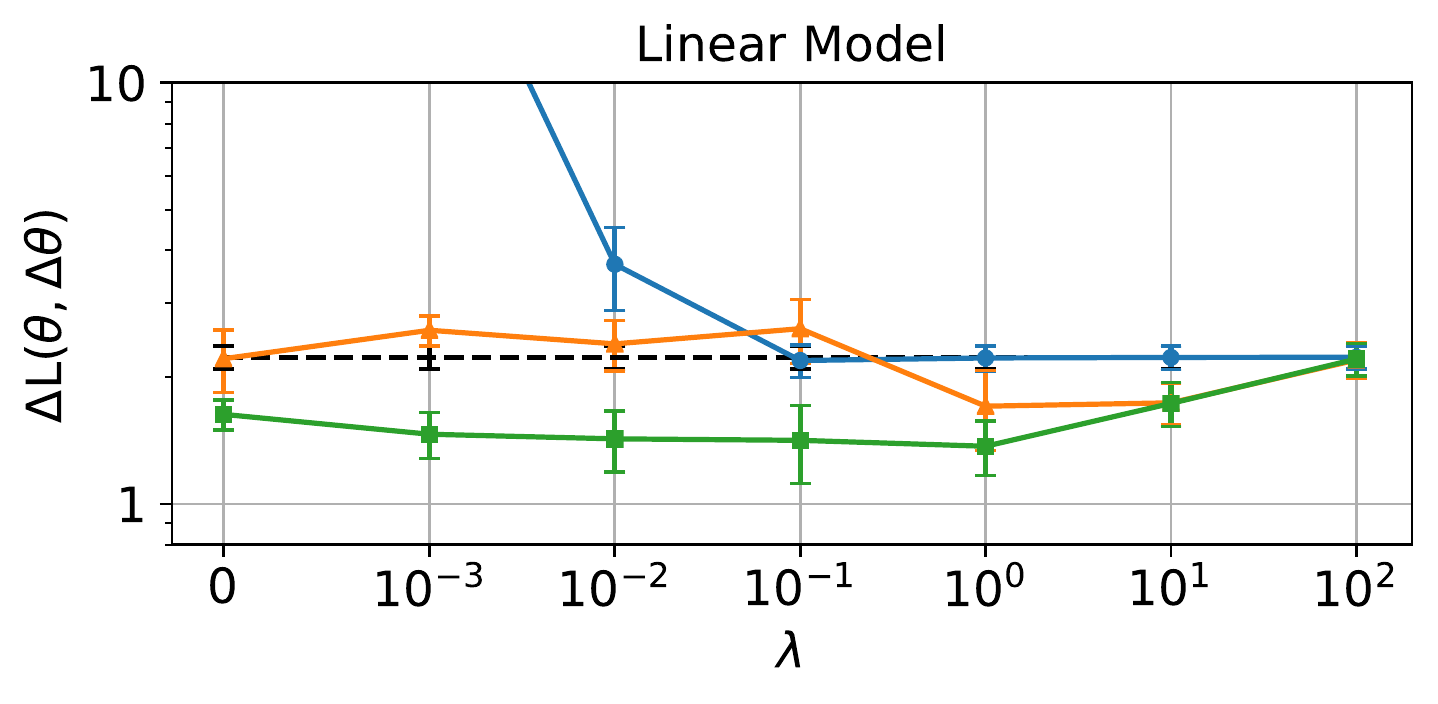}
        \includegraphics[width=0.32\textwidth]{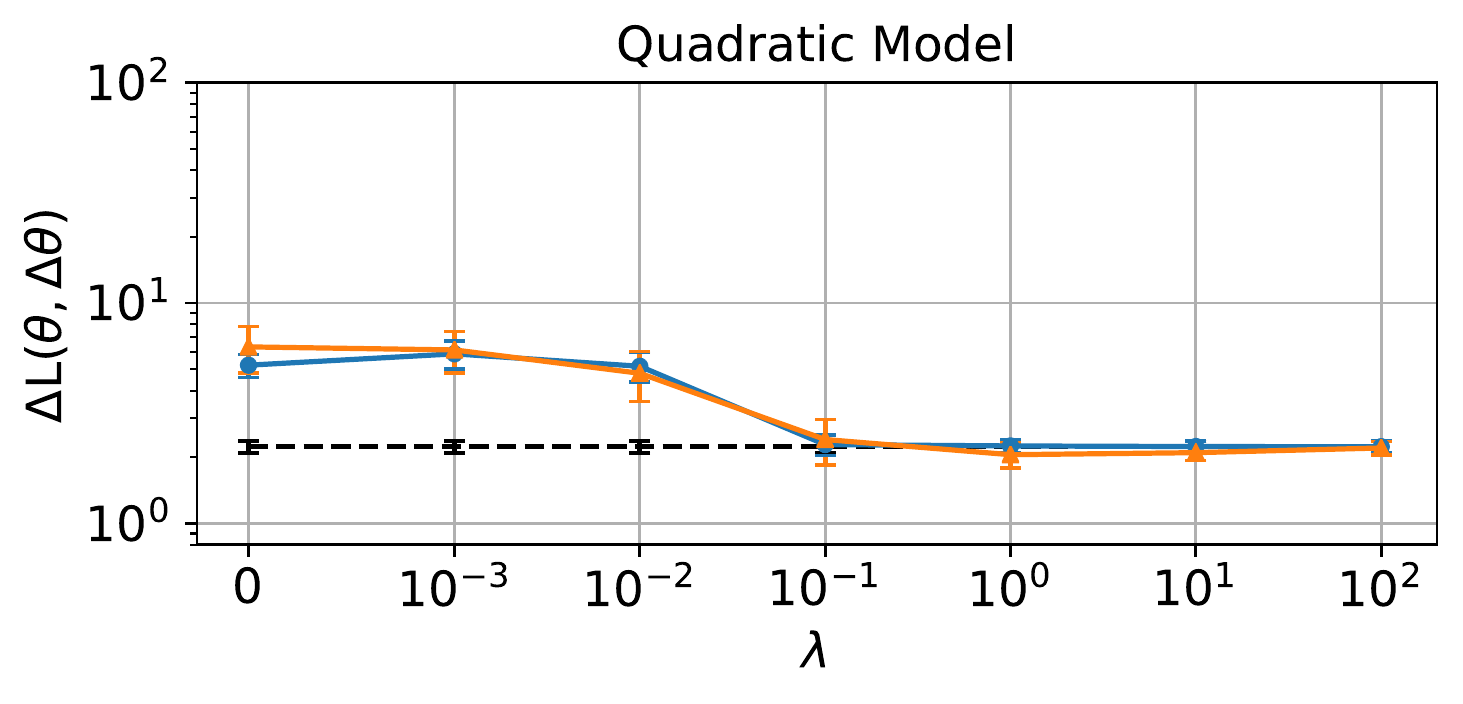}
        \includegraphics[width=0.32\textwidth]{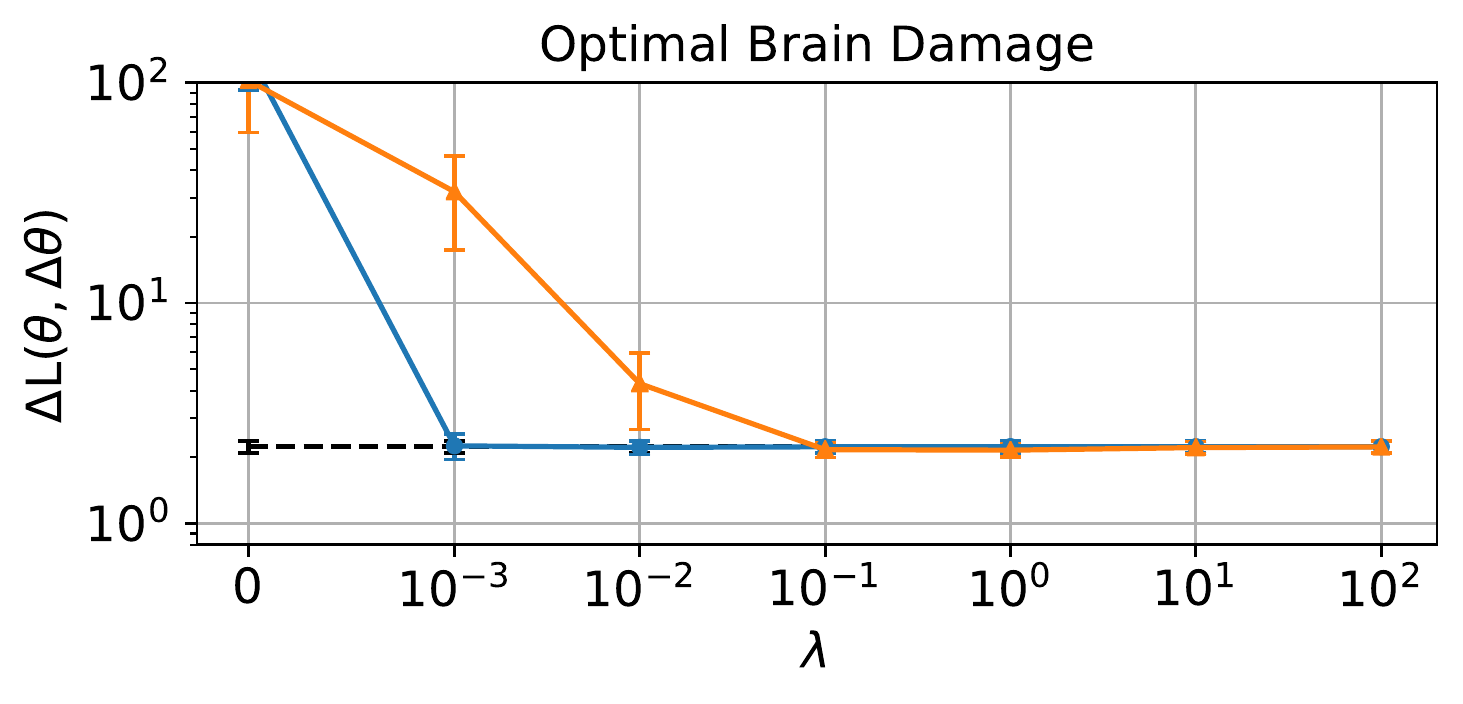}
        \label{fig:loss_eq_resnet}
        \caption{PreActResNet18 on CIFAR10 with 95.6\% sparsity.}
    \end{subfigure}
    \caption{Same as Figure~\ref{fig:loss_lt}, but using equally spaced pruning steps. Note the difference in number of pruning iterations.}
    \label{fig:loss_eq}
\end{figure}

\paragraph{Linear step size} Figure~\ref{fig:loss_eq} contains the same experiments than Figure~\ref{fig:loss_eq}, but using a linear step size rather than exponential. There is a drastic difference in performances: One need roughly 10x more iterations with the linear step size to reach the training gap of the exponential.

\subsection{Performances after Fine-tuning}
\label{app:finetune}

\begin{figure}[htb]
    \centering
    \includegraphics[width=0.32\textwidth]{figures/oneshot/legend_140.pdf}
    \begin{subfigure}[b]{\textwidth}
        \centering
        \includegraphics[width=0.32\textwidth]{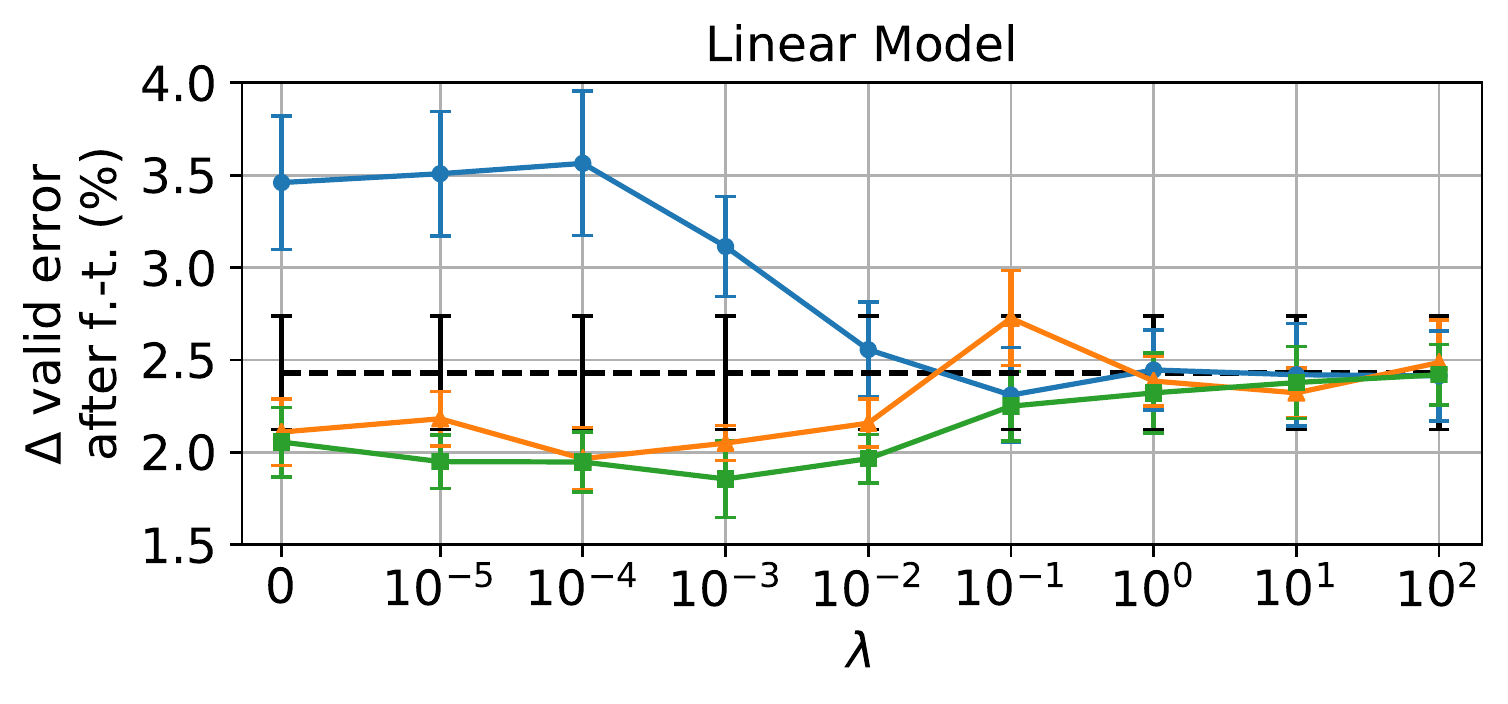}
        \includegraphics[width=0.32\textwidth]{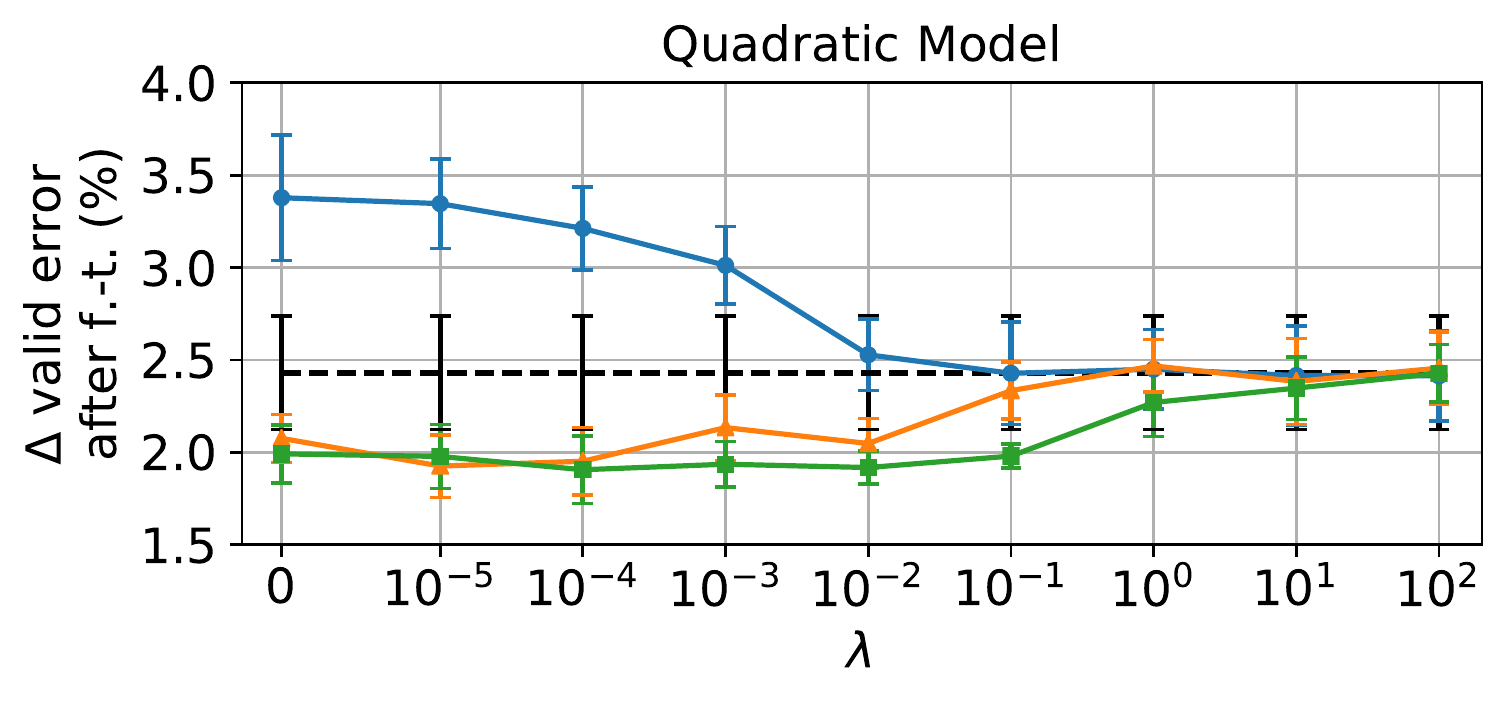}
        \includegraphics[width=0.32\textwidth]{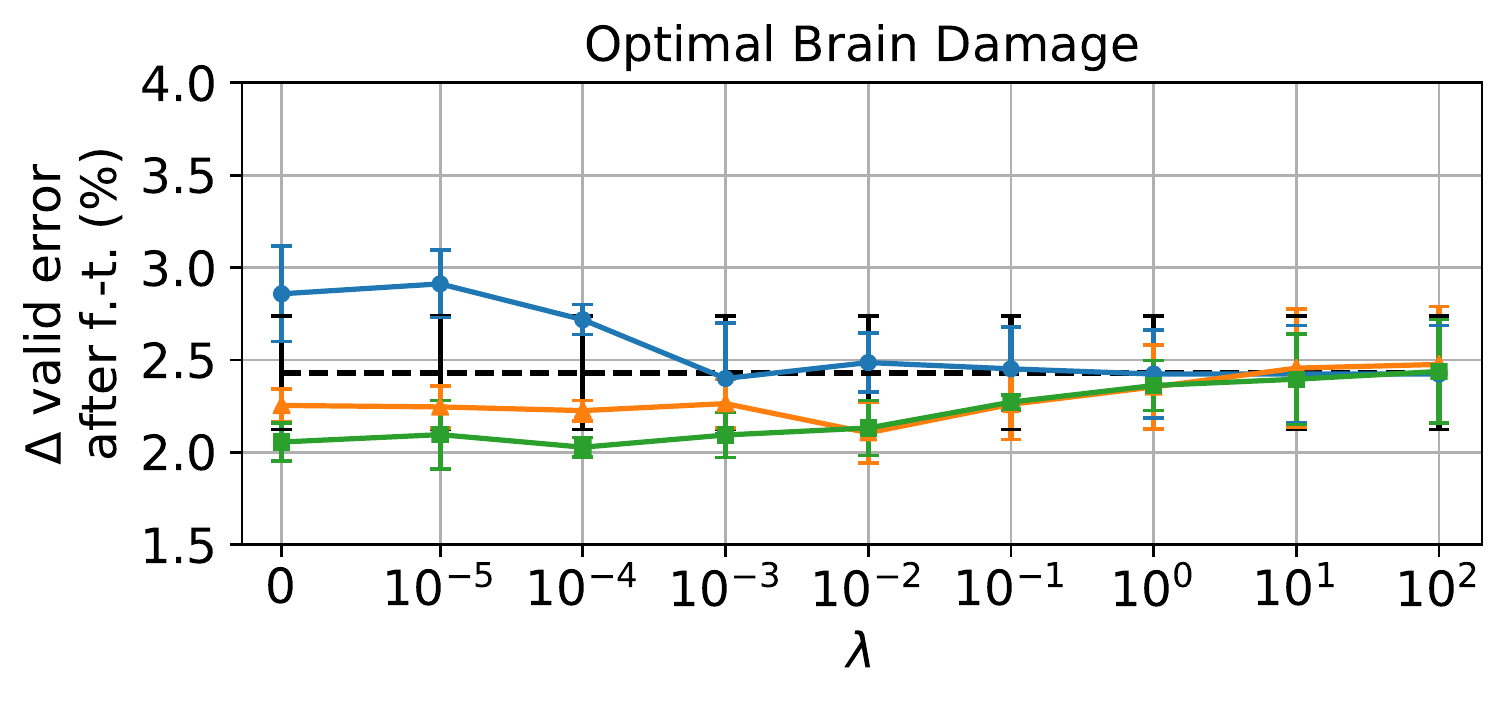}
        \caption{MLP on MNIST with 98.8\% sparsity.}
    \end{subfigure}\\
    \begin{subfigure}[b]{\textwidth}
        \centering
        \includegraphics[width=0.32\textwidth]{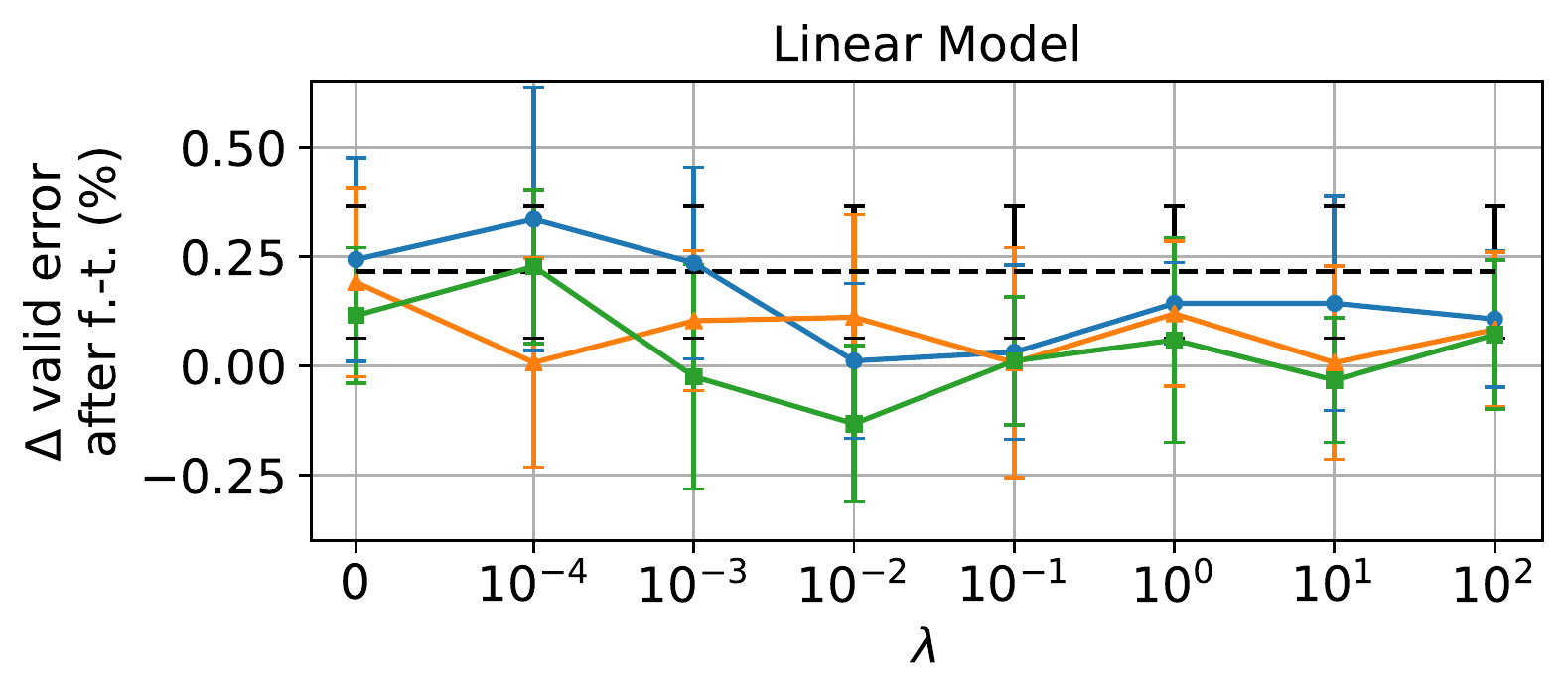}
        \includegraphics[width=0.32\textwidth]{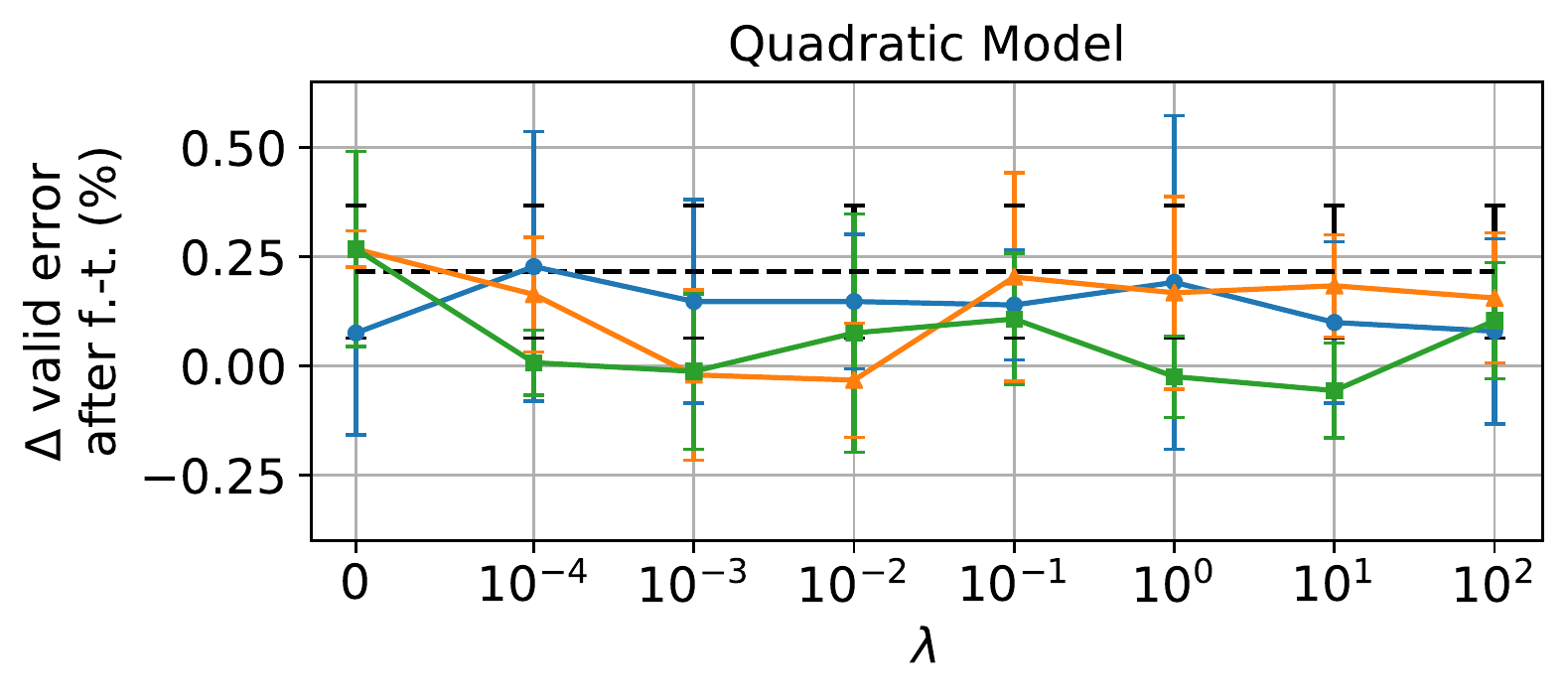}
        \includegraphics[width=0.32\textwidth]{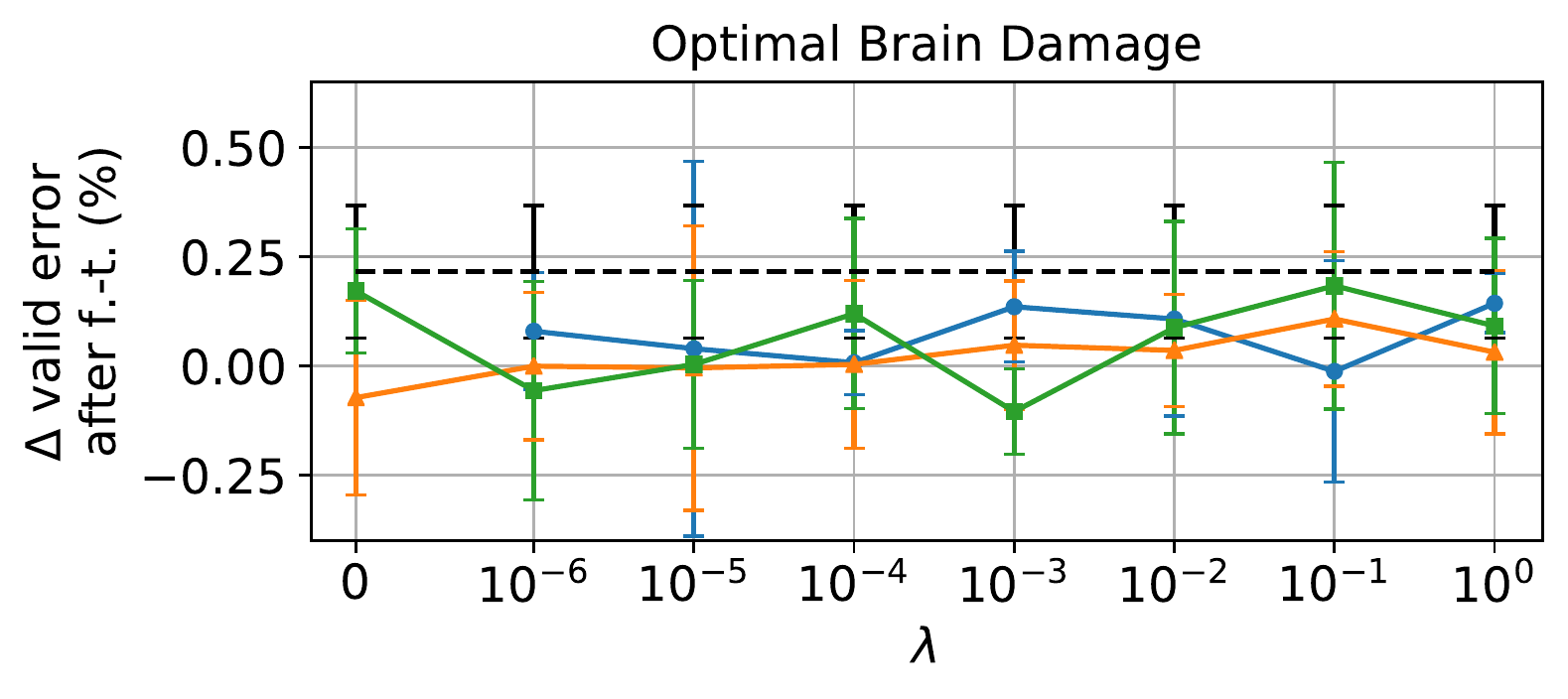}
        \caption{VGG11 on CIFAR10 with 95.6\% sparsity.}
    \end{subfigure}\\
    \begin{subfigure}[b]{\textwidth}
        \centering
        \includegraphics[width=0.32\textwidth]{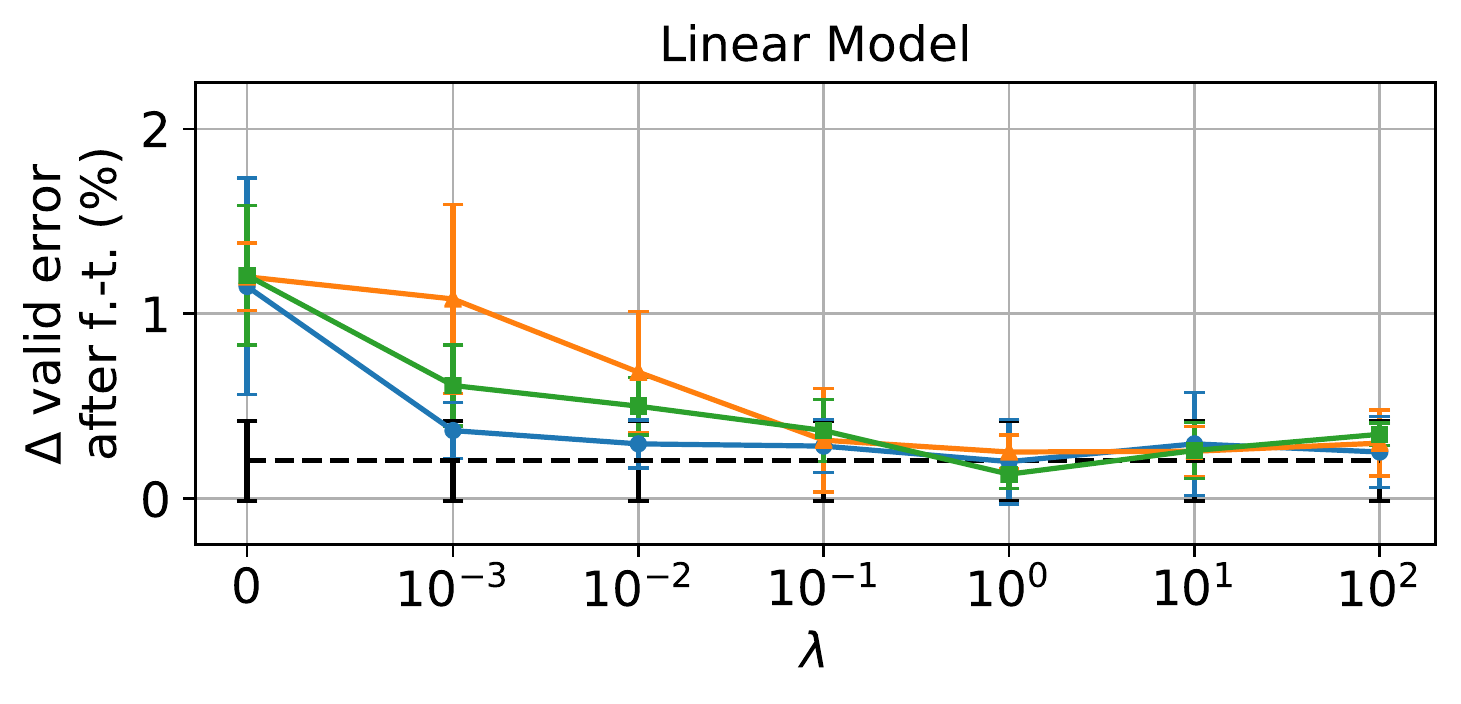}
        \includegraphics[width=0.32\textwidth]{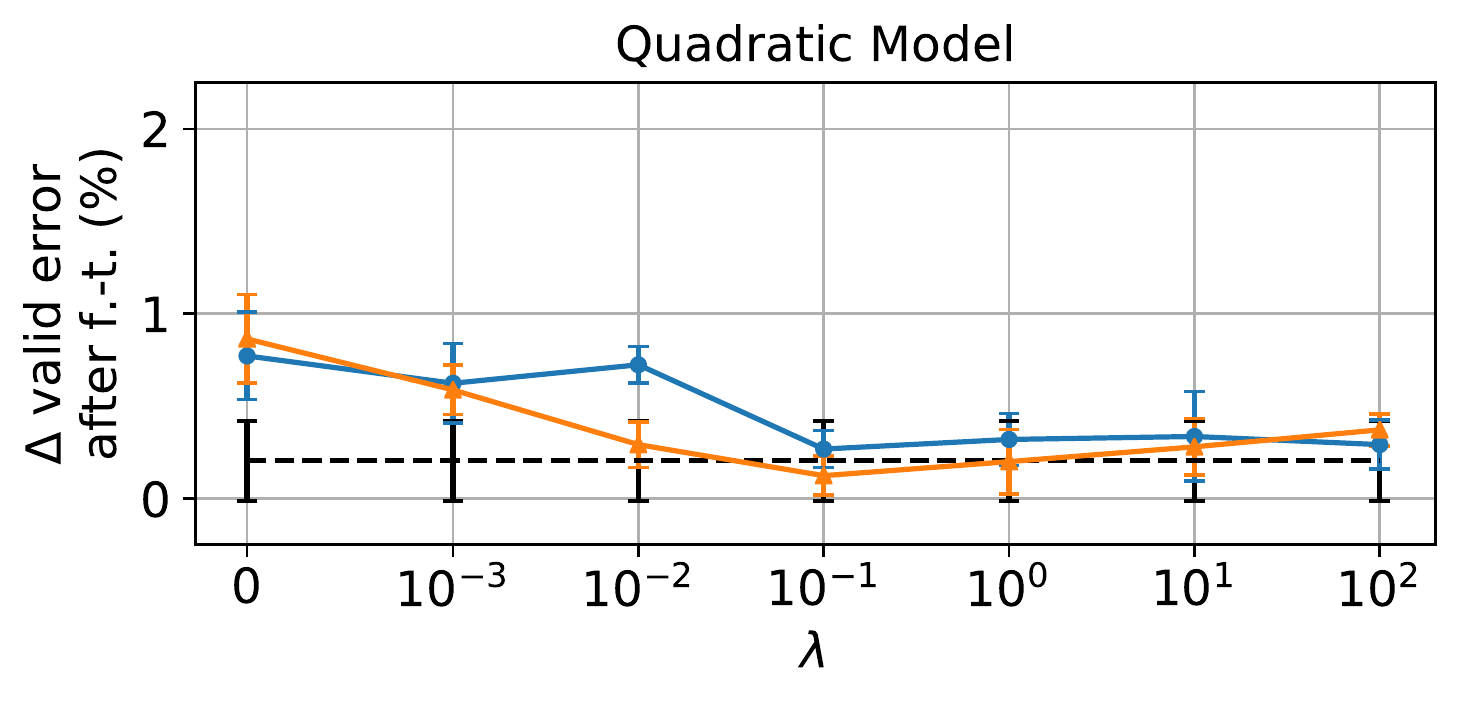}
        \includegraphics[width=0.32\textwidth]{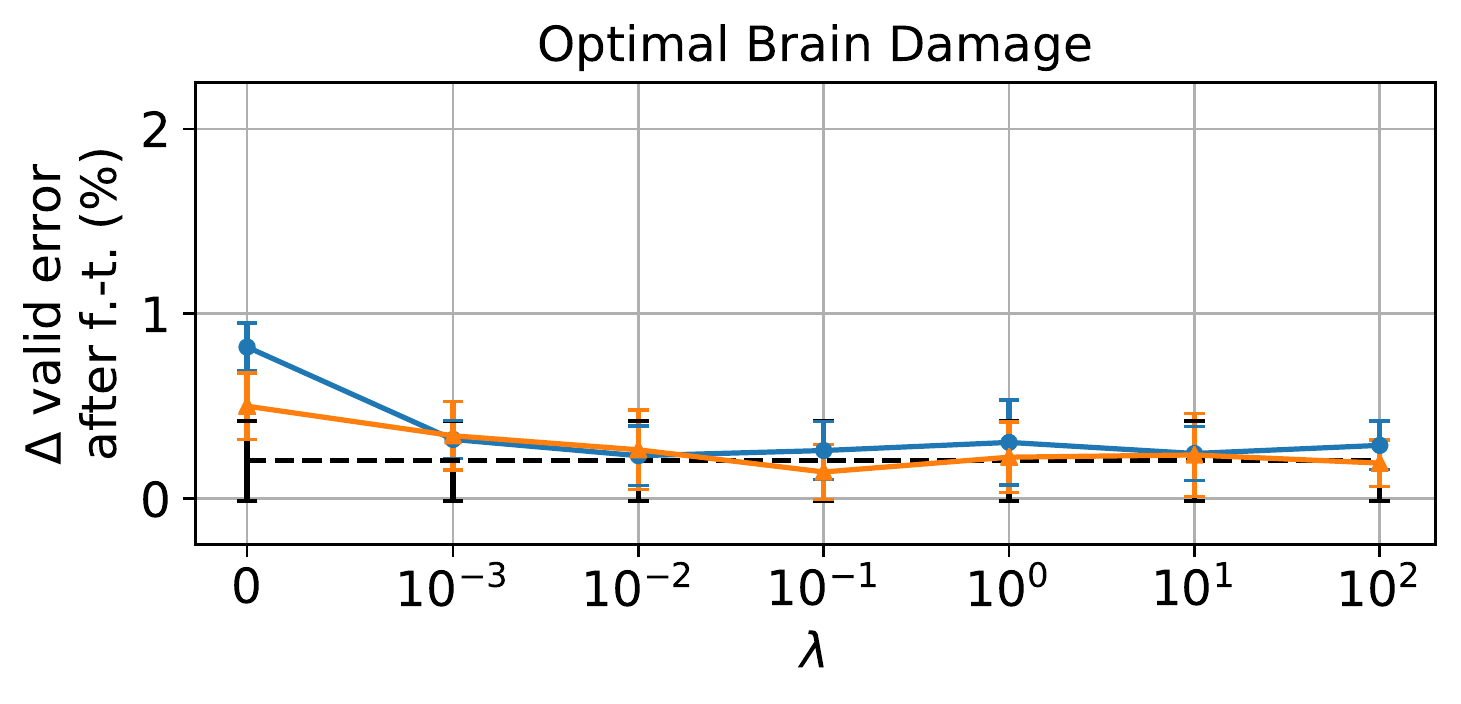}
        \caption{PreActResNet18 on CIFAR10 with 95.6\% sparsity.}
    \end{subfigure}
    \caption{Same as Figure~\ref{fig:loss_lt}, but displaying the validation error gap after fine-tuning.}
    \label{fig:val_lt}
\end{figure}

\begin{figure}[htb!]
    \centering
    \includegraphics[width=0.32\textwidth]{figures/oneshot/legend_1000.pdf}
    \begin{subfigure}[b]{\textwidth}
        \centering
        \includegraphics[width=0.32\textwidth]{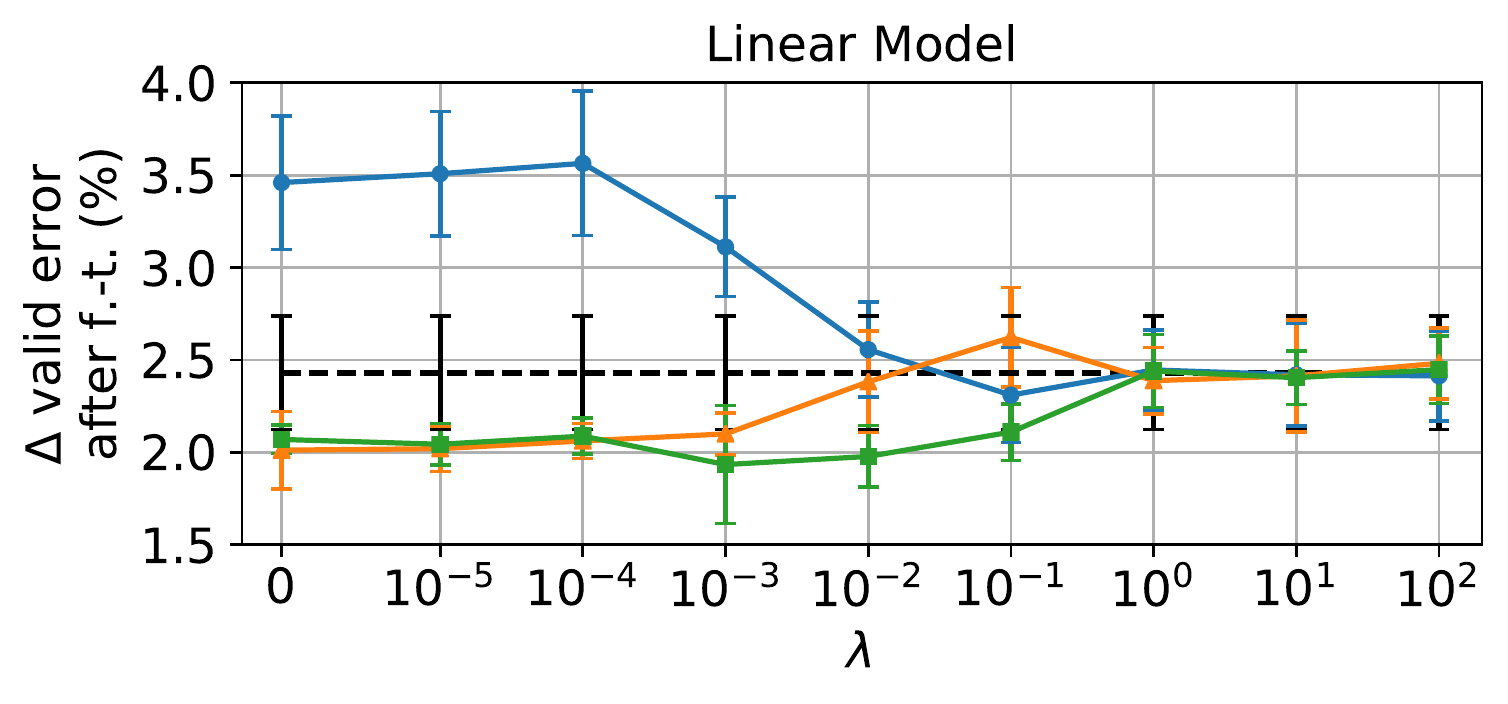}
        \includegraphics[width=0.32\textwidth]{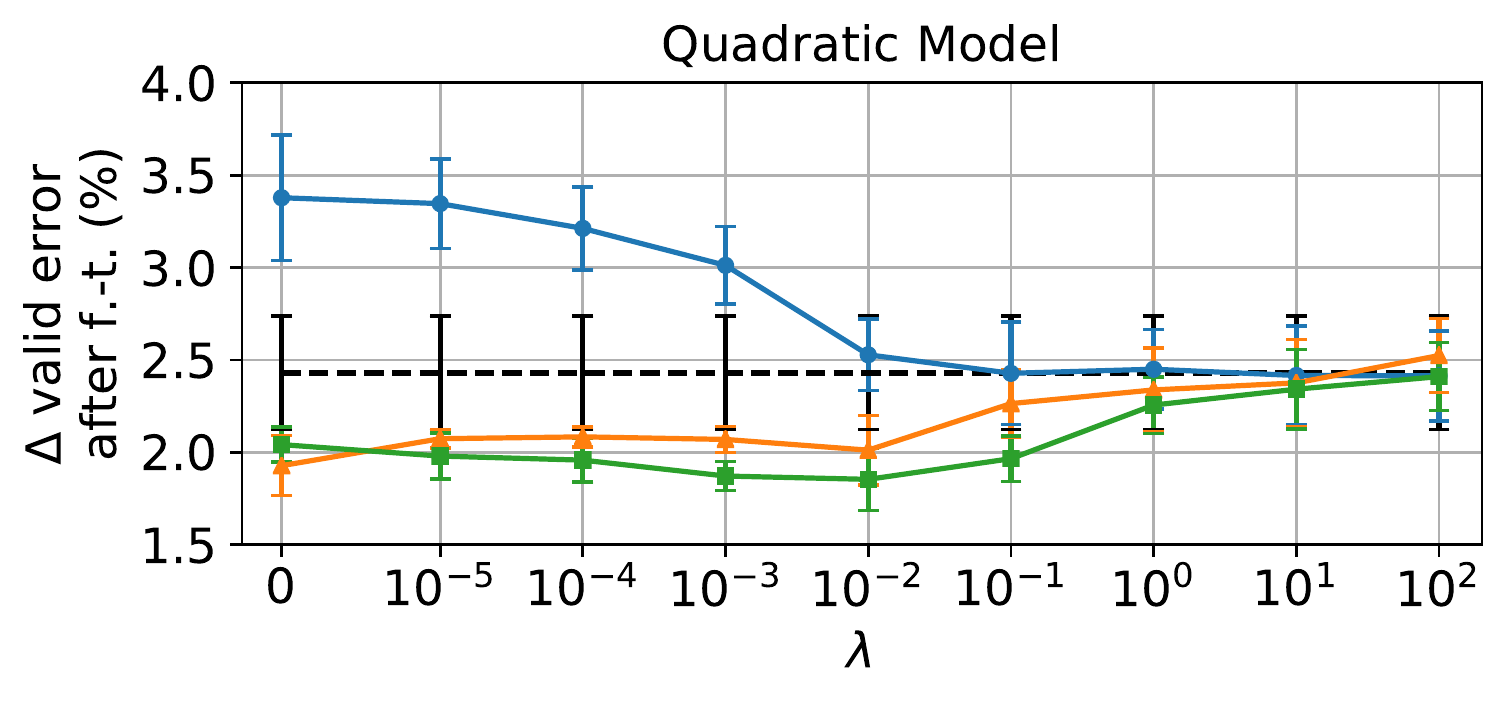}
        \includegraphics[width=0.32\textwidth]{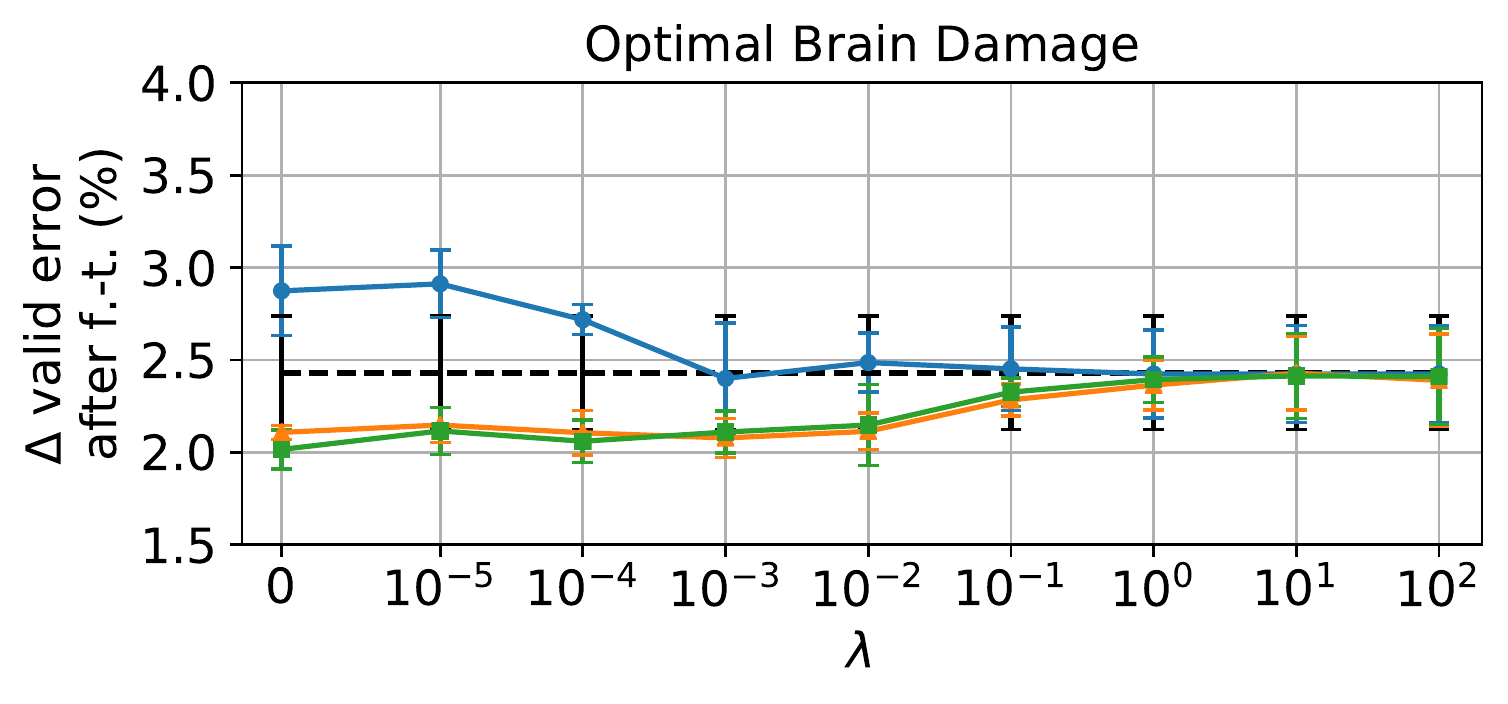}
        \caption{MLP on MNIST with 98.8\% sparsity.}
    \end{subfigure}\\
    \begin{subfigure}[b]{\textwidth}
        \centering
        \includegraphics[width=0.32\textwidth]{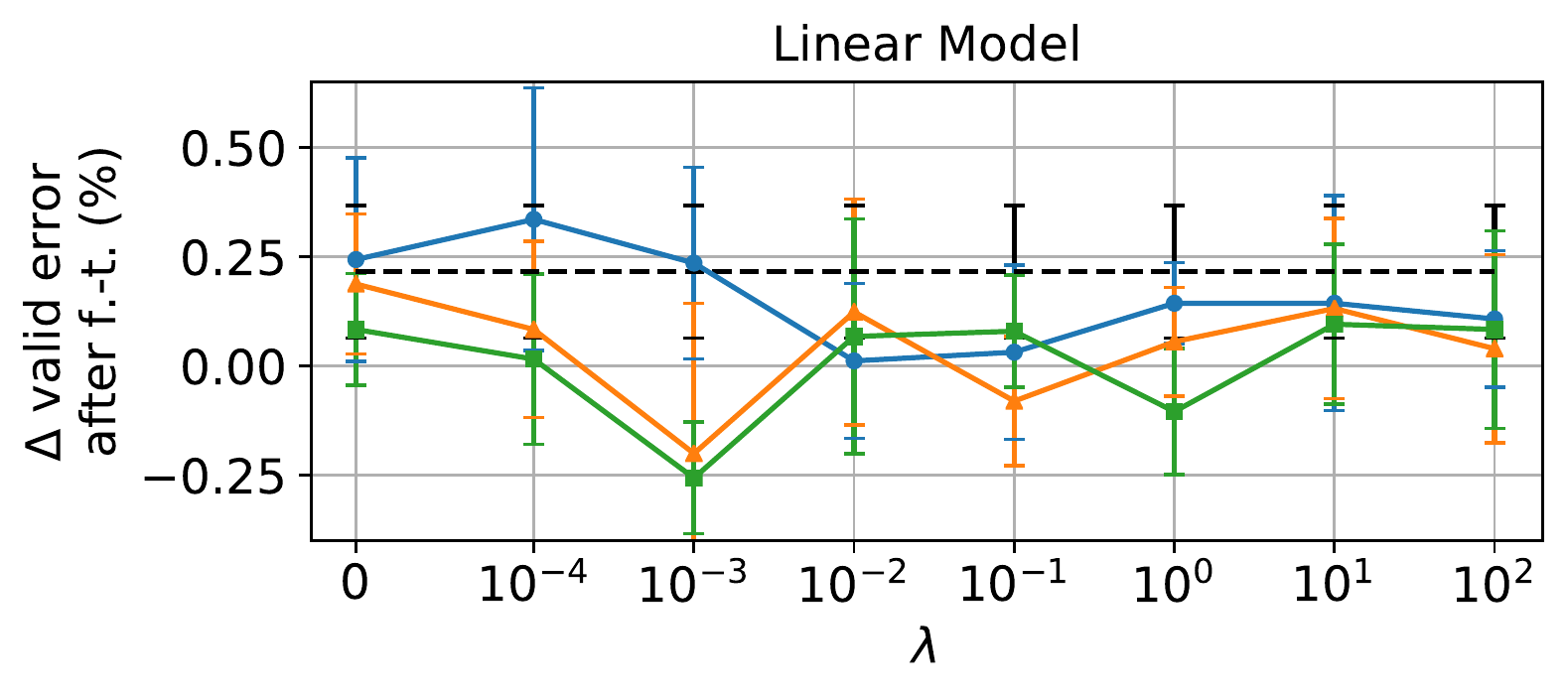}
        \includegraphics[width=0.32\textwidth]{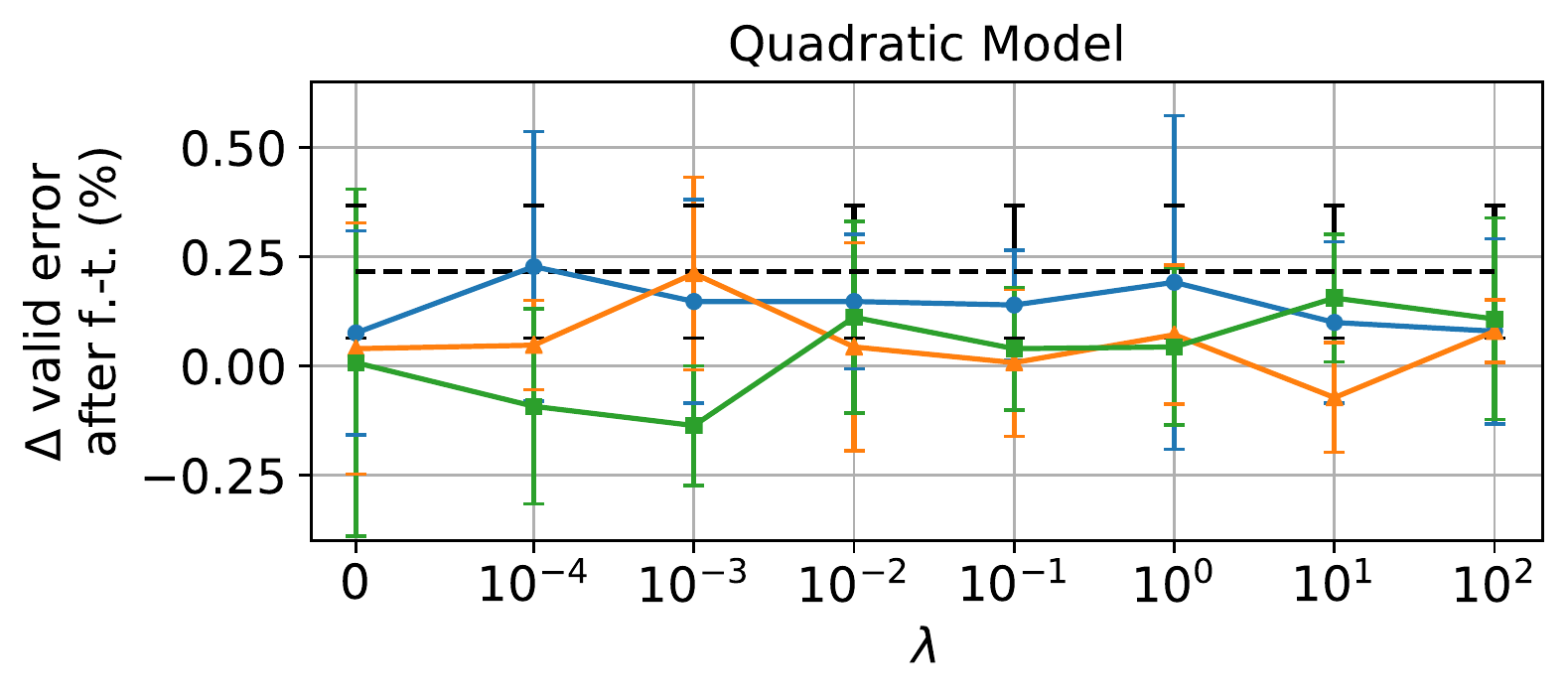}
        \includegraphics[width=0.32\textwidth]{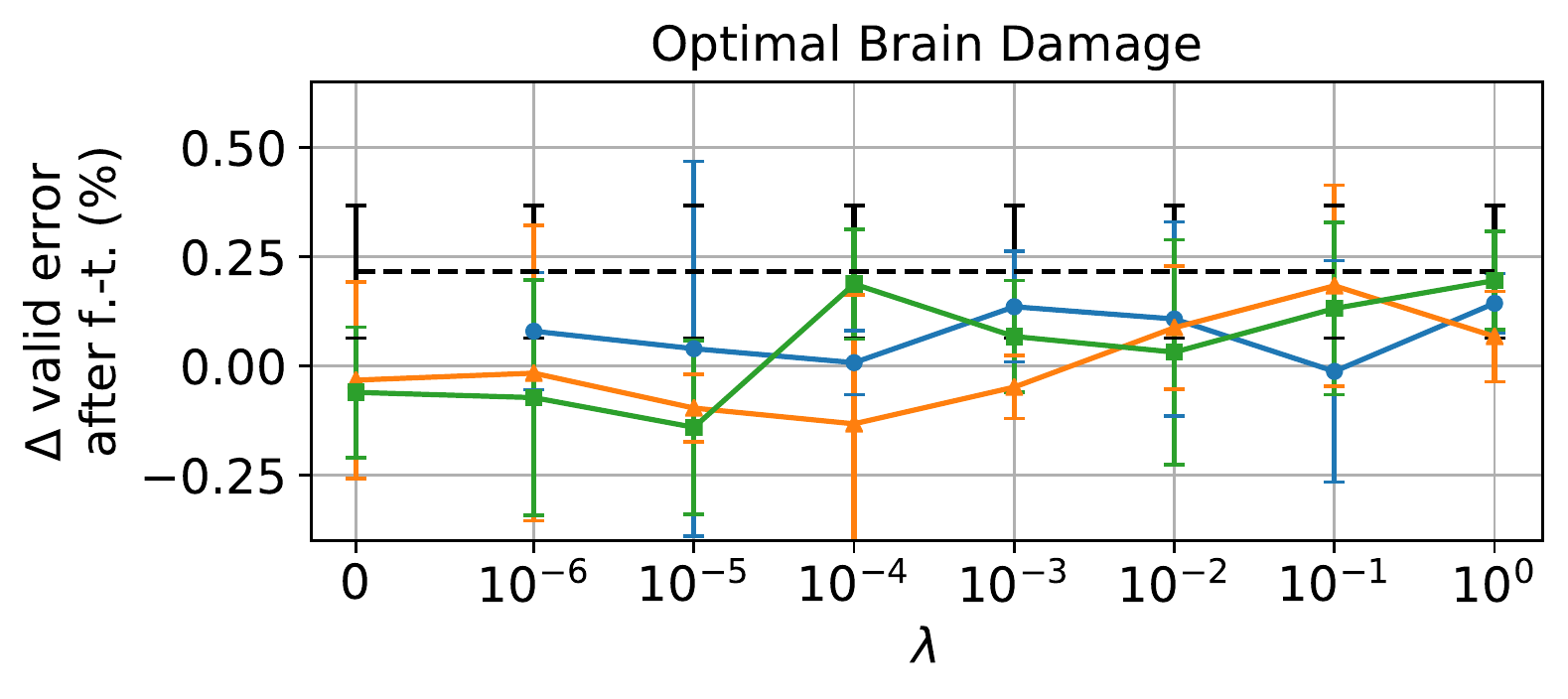}
        \caption{VGG11 on CIFAR10 with 95.6\% sparsity.}
    \end{subfigure}\\
    \begin{subfigure}[b]{\textwidth}
        \centering
        \includegraphics[width=0.32\textwidth]{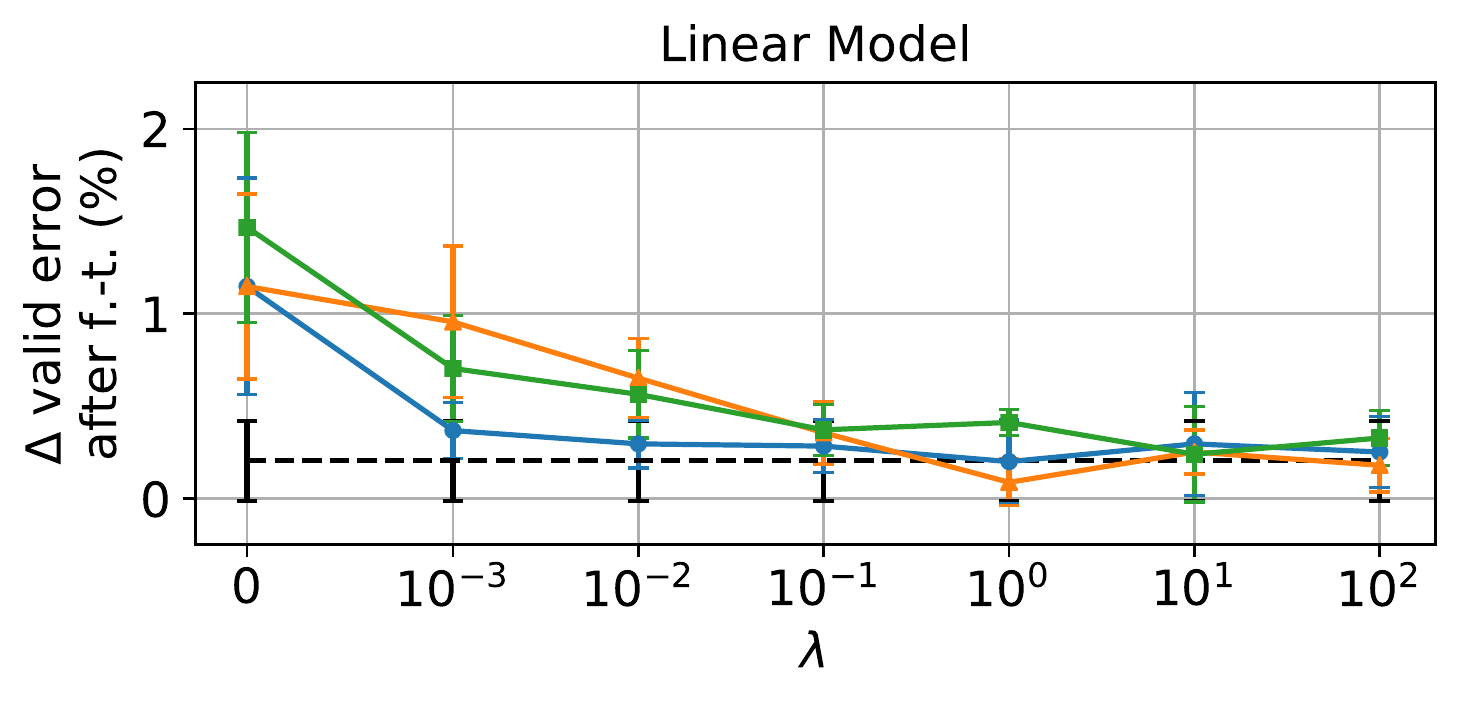}
        \includegraphics[width=0.32\textwidth]{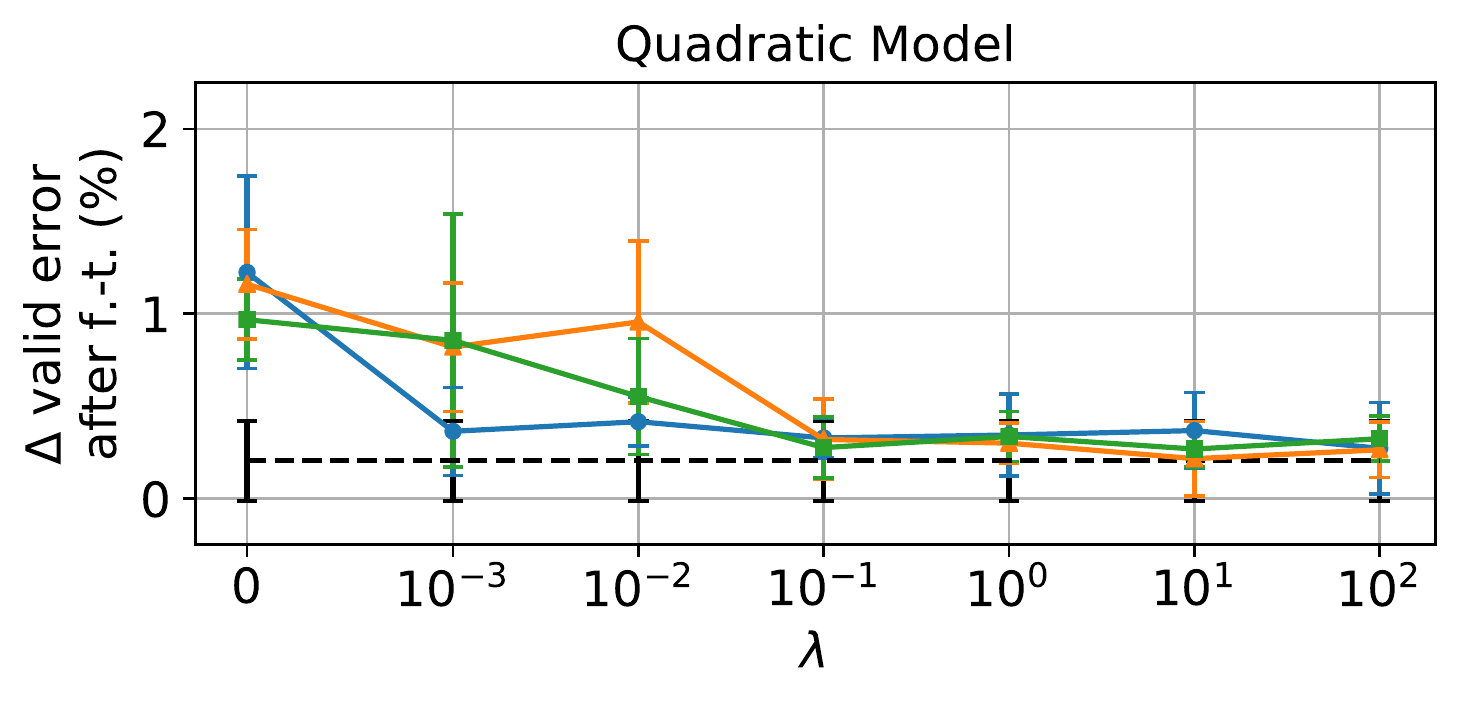}
        \includegraphics[width=0.32\textwidth]{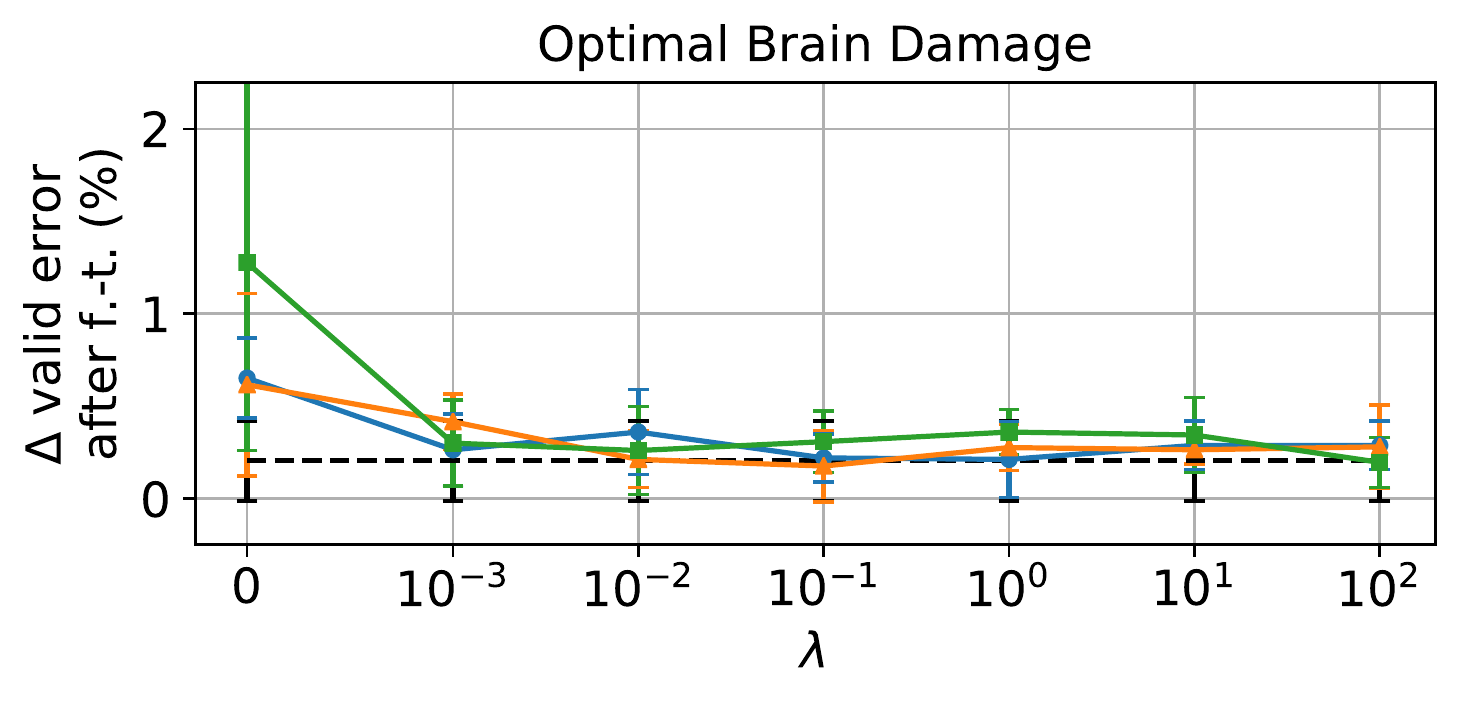}
        \caption{PreActResNet18 on CIFAR10 with 95.6\% sparsity.}
    \end{subfigure}
    \caption{Same as Figure~\ref{fig:val_lt}, but using equally spaced pruning steps. Note the difference in number of pruning iterations.}
    \label{fig:val_eq}
\end{figure}

\paragraph{Validation error figures} Figures~\ref{fig:val_eq} and~\ref{fig:val_lt} contain the same experiments than Figure~\ref{fig:loss_eq}, but displaying the validation error gap, for linear and exponential step size, respectively.

\begin{figure}[htb!]
    \centering
    \includegraphics[width=0.32\textwidth]{figures/oneshot/legend_gap.pdf}\\
    \includegraphics[height=0.32\textwidth]{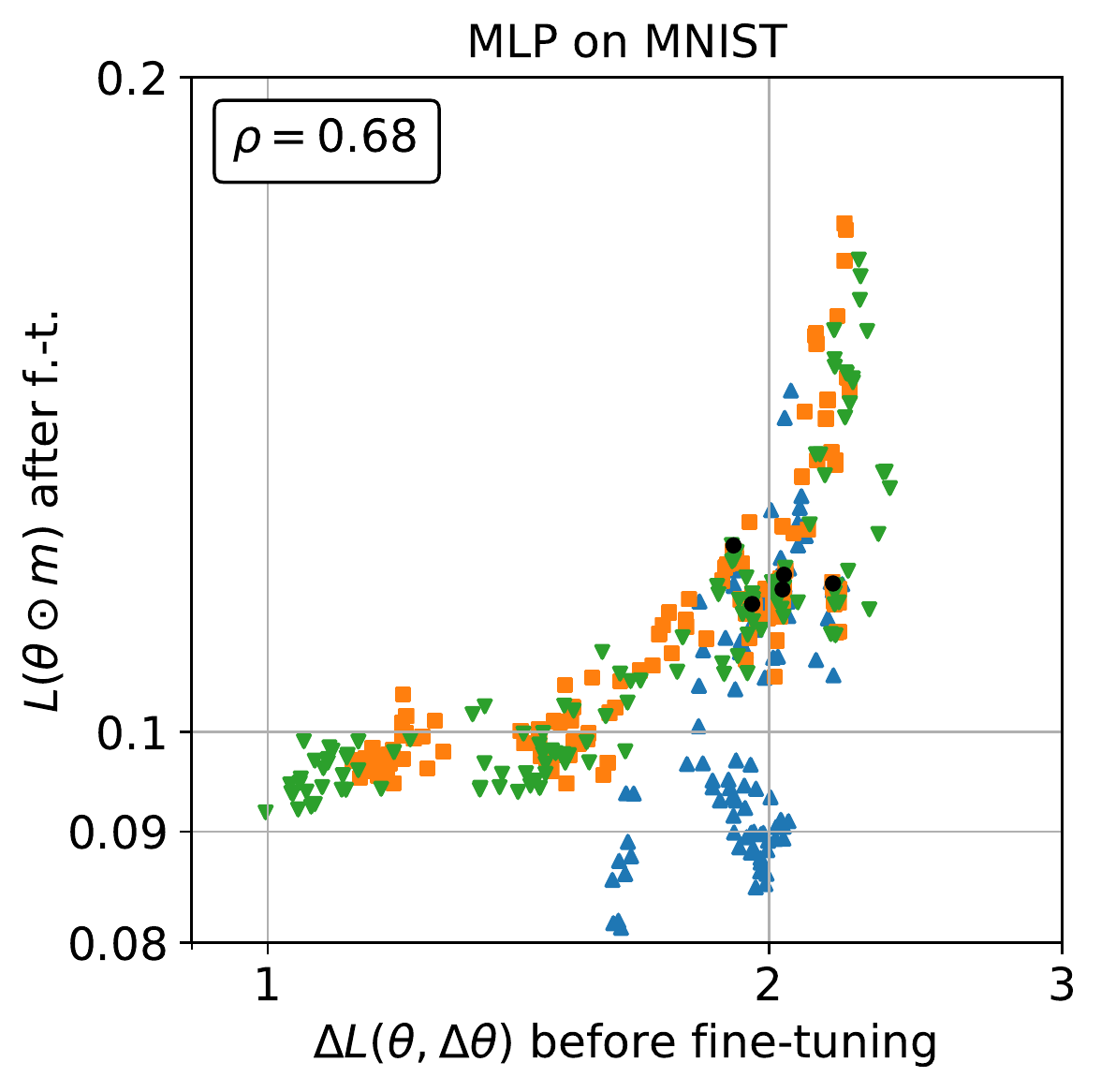}
    \includegraphics[height=0.32\textwidth]{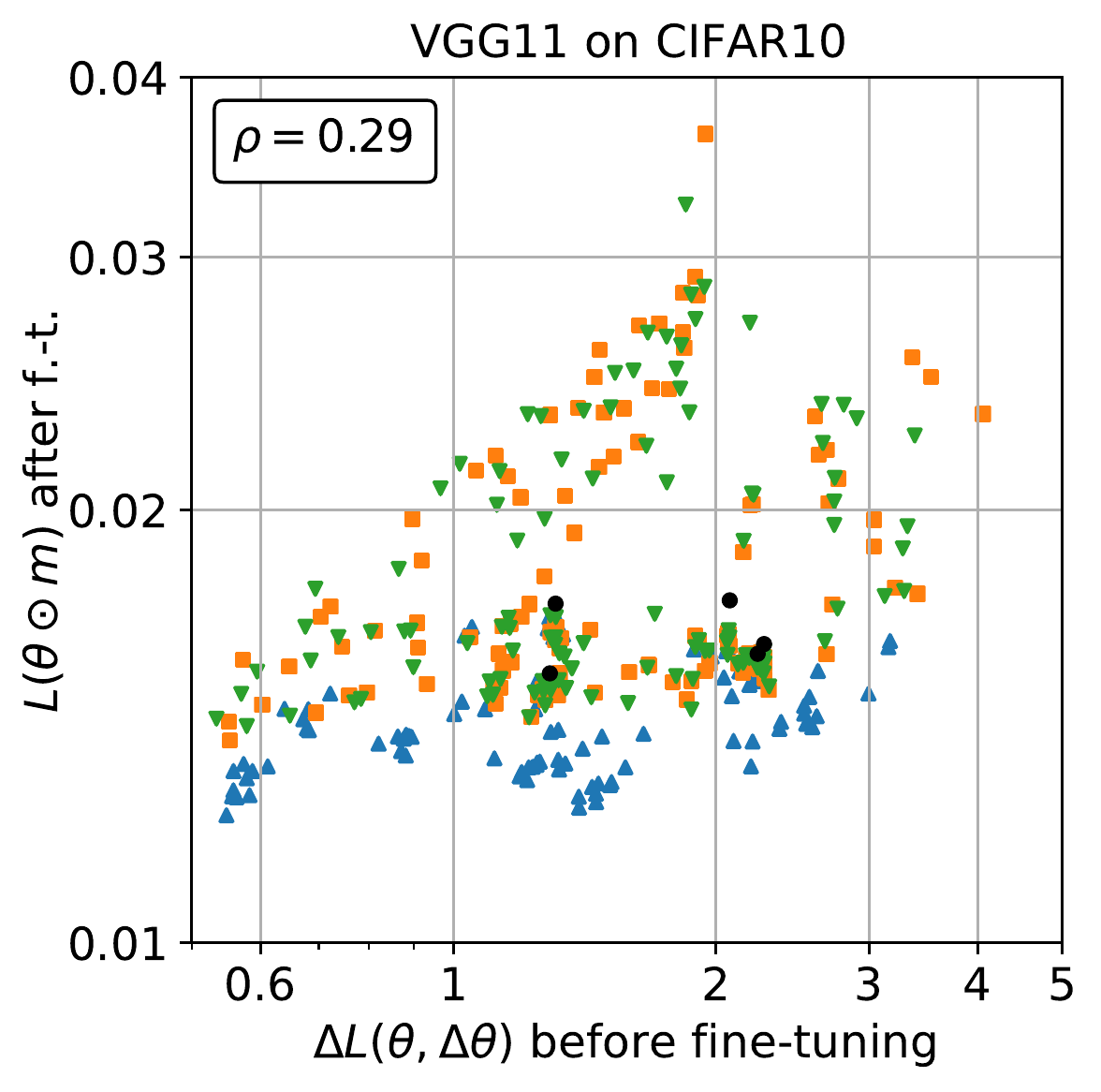}
    \includegraphics[height=0.32\textwidth]{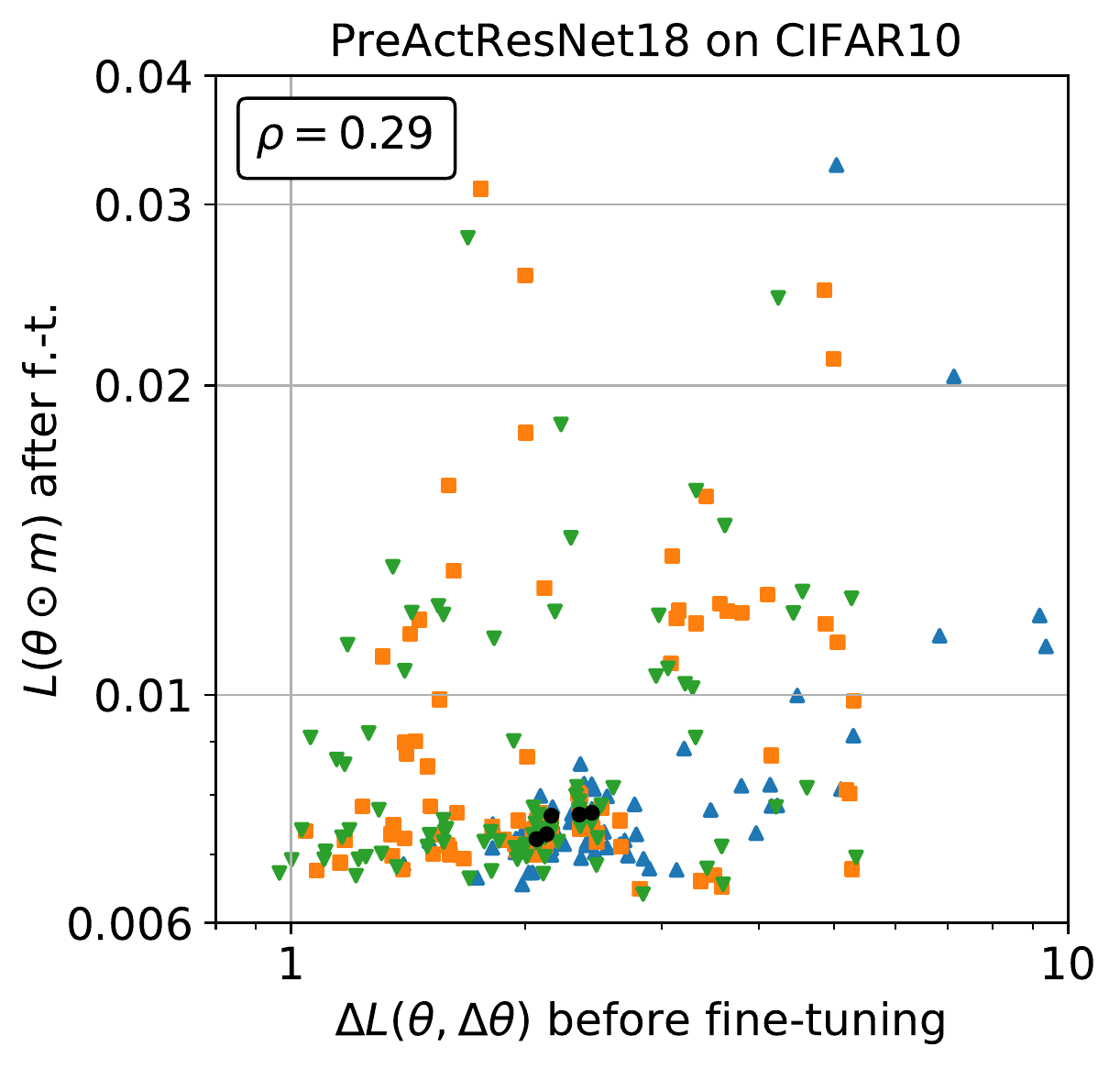}
    \caption{Same as Figure~\ref{fig:scatter}, but showing $\mathcal L(\vtheta \odot \mathbf m)$ after fine-tuning as a function of $\Delta \mathcal L(\vtheta, \Delta\vtheta)$. Except for the MLP on MNIST, there is only weak correlations between $\Delta \mathcal L(\vtheta, \Delta\vtheta)$ and $\mathcal L(\vtheta \odot \mathbf m)$ after fine-tuning.}
    \label{fig:scatter_loss}
\end{figure}

\paragraph{Training loss after fine-tuning} Figure~\ref{fig:scatter_loss} is the same as Figure~\ref{fig:scatter} but showing $\mathcal L(\vtheta \odot \mathbf m)$ after fine-tuning as a function of $\Delta \mathcal L(\vtheta, \Delta\vtheta)$. It has a similar trend as Figure~\ref{fig:scatter}: there is not much correlation between the loss before and after fine-tuning, except on MNIST.

\begin{figure}[htb!]
    \centering
    \includegraphics[width=0.32\textwidth]{figures/oneshot/legend_gap.pdf}\\
    \includegraphics[height=0.32\textwidth]{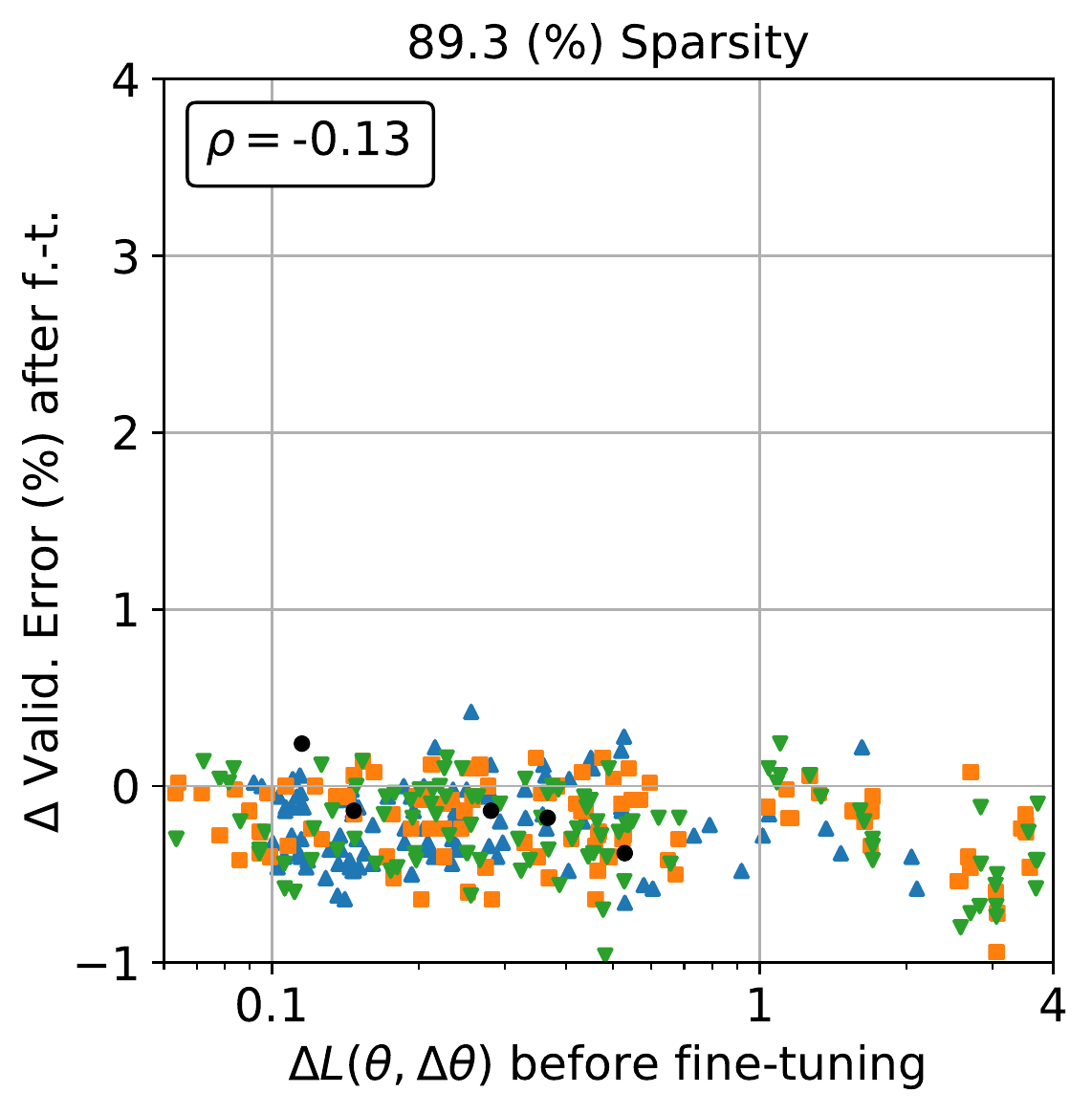}
    \includegraphics[height=0.32\textwidth]{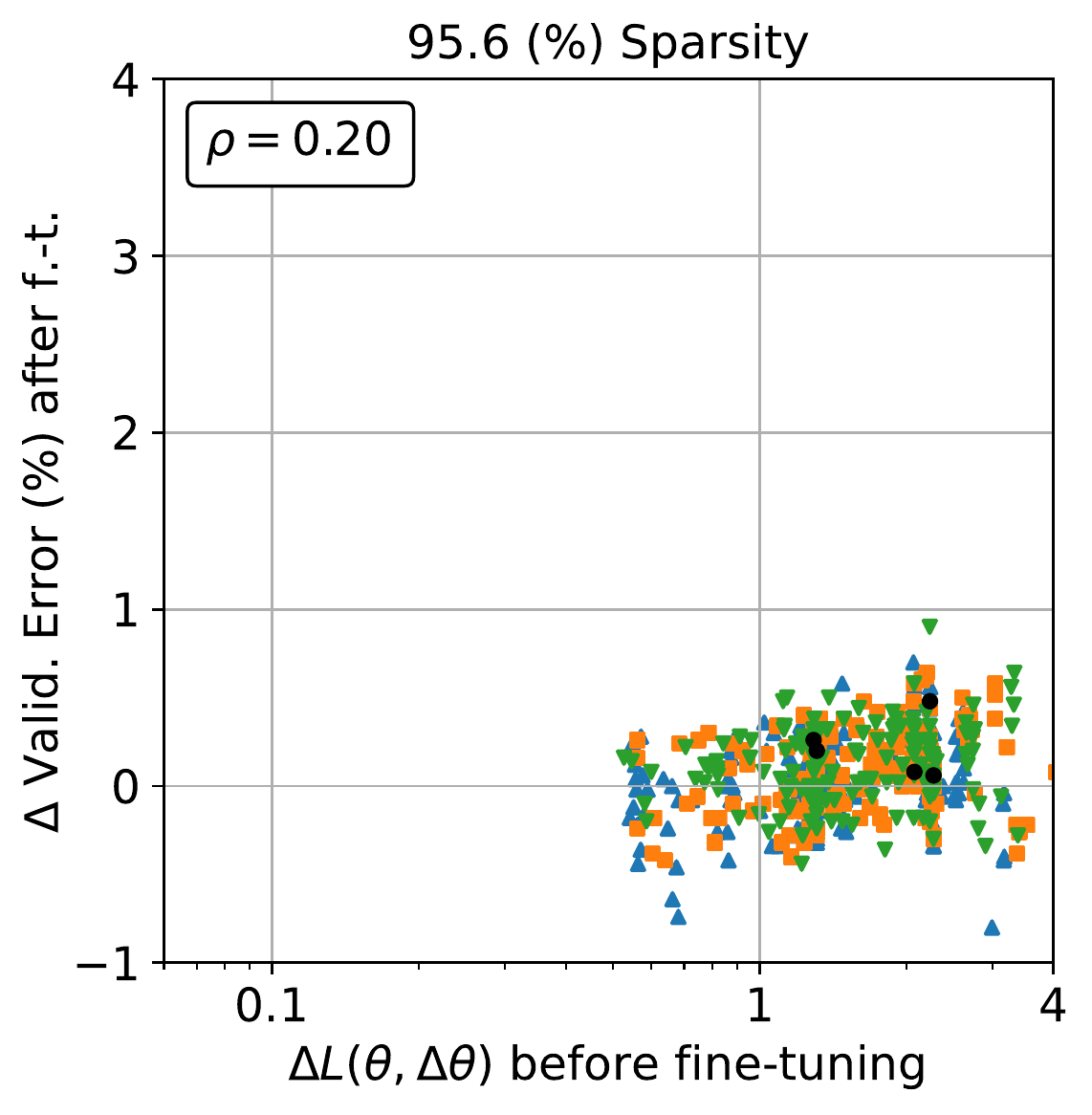}
    \includegraphics[height=0.32\textwidth]{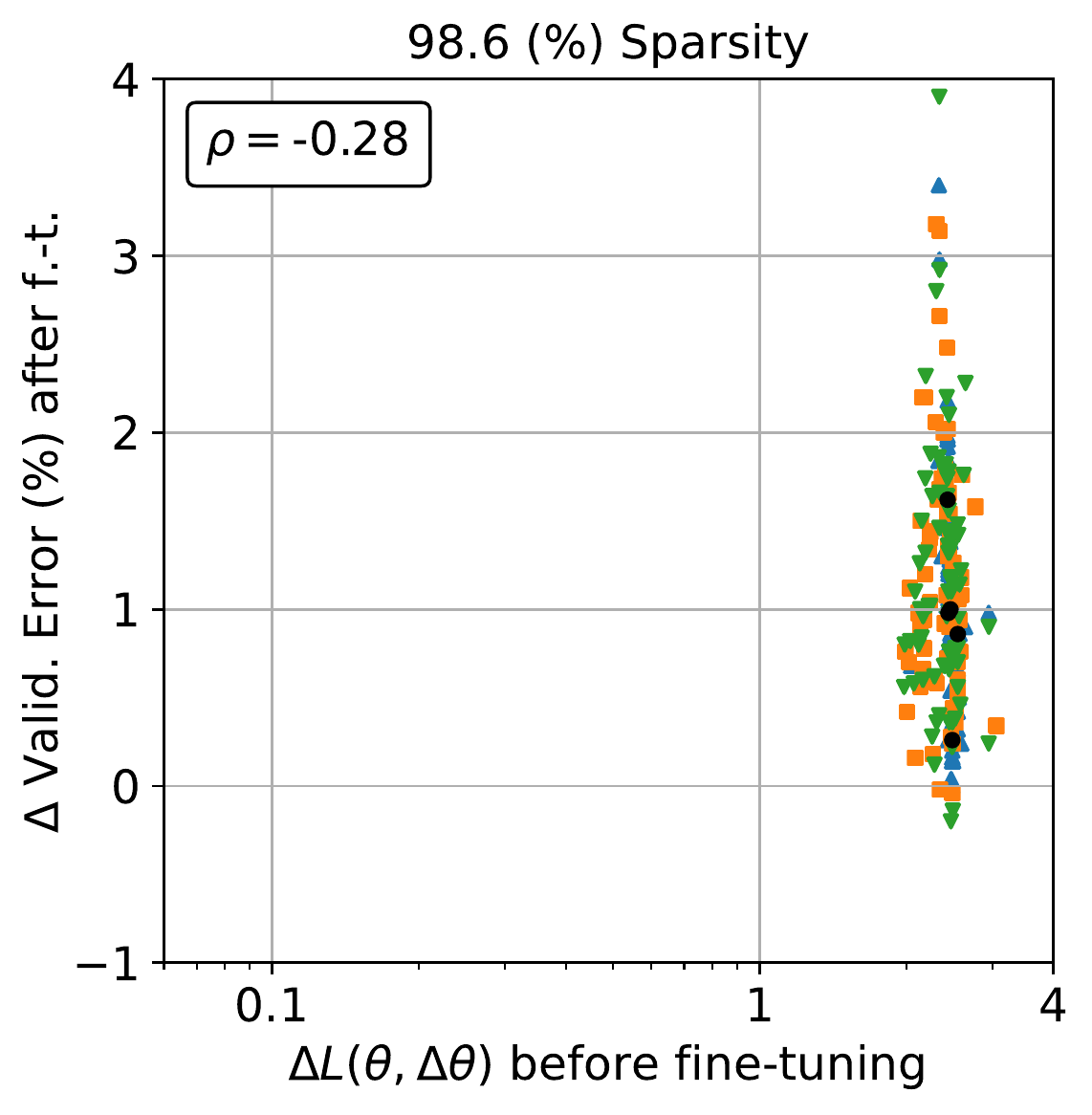}
    \caption{Same as Figure~\ref{fig:scatter}, but for different sparsity levels on the VGG11 on CIFAR10. When the sparsity is low, the network has enough capacity to return to its original performances after fine-tuning. When the sparsity is too high, then all criteria produce networks with random predictions.}
    \label{fig:scatter_pr}
\end{figure}

\paragraph{Different sparsity levels} Figure~\ref{fig:scatter_pr} shows the performances of different criteria on VGG11 on CIFAR10, for different sparsity levels. When the sparsity is low (89.3~\%), the network has enough capacity to return to its original performances after fine-tuning. When the sparsity is too high (98.6~\%), then all criteria produce networks with random predictions. There might be a sweet spot in between, but one would require more powerful model to verify this supposition.

\begin{figure}[htb!]
    \centering
    \includegraphics[width=0.32\textwidth]{figures/oneshot/legend_gap.pdf}\\
    \includegraphics[width=0.32\textwidth]{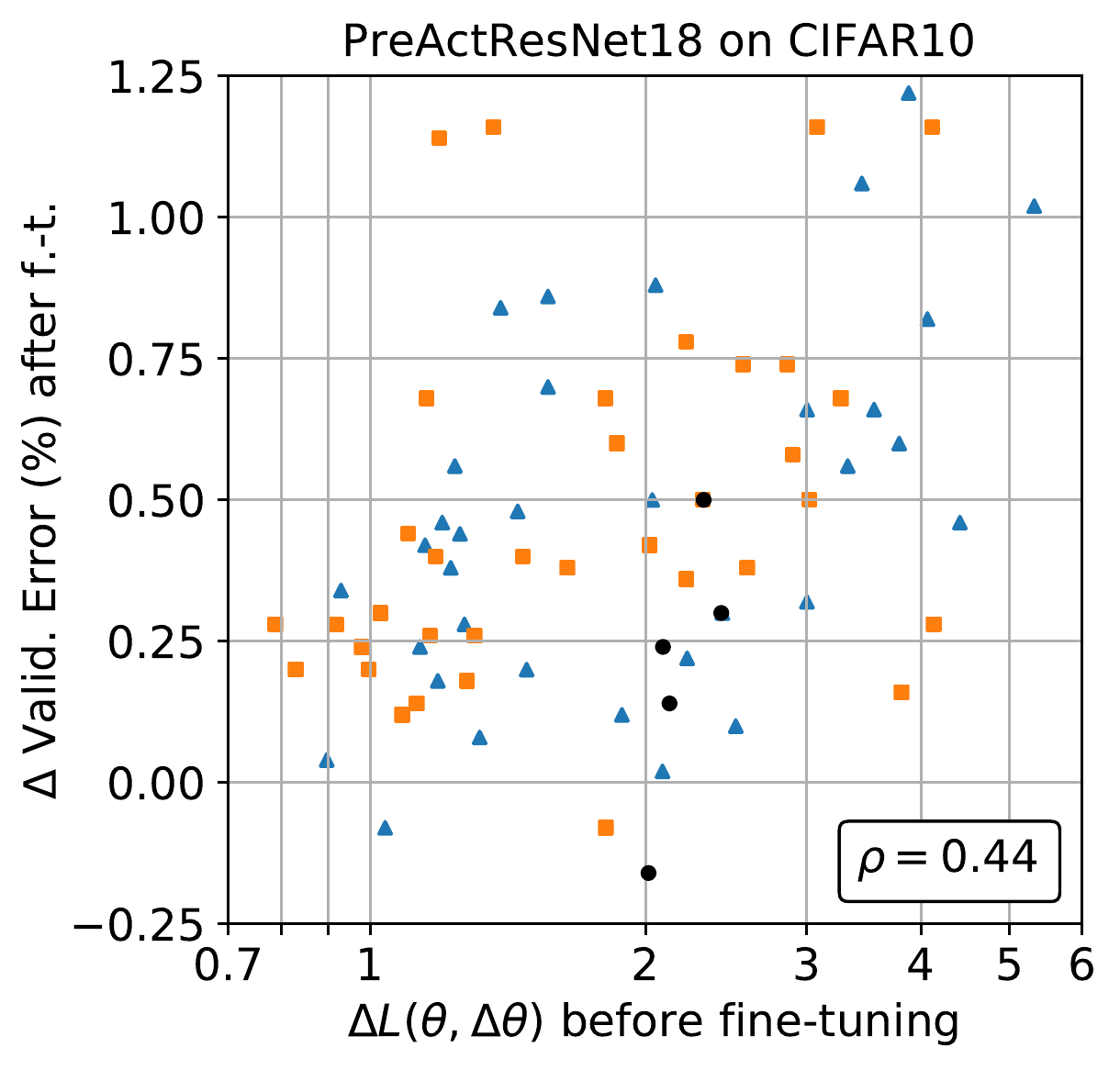}
    \includegraphics[width=0.32\textwidth]{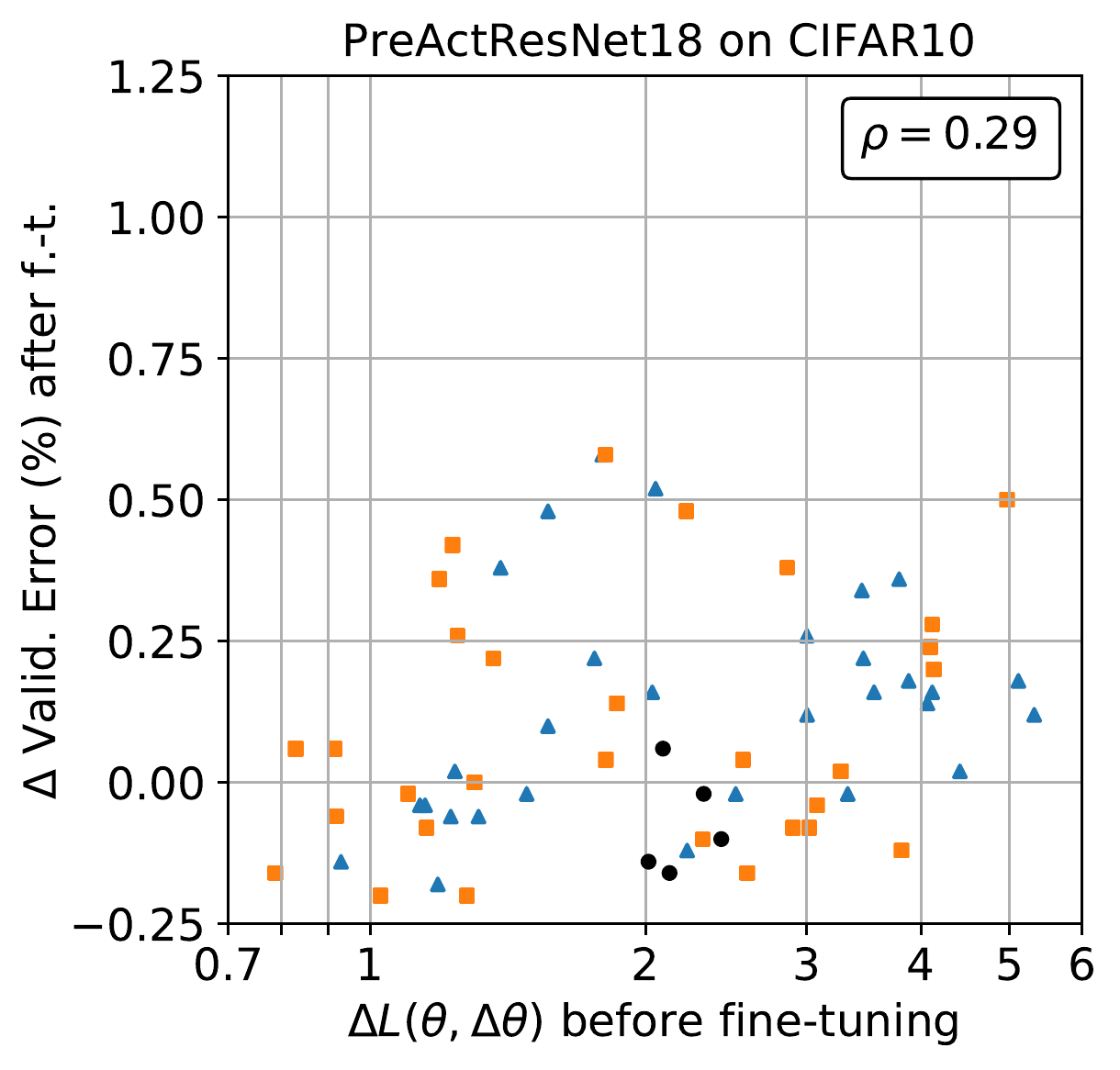}
    \caption{Left: Using the same hyper-parameters for fine-tuning as the ones of the original training. Right: Performing hyper-parameters optimisation for the fine-tuning. This figure shows that optimizing the hyper-parameters for fine-tuning can improve the performances of the network after pruning. However, it reduces the correlation between the performances after fine-tuning and $\Delta \mathcal L(\vtheta, \Delta\vtheta)$. The lack of correlation can thus not be explained by poor fine-tuning hyper-parameters.}
    \label{fig:scatter_hp}
\end{figure}

\paragraph{Hyper-parameters optimisation}
Figure~\ref{fig:scatter_hp} shows the impact of  hyper-parameter optimization for the fine-tuning phase. We performed a grid search with three different learning rate (0.1, 0.01, 0.03) and three different l2-regularisation (0, 5e-4, 5e-5). All 9 sets of hyper-parameters were tested on LM, QM and MP on 5 different random seeds.
In this set of experiments, we used $\lambda \in \{0, 0.01, 0.1, 1\}$ and $\pi \in \{14, 140\}$.
Optimizing hyper-parameters for fine-tuning can lead to better performance after fine-tuning, but does not increases the correlation between the performances after fine-tuning and $\Delta \mathcal L(\vtheta, \Delta\vtheta)$. The lack of correlation can thus not be explained by bad fine-tuning hyper-parameters.

\end{document}